\documentclass[3p,final]{elsarticle}

\usepackage{color}
\usepackage{xfrac}
\usepackage{subcaption}

\usepackage{tikz}
\newcommand{\tikzcircle}[2][black,fill=black]{\tikz[baseline=-.7ex]\draw[#1,radius=#2] (0,0) circle ;}%
\newcommand{\tikzrect}[2][fill=black]{\tikz[baseline=0.ex]\draw[rounded corners=0.7pt, fill=black]
	(0,0) rectangle ++(0.32,0.2);}
\newcommand{\tikzemptyrect}[2][fill=black]{\tikz[baseline=0.ex]\draw[rounded corners=0.7pt, fill=white]
	(0,0) rectangle ++(0.32,0.2);}

\usepackage{multicol} 
\usepackage{framed}
\usepackage{booktabs}

\usepackage{amsmath}
\newcommand{\overbar}[1]{\mkern 1.5mu\overline{\mkern-1.5mu#1\mkern-1.5mu}\mkern 1.5mu}
\journal{Journal}
\usepackage{multirow}
\usepackage{adjustbox}
\usepackage{geometry}

\begin{document}

	
	\begin{frontmatter}%
		
		\title{Thermodynamics-based Artificial Neural Networks for constitutive modeling}%

		\author[addr1,addr2]{Filippo Masi}%
		\ead{filippo.masi@ec-nantes.fr}
		
		\author[addr1]{Ioannis Stefanou\corref{cor1}}
		\cortext[cor1]{Corresponding author.}
		\ead{ioannis.stefanou@ec-nantes.fr}%
		
		\author[addr3]{Paolo Vannucci}%
		\ead{paolo.vannucci@uvsq.fr}%
			
		\author[addr2]{Victor Maffi-Berthier}%
		\ead{victor.maffi-berthier@ingerop.com}

		\address[addr1]{Institut de Recherche en Génie Civil et Mécanique,\\UMR 6183, CNRS, Ecole Centrale de Nantes, Université de Nantes,\\1 rue de la N\"{o}e, F-44300, Nantes, France.\\ \medskip}
		
		\address[addr2]{Ing\'{e}rop Conseil et Ing\'{e}nierie, \\ 18 rue des Deux Gares, F-92500, Rueil-Malmaison, France.\\ \medskip}%
		
		\address[addr3]{LMV, UMR 8100, Universit\'{e} de Versailles et Saint-Quentin,\\ 55 avenue de Paris, F-78035, Versailles, France.\\}
		
		\begin{abstract}
		Machine Learning methods and, in particular, Artificial Neural Networks (ANNs) have demonstrated promising capabilities in material constitutive modeling. One of the main drawbacks of such approaches is the lack of a rigorous frame based on the laws of physics. This may render physically inconsistent the predictions of a trained network, which can be even dangerous for real applications.
		
		Here we propose a new class of data-driven, physics-based, neural networks for constitutive modeling of strain rate independent processes at the material point level, which we define as Thermodynamics-based Artificial Neural Networks (TANNs). The two basic principles of thermodynamics are encoded in the network's architecture by taking advantage of automatic differentiation to compute the numerical derivatives of a network with respect to its inputs. In this way, derivatives of the free-energy, the dissipation rate and their relation with the stress and internal state variables are hardwired in the network. Consequently, our network does not have to identify the underlying pattern of thermodynamic laws during training, reducing the need of large data-sets. Moreover the training is more efficient and robust, and the predictions more accurate. Finally and more important, the predictions remain thermodynamically consistent, even for unseen data. Based on these features, TANNs are a starting point for data-driven, physics-based constitutive modeling with neural networks.
		
		We demonstrate the wide applicability of TANNs for modeling elasto-plastic materials, with strain hardening and strain softening. Detailed comparisons show that the predictions of TANNs outperform those of standard ANNs. TANNs ' architecture is general, enabling applications to materials with different or more complex behavior, without any modification.

		\end{abstract}

		
	\end{frontmatter}%
	
\section{Introduction}

A large spectrum of constitutive models have been proposed in the literature, based on observations and experimental testing. Existing constitutive laws can account for phenomena taking place in various length scales. This is achieved either through heuristic approaches and assumptions or through asymptotic approximations and averaging (e.g. \cite{lloberas2019reduced, Nitka2011, Feyel2003, Bakhvalov1989}). The history and the state of a material is commonly taken into account through ad hoc enrichment of simpler constitutive laws and extensive calibration. For this purpose, the laws of thermodynamics offer a useful framework for deriving more sophisticated laws, by intrinsically respecting the energy balance and the entropy production requirements  (see e.g. \cite{houlsby2007principles, einav2007coupled, houlsby2000thermomechanical} among others).

An important limitation in constitutive modeling is the availability of data at different time- and length-scales. However, with the increase of computational power, it is nowadays possible to foresee micromechanical simulations that can account for realistic physics and explore stress paths and non-linear phenomena, which are experimentally inaccessible with the current methods. Of course, some constitutive assumptions will be always necessary, but these might be at a smaller scale, where the material properties are measurable and easier to identify. This scale is for instance the scale of the microstructure of a material (e.g. the scale of sand grains, crystals, alloys' grains, composites' fibers, masonry bricks' etc. including their topological configuration).

However, it is likely that the existing constitutive models might not be sufficient for describing complex material behaviors emerging from the microstructure. Therefore, calibration (parameter fitting) of known constitutive descriptors might be insufficient for representing the full space of material response, provided by sophisticated micromechanical simulations. Moreover, micromechanical simulations have currently a tremendous calculation cost, which is impossible to afford in large-scale, non-linear, incremental simulations (e.g. Finite Elements) that are usually needed in applications (cf. \cite{masi2020, masi, RATTEZ201854, RATTEZ20181, COLLINSCRAFT2020103975, lloberas2019reduced, Nitka2011, Eijnden2016, Feyel2003}).

A promising solution to this issue seems to be \textit{Machine Learning}. According to \cite{Geron2015}, ``Machine Learning is the science (and art) of programming computers so they can learn from data''. In the context of computer programming, learning is defined by E. Tom Mitchell \cite{mitchell1997machine} as follows: ``A computer program is said to learn from experience $E$ with respect to some task $T$ and some performance measure $P$, if its performance on $T$, as measured by $P$, improves with experience''. In the frame of constitutive modeling, a Machine Learning program can learn the stress-strain behavior of a material, given examples of stress-strain increments, which are either determined experimentally or through detailed micromechanical simulations. The data that the system uses to learn are called the training data-set and each training example is called a training instance (or sample). In our case, the task $T$, for instance, can be the prediction of the stress for a given increment and internal state of the material. The experience $E$ is the training data-set and the performance measure $P$ can be the prediction error. Machine Learning is a general term to describe a large spectrum of numerical methods. Some of them offer very rich interpolation spaces, which, in theory, could be used for approximating complicated functions belonging to uncommon spaces. Here we focus on the method of Artificial Neural Networks (ANNs), which is considered to be a sub-class of Machine Learning methods. According to \citet{chen1995universal} and \citet{cybenko1989approximation}, ANNs have proved to be universal approximators, due to their rich interpolation space. Therefore, they seem to be a useful and promising tool for data-driven constitutive modeling of many materials (e.g. sand, masonry, alloys, ceramics, composites etc.).

Recognizing this potential, there is an increasing amount of new literature employing ANNs in constitutive modeling. Starting form the seminal work of \citet{ghaboussi1991} and without being exhaustive, we refer to \citet{lefik2003artificial,ghaboussi1998new, jung2006neural, HEIDER2020112875,settgast2019constitutive, ghavamian2019accelerating, liu2019exploring, lu2019data, xu2020learning, huang2020learning, mozaffar2019deep, frankel2019predicting, liu2019exploring,gajek2020micromechanics} and references therein. The main idea in these works is to appropriately train ANNs, feeding them with material data, and predict the material response at the material point level. In this sense ANNs can be seen as rich interpolation spaces, able to represent complex material behavior. The Boundary Value Problem (BVP), set to determine the behavior of a solid under mechanical and/or multiphysics couplings, is then solved by replacing the standard constitutive equations or algorithms by the trained ANN. This replacement is straightforward and non-intrusive in Finite Element codes. It is worth emphasizing that the aforementioned data-driven approaches are different from another promising data-driven method (i.e., data driven computing \cite{Kirchdoerfer2016DatadrivenCM}) in which the BVP is solved directly from experimental material data (measurements), bypassing the empirical material modeling step, involving the calibration of constitutive parameters \cite{Kirchdoerfer2016DatadrivenCM,ibanez2017data, doi:10.1002/nme.5716, kirchdoerfer2018data,ibanez2018manifold, EGGERSMANN201981}. While data-driven computing can be extremely powerful in many applications \cite{EGGERSMANN201981}, the first class of methods above-mentioned (based on the constitutive behavior at the material point level) can be advantageous when modeling complex and abstract constitutive behaviors, which are not a priori known. Moreover, they can be used even if the BVP does not have a unique solution due to important non-linearities and bifurcation phenomena (e.g. loss of uniqueness, strain localization at the length of interest, multiphysics, runway instabilities etc.).

Nevertheless, until now ANNs for constitutive modeling are mainly used as a `black-box' mathematical operator, which once trained on available data-sets, does not embody the basic laws of thermodynamics. As a result, vast amount of high quality data (e.g. with reduced noise and free of outliers) are needed to enable ANNs to identify and learn the underlying thermodynamic laws. Moreover, nothing guarantees that the predictions of trained ANNs will be thermodynamically consistent, especially for unseen data.

In this paper, we encode in the ANN architecture the two basic laws of thermodynamics. This assures thermodynamically consistent predictions, even for unseen data (which can exceed the range of training data-sets). Therefore, we assure thermodynamically consistent network's predictions, both for seen and unseen data (which can exceed the range of the training data-sets). Moreover, our network does not have to identify/learn the underlying pattern of thermodynamical laws. Consequently, smaller data-sets are needed in principle, the training is more efficient and the accuracy of the predictions higher. The price to pay, in comparison with existing approaches, is the need of two additional scalar functions (outputs) in the training data-set. These are the free-energy and the dissipation rate. However, these quantities are easily accessible in micromechanical simulations (e.g. \cite{Nitka2011, Eijnden2016, Feyel2003}) and can also be obtained experimentally in some cases. Then, based on classical derivations in thermodynamics (e.g. \cite{houlsby2007principles, einav2012unification}) specific interconnections are programmed inside our ANN architecture to impose the necessary thermodynamic restrictions. These thermodynamic restrictions concern the stresses and internal state variables and their relation with the free-energy and the dissipation rate. Our approach is inspired by the so-called Physics-Informed Neural Networks (PINNs) \cite{raissi2019physics}, in which \textit{reverse-mode autodiff} \cite{baydin2017automatic} is used, allowing the numerical calculation of the derivative of an ANN with respect to its inputs.

The calculation of these derivatives, imposes some numerical requirements regarding the mathematical class of the activation functions to be used. More specifically, the internal ANN restrictions, derived from the first law of thermodynamics, require activation functions whose second gradient does not vanish. Otherwise, the problem of \textit{second-order vanishing gradients}, as it is called here (cf. classical \textit{vanishing gradients} problem in ANNs \cite[e.g.][]{Geron2015}), can inhibit back-propagation and make training to fail. This new problem and its remedy is extensively explored and discussed herein.

For the sake of simplicity and for distinguishing our approach from existing ones, we call the proposed ANN architecture Thermodynamics-based Artificial Neural Networks (TANNs). In our opinion TANN should be the starting point for data-driven and physics-based constitutive modeling at the material point level. For the implementation of TANNs, we leverage Tensorflow v2.0, an open-source symbolic tensor manipulation software library \cite{abadi2016tensorflow}. Other libraries/packages can be used as well.

The paper is structured as follows. Section \ref{sec:theory} presents a brief summary of the theoretical background of thermodynamics. In Section \ref{sec:ann} an overview of the methodology proposed and architecture of TANN is given. The main differences with classical, standard ANNs for material constitutive modeling are also discussed. Particular attention is given to the choice of activation functions and the issue of second-order vanishing gradient is investigated in detail. Generation of material data-sets, with which the training of ANNs is performed, is presented in Section \ref{sec:generation}. Some applications of TANN for uni-dimensional and three-dimensional elasto-plastic material models are presented in Section \ref{sec:application}. Extensive comparisons with standard ANNs which are not based on thermodynamics are also presented. Supplementary figures and data are available in Supplementary Material (SM) file. All code accompanying this manuscript is available on request.
\section{Thermodynamics principles: energy conservation and dissipation inequality}
\label{sec:theory}

\subsection{Energy conservation}
A convenient way to express the (local) energy conservation is
\begin{equation}
 \rho \dot{\text{\textsf{e}}} = \sigma\cdot \text{D}^{\text{Sym}}\text{v} - \text{div} q +\rho h,
\label{eq:firstP4}
\end{equation}
with $\rho$ being the material density; $\text{\textsf{e}}$ the specific internal energy (per unit mass); $\sigma$ the Cauchy stress tensor; $\text{D} \text{v}$ the spatial velocity gradient tensor; $q$ the rate of heat flux per unit area; $h$  the specific energy source (supply) per unit mass, and "$\cdot$" denotes contraction of adjacent indices.

\subsection{Second principle}
The second law of thermodynamics can be formulated in terms of the local Clausius-Duhem inequality
\begin{equation}
\rho \dot{\text{\textsf{s}}} \geq \frac{\rho h}{\theta}- \text{div}\,\left( \frac{q\cdot n}{\theta} \right),
\label{eq:secondP3}
\end{equation}
with $\text{\textsf{s}}$ being the specific (per unit mass) entropy; $h/\theta$ and $-(q\cdot n) /\theta$ the rate of entropy supply and flux, respectively.
By removing the heat supply $h$ between the energy equation (\ref{eq:firstP4}) and the entropy inequality (\ref{eq:secondP3}) leads to
\begin{equation}
\rho \left( \theta \dot{\text{\textsf{s}}}-\dot{\text{\textsf{e}}}\right) + \sigma\cdot \text{D}^{\text{Sym}}\text{v} - \dfrac{q\cdot \text{D}\theta}{\theta} \geq0,
\label{eq:thermo1}
\end{equation}
where the first two terms represent the rate of mechanical dissipation $ \text{\textsf{D}}= \rho \left( \theta \dot{\text{\textsf{s}}}-\dot{ \text{\textsf{e}}}\right) + \sigma\cdot \text{D}^{\text{Sym}}\text{v}$ and the latter the thermal dissipation rate, i.e., $ \text{\textsf{D}}^{th}= - \frac{q\cdot \text{D}\theta}{\theta}$. The thermal dissipation is non-negative because heat only flows from regions of higher temperature to lower temperature$-$that is, the heat flux $q$ is always in the direction of the negative thermal gradient. As it follows we argue that the mechanical dissipation rate must itself be non-negative (point-wise), i.e., $ \text{\textsf{D}}\geq0$.

\subsection{Dissipation function}
The definition of the (mechanical) dissipation rate $ \text{\textsf{D}}$ leads to
\begin{equation}
\rho \dot{\text{\textsf{e}}}= \rho\theta \dot{\text{\textsf{s}}} + \sigma\cdot \text{D}^{\text{Sym}}\text{v} -  \text{\textsf{D}}.
\label{eqh:udot}
\end{equation}
Let define the specific (per unit volume) internal energy $\text{\textsf{E}}= \rho \text{\textsf{e}}$ and entropy $\text{\textsf{S}}= \rho \text{\textsf{s}}$ and further assume constant material density, i.e., $\frac{\text{d}}{\text{d}t}\rho=0-$that is, $\dot{\text{\textsf{E}}}= \rho \dot{\text{\textsf{e}}}$ and $\dot{\text{\textsf{S}}}= \rho \dot{\text{\textsf{s}}}$.
We shall assume a small strain regime, i.e., $\text{D} u\ll 1$, with $\varepsilon:=\text{D}^{\text{Sym}} u$ the small strain tensor, where $u$ is the displacement vector field, and $\dot{\varepsilon}:=\text{D}^{\text{Sym}} \text{v}$ its rate of change.
Equation (\ref{eqh:udot}) hence becomes
\begin{equation}
\dot{\text{\textsf{E}}} = \theta \dot{\text{\textsf{S}}}  + \sigma\cdot \dot{\varepsilon} -  \text{\textsf{D}}.
\label{eqh:Fdot}
\end{equation}
Let assume a strain-rate independent material such that
\begin{equation}
\text{\textsf{E}}:=\widetilde{\text{\textsf{E}}}\left(\text{\textsf{S}}, \varepsilon, \mathcal{Z}\right),
\end{equation}
and
\begin{equation}
 \text{\textsf{D}}:=\widetilde{ \text{\textsf{D}}}\left(\text{\textsf{S}}, \varepsilon, \mathcal{Z},  \dot{\mathcal{Z}}\right),
\end{equation}
where $\mathcal{Z}=(\zeta_i, \dots, \zeta_N)$ denotes a set of $N$ (additional) internal state variables, $\zeta_i$, $i=1, \dots, N$. We define here (thermodynamic) state variables those macroscopic quantities characterizing the state of a system, see e.g. \cite{maugin1994thermodynamics}. The physical representation of $\zeta_i$ is not \textit{a priori} prescribed. For instance, in the case of isotropic damage, $\zeta$ is a scalar; in anistotropic damage, a tensor; in the case of elasto-plasticity, a second order tensor, etc.  The generalization to a finite-strain formulation can be achieved by considering the deformation gradient, $F$, and the first Piola-Kirchhoff tensor, $P$, as strain and stress measures, respectively (see e.g. \cite{mariano2015fundamentals} and \cite{ANAND2012116}). Nevertheless, as it would presented in Section \ref{sec:ann}, an incremental formulation of the material response is herein adopted. Therefore, the hypothesis of a small strain regime is usually realistic, at least for a large class of materials and an updated Lagrangian scheme.\\
Time differentiation of the internal energy gives
\begin{equation}
\dot{\text{\textsf{E}}}= \dfrac{\partial \text{\textsf{E}}}{\partial\text{\textsf{S}}}\cdot \dot{\text{\textsf{S}}}+\dfrac{\partial \text{\textsf{E}}}{\partial\varepsilon}\cdot \dot{\varepsilon} + \sum_{i=1}^N\dfrac{\partial\text{\textsf{E}}}{\partial\zeta}_i\cdot \dot{\zeta}_i,
\end{equation}
which is equal to (\ref{eqh:Fdot}) and, grouping terms, it leads to
\begin{equation}
\left(\dfrac{\partial \text{\textsf{E}}}{\partial \text{\textsf{S}}} -\theta \right) \dot{\text{\textsf{S}}} +
\left(\dfrac{\partial \text{\textsf{E}}}{\partial \varepsilon}  - \sigma \right)\cdot \dot{\varepsilon} - \left(\sum_{i=1}^N\dfrac{\partial\text{\textsf{E}}}{\partial\zeta}_i\cdot \dot{\zeta}_i  +  \text{\textsf{D}} \right)=0.
\label{eqh:equatingu}
\end{equation}
The arbitrariness of $\dot{\text{\textsf{S}}}$, $\dot{\varepsilon}$, and $\dot{\zeta}$ leads to the following relations
\begin{subequations}
	\begin{align}
	\theta = \dfrac{\partial  \text{\textsf{E}}}{\partial  \text{\textsf{S}}},\\
	\sigma = \dfrac{\partial  \text{\textsf{E}}}{\partial \varepsilon},\\
	\sum_{i=1}^N\dfrac{\partial\text{\textsf{E}}}{\partial\zeta}_i\cdot \dot{\zeta}_i  +  \text{\textsf{D}} =0.
\end{align}
\end{subequations}
Further introducing the thermodynamic stress, conjugate to $\zeta_i$, $\mathcal{X}=(\chi_1, \ldots, \chi_N)$, with 
\begin{equation}
\chi_i := - \frac{\partial \text{\textsf{E}}}{\partial \zeta_i} \qquad \forall \: i \in \left[ 1,N\right],
\label{eq:chi}
\end{equation}
we obtain the following, alternative definition of the dissipation
\begin{equation}
 \text{\textsf{D}}=\sum_{i=1}^{N} \chi_i\cdot \dot{\zeta}_i
\end{equation}

\subsection{Isothermal processes}
In the case of isothermal process, the (specific) Helmholtz free-energy, $\text{\textsf{F}}:=\text{\textsf{E}}-\text{\textsf{S}}\theta = \widetilde{\text{\textsf{F}}}(\theta, \varepsilon, \mathcal{Z})$, which is the Legendre transform conjugate of $\text{\textsf{e}}$, is preferable. In this case, the dissipation rate is such that $\text{\textsf{D}} := \widetilde{\text{\textsf{D}} }(\theta, \varepsilon, \mathcal{Z}, \dot{\mathcal{Z}})$. The equations presented above (\ref{eqh:equatingu}-\ref{eqh:normality}) still hold (by replacing $\text{\textsf{E}}$ with $\text{\textsf{F}}$)
\begin{equation}
\label{eq:energy_expr}
 \text{\textsf{S}} =- \dfrac{\partial  \text{\textsf{F}}}{\partial \theta}, \quad \sigma = \dfrac{\partial  \text{\textsf{F}}}{\partial \varepsilon}, \quad  \text{\textsf{D}}=-\sum_i \dfrac{\partial  \text{\textsf{F}}}{\partial \zeta_i}\cdot \dot{\zeta}_i=\sum_i \chi_i \cdot \dot{\zeta}_i.
\end{equation}

\section{Thermodynamics-based Artificial Neural Networks}
\label{sec:ann}
Within the framework of ANN material models, we can distinguish two main classes. The first consists of direct, so-called ``black-box'', approaches, where the information flow passes through the machine learning tool (usually feed-forward or recurrent artificial neural networks) which operates as a mere regression operator, see e.g. \cite{ghaboussi1991, lefik2003artificial}. The second class coincides with ANN models incorporating some knowledge in an informed, guided graph, see e.g. \cite{HEIDER2020112875}. Both classes, however, are affected by the lack of physics, being the predictions not always compatible with thermodynamic principles (at least). 
\begin{figure*}[ht]
\centering
\begin{subfigure}[b]{0.3\textwidth}
  \centering
  \includegraphics[width=0.4\linewidth]{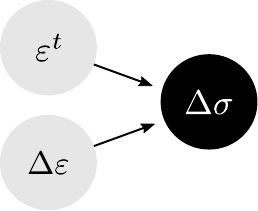}
	\caption{\footnotesize black-box (BB) network.}
	\label{f:blackbox_a}
\end{subfigure}
\begin{subfigure}[b]{0.3\textwidth}
  \centering
  \includegraphics[width=0.4\linewidth]{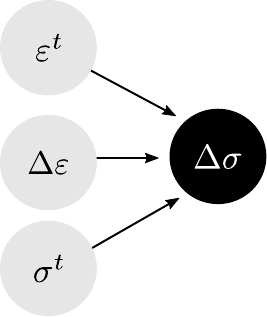}
	\caption{\footnotesize \textit{informed} neural network (i-NN1).}
	\label{f:blackbox_b}
\end{subfigure}
\begin{subfigure}[b]{0.3\textwidth}
  \centering
  \includegraphics[width=0.4\linewidth]{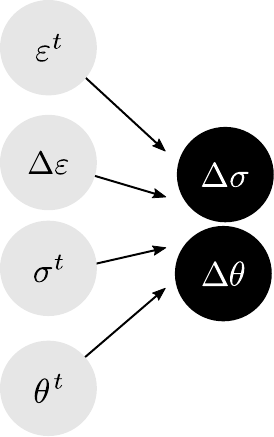}
	\caption{\footnotesize \textit{informed} neural network (i-NN1).}
	\label{f:blackbox_c}
\end{subfigure}
\caption{Examples of direct, black-box (\textsf{BB}) (left) and informed (right) neural networks for material laws modeling. Inputs are highlighted in gray (\tikzcircle[lightgray, fill=lightgray]{3pt}), outputs in black (\tikzcircle{3pt}).}
	\label{f:blackbox}
\end{figure*}
Figure \ref{f:blackbox_a} depicts the direct approach (\textsf{BB}), in which ANNs, either forward fed or recurrent, are used to predict the stress increment, (output, $\mathcal{O}$) $\mathcal{O}=\Delta \sigma=\sigma^{t+\Delta t} - \sigma^t$, from the input $\mathcal{I}=(\varepsilon^{t}, \Delta \varepsilon)$, being $\varepsilon^{t}$ the precedent strain state and $\Delta \varepsilon$ its increment. In concise form, we write $\mathcal{O} = \textsf{BB}@\, \mathcal{I}$. In this scheme, $\varepsilon^t$ and $\Delta \varepsilon$ can be regarded as the state variables, namely the \textit{ANN state variables} (not necessarily coinciding with those introducted in Sect. \ref{sec:theory}), on which the updated material stress depends on. Two examples of guided, informed ANNs are illustrated in Figures \ref{f:blackbox_b} and \ref{f:blackbox_c}. In both cases, the ANN intrinsically accounts for path-dependency, see e.g. \cite{HEIDER2020112875}, making sequence of predictions of the main output. The network \textsf{i-NN1} makes use of the last predicted output, i.e., $\sigma^t$, to make predictions of the next output, $\mathcal{O}=\Delta \sigma$. The inputs are hence $\mathcal{I} = (\varepsilon^t, \Delta \varepsilon, \sigma^t)$. We shall notice that, differently from \textsf{BB}, the stress at the precedent state, $\sigma^t$, is also considered to be an ANN state variable. Other alternatives exist in the selection of the ANN variables of state. One may chose, as we shall see in Section \ref{sec:application}, (thermodynamic) state variables  to be ANN state variables.\\
In the case of temperature-dependent material response, the second case (\textsf{i-NN2}) allows to make predictions that depend on the precedent temperature state, $\theta^t$, namely $\mathcal{O} = \textsf{i-NN2}@\, \mathcal{I}$, with $\mathcal{I} = (\varepsilon^t, \Delta \varepsilon, \sigma^t, \theta^t)$ and $\mathcal{O}= (\Delta \sigma, \Delta \theta)$.\\

\begin{figure}[h]
	\centering
	\includegraphics[width=0.3\textwidth]{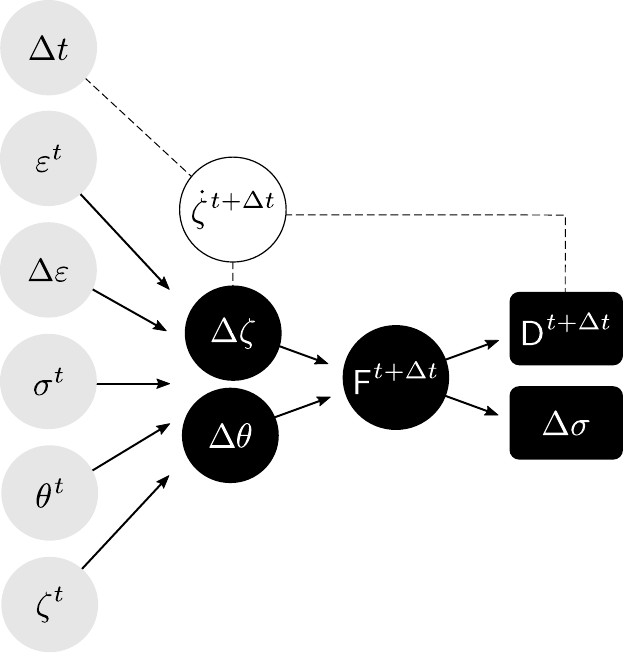}
	\caption{Schematic architecture of \textsf{TANN}. Inputs are highlighted in gray (\tikzcircle[lightgray, fill=lightgray]{3pt}); outputs in black, (\tikzcircle{3pt}) and (\tikzrect{0.1pt}); and intermediate quantities in white (\tikzcircle[black, fill=white]{3pt}). Dashed lines represent definitions, while arrows are used to denote ANN.}
	\label{f:TbNN}
\end{figure}

The main aim of this work is to change the classical paradigm of data-driven ANN material modeling into physics-based ANN material modeling. We propose a new class of ANN based on thermodynamics, which are thermodynamics-based artificial neural networks (TANN). By exploiting the theoretical background presented in Section \ref{sec:theory}, we propose neural networks which, by definition, respect the thermodynamic principles, holding true for any class of material. In this framework, TANN posses the special feature that the entire constitutive response of a material can be derived from definition of only two potential functions: an energy potential and a dissipation (pseudo-) potential \cite{houlsby2007principles}. TANNs are fed with thermodynamics "information" by relying on the automatic differentiation technique \cite{baydin2017automatic} to differentiate neural networks outputs with respect to their inputs. This strategy allows to construct a general framework of neural networks material models which, in principle, can be exploited to predict the behavior of any material and assure that the predictions of TANN will be thermodynamically consistent even for inputs that exceed the training range of data. In this paper, we only focus on strain-rate independent processes. Moreover, our approach can be extended, following the developments in \cite{houlsby2000thermomechanical}, to materials showing viscosity and strain-rate dependency.\\
The model relies on an incremental formulation and can be used in existing Finite Element formulations (among others), see e.g. \cite{lefik2003artificial}. Figure \ref{f:TbNN} illustrates the scheme of \textsf{TANN}. The model inputs are the strain increment, the previous material state at time $t$, which is identified herein through the material stress, $\sigma^t$, temperature, $\theta^t$, and the internal state variables, $\zeta_i^t$, as well as the time increment $\Delta t$, namely $\mathcal{I} = (\varepsilon^t, \Delta\varepsilon, \sigma^t, \theta^t, \zeta_i^t, \Delta t)$. The \textit{primary} outputs, $\mathcal{O}_1$, are internal variables increment, $\Delta \zeta_i$, the temperature increment, $\Delta \theta$, and the energy potential at time $t+\Delta t$, $\textsf{F}^{t+\Delta t}$, i.e. $\mathcal{O}_1=(\Delta \zeta_i, \Delta \theta, \textsf{F}^{t+\Delta t})$. \textit{Secondary} outputs, $\mathcal{O}_2-$that is, outputs computed by differentiation of the neural network with respect to the inputs$-$are the stress increment, $\Delta \sigma$, and the dissipation rate, $\textsf{D}^{t+\Delta t}$, which we denote as $\mathcal{O}_2 =\nabla_{\mathcal{I}}\mathcal{O}_1 = (\Delta \sigma, \textsf{D}^{t+\Delta t})$.\\
The class of neural network we propose differs from the previous ones by the fact that the quantity of main interest, i.e., the stress increment, is obtained as a derived one, which intrinsically satisfies the first principle of thermodynamics (and, as we shall see, the second principle, as well). In the following, we briefly recall the basic concepts of artificial neural networks (paragraph \ref{par:ann_overview}), we then focus on the issue of the second-order vanishing gradients that may afflict the training and the performance of an ANN model (paragraph \ref{par:vanishing_gradient}). In particular, it is shown that, in the framework of thermodynamics-based artificial neural networks, particular attention has to be paid to the selection of activation functions. Finally, we present in detail the architecture of our model (paragraph \ref{par:tann}).

\subsection{Artificial neural networks overview}
\label{par:ann_overview}
\begin{figure}[h]
	\centering
	\includegraphics[width=0.25\textwidth]{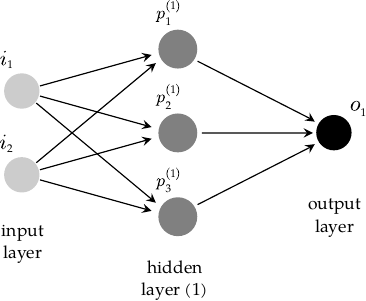}
	\caption{Graph illustration of an ANN structure with two inputs, one output, and one hidden layer with three nodes.}
	\label{f:NN_theory}
\end{figure}

We give herein a brief overview of the basic concepts of artificial neural networks (ANNs). For more details, we refer to \cite{hu2002handbook, geron2019hands}. ANNs can be regarded as non-linear operators, composed of an assembly of mutually connected processing units$-$nodes$-$, which take an input signal $\mathcal{I}$ and return the output $\mathcal{O}$, namely
\begin{equation}
\mathcal{O} = \textsf{ANN}@\mathcal{I}.
\end{equation}
ANN consist of at least three types of layers: input, output and hidden layers, with equal or different number of nodes. Figure \ref{f:NN_theory} depicts a network composed  of one hidden layer, with 3 nodes, an input layer with 2 inputs, and an output layer with 1 node. When an ANN has two or more hidden layers, it is called a deep neural network \cite{geron2019hands}. Denoting the input array with $\mathcal{I}=\left(i_t\right)$, with $t=1,2 \ldots, n_{\mathcal{I}}$ ($n_{\mathcal{I}}$ is the number of inputs), and the outputs with $\mathcal{O}=\left(o_j\right)$, with $j=1,2 \ldots, n_{\mathcal{O}}$ ($n_{\mathcal{O}}$ is the number of outputs), the signal flows from layer $(l-1)$ to layer $(l)$ according to
\begin{equation}
p^{(l)}_k = \mathcal{A}^{(l)}\left(z_k^{(l)} \right), \quad \text{with} \quad z_k^{(l)} = \sum_s^{n^{(l-1)}_{\mathcal{N}}} \left( w^{(l)}_{ks} p^{(l-1)}_s \right) + b^{(l)}_k,
\end{equation}
where $p^{(l)}_k $ is the output of node $k$, at layer $(l)$; $\mathcal{A}^{(l)}$ is the activation function of layer $(l)$; $n^{(l-1)}_{\mathcal{N}}$ is the number of neurons in layer $(l-1)$; $w^{(l)}_{ks}$ is the \textit{weight} between the $s$-th node in layer $(l-1)$ and the $k$-th node in
layer $(l)$; and $b^{(l)}_k$ are the \textit{biases} of layer $(l)$.
With reference to Figure \ref{f:NN_theory}, the output is given by
\begin{eqnarray*}
\mathcal{O} =& \mathcal{A}^{(\text{o})}\left(z^{(\text{o})}\right)\quad \text{with} \quad z^{(\text{o})} =\sum_r w_{r}^{(2)} p_r^{(1)} +b^{(2)} \\
p_r^{(1)} =& \mathcal{A}^{(1)}\left( z^{(1)}_r\right) \quad \text{with} \quad z^{(1)}_r =\sum_t w_{rt}^{(1)} i_t +b_r^{(1)},
\end{eqnarray*}
where the activation function of the output layer, $\mathcal{A}^{(\text{out})}$, in a regression problem, is a linear function, in the most part of applications. The weights and biases of interconnections are adjusted, in an iterative procedure (gradient descent algorithm \cite{geron2019hands}), to minimize the error between the benchmark, $\overbar{\mathcal{O}}$, and prediction, $\mathcal{O}$, that is measured by a loss function, $\mathcal{L}$. In the following, the Mean (over a set of $N$ samples) Absolute Error is used as loss function, i.e.,
\begin{equation}
\mathcal{L} = \dfrac{\sum_{i=1}^{N} |\overbar{\mathcal{O}}_i - \mathcal{O}_i|}{N},
\end{equation}
where $i=1,2,\ldots N$. The errors related to each node of the output layer are hence back-propagated to the nodes in the hidden layers and used to calculate the gradient of the loss function, namely
\begin{equation}
\dfrac{\partial \mathcal{L}}{\partial w_{ks}^{(l)}} = \dfrac{\partial \mathcal{L}}{\partial o_j} \dfrac{\partial o_j}{\partial z_r^{(l+m)}}\dfrac{\partial z_r^{(l+m)}}{\partial p_r^{(l+m -1)}} \cdots\dfrac{\partial z_k^{(l+1)}}{\partial p_k^{(l)}} \dfrac{\partial p_k^{(l)}}{\partial w_{ks}^{(l)}},
\end{equation}
which is also used to update weights and biases, and force the minimization of the loss function values, i.e.
\begin{equation}
\label{eq:update}
w_{ks}^{(l)-{\text{new}}} := w_{ks}^{(l)} - \epsilon \dfrac{\partial \mathcal{L}}{\partial w_{ks}^{(l)}},
\end{equation} 
where $\epsilon$ is the so-called learning rate.
The weights and biases updating, the so-called training process, is performed on a subset of the input-output data-set, defined as training set, known from experimental tests or numerical simulations of the phenomenon investigated. The ANN is trained. The training process is stopped as the loss function is below a specific tolerance. Then a test set, a subset of the input-output data-set different to the training set, is used to check the error of the network predictions. Once the ANN is trained, it is used in recall mode to obtain the output of the problem at hand.\\
Due to their rich interpolation space, ANNs have proved to be universal approximators, see e.g. \cite{chen1995universal,cybenko1989approximation}, although the choice of hyper-parameters, such as the number of neurons, the network topology, the weights, etc. are problem-dependent. The same stands for the activation functions, which may be chosen to have some desirable properties of non-linearity, differentiation, monotonicity, etc. Most of these properties mostly stem from issues related to the gradient descent algorithm and the so-called (first-order) vanishing gradient problem. As it follows, we briefly present this well-known issue and we further give insights in a variation of it: the second-order vanishing gradient.

\subsection{First- and second-order vanishing gradients}
\label{par:vanishing_gradient}
During the training process, if the gradient of the loss function with respect to a certain weight tends to zero$-$that is, see Eq. (\ref{eq:update}), when $\mathcal{A'}^{(l)} = \partial p_j^{(l)} / \partial z_j^{(l)} \approx 0$ (with $\mathcal{A}'$ the first-derivative of the activation function with respect to its arguments)$-$the update operation can fail, and the weight and biases values are not updated. In this case, we have the so-called \textit{first-order} vanishing gradient \cite{geron2019hands}. 
\begin{figure*}[h]
	\centering
	\includegraphics[width=0.85\textwidth]{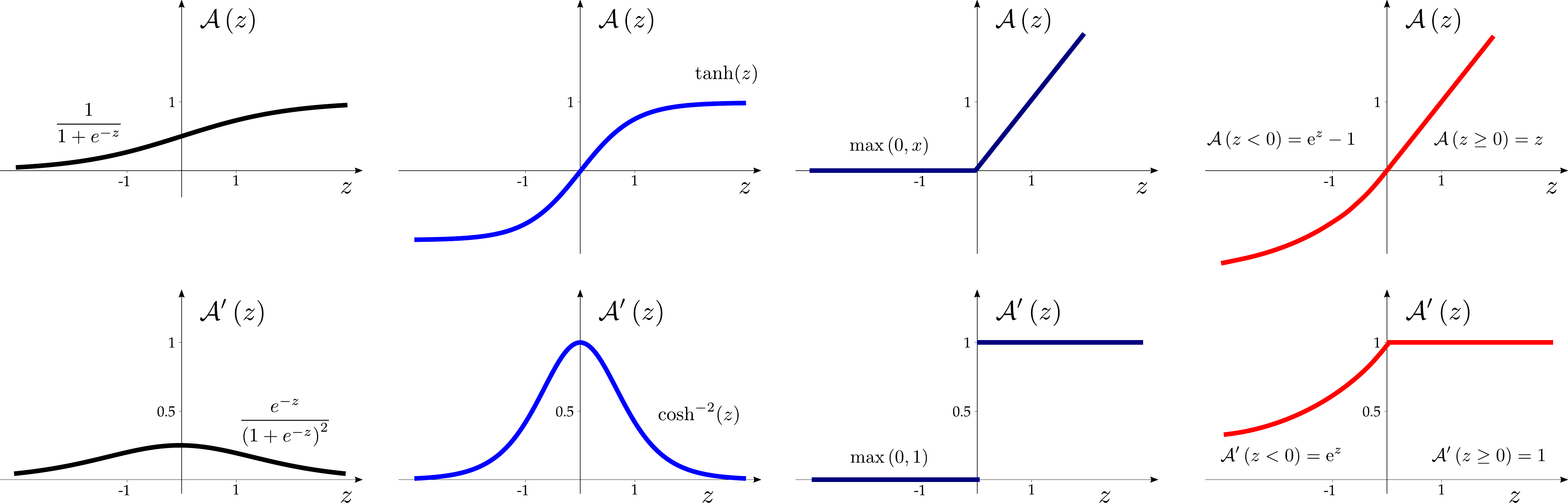}
	\caption{Some of the most common activation functions and their first-order gradient. From left to right: the logistic (sigmoid) function, the hyperbolic tangent, the Rectified Linear Unit (ReLU), and the Exponential Linear Unit (ELU).}
	\label{f:activation}
\end{figure*}
Figure \ref{f:activation} displays some of the most common activation functions and their derivatives$-$that is, the logistic (sigmoid) function, the hyperbolic tangent, the Rectified Linear Unit (ReLU), and the Exponential Linear Unit (ELU). The sigmoid function is S-shaped, continuous, differentiable, its output values range from 0 to 1, and its first-order gradient (derivative) assumes values much smaller than 1. When inputs become large (negative or positive), the function saturates at 0 or 1, with a derivative extremely close to 0. Thus when backpropagation kicks in, it has virtually no gradient to propagate back through the network, which is problematic for training. The hyperbolic tangent activation function is very similar to the sigmoid, but it is centered at zero allowing to maintain the output values within a normalized range (between -1 and 1). Nevertheless, it suffers from saturated gradients (at $z=0$, for $z<<-1$ and $z>>1$). ReLU is continuous but not differentiable at $z=0$. Nevertheless it is an unsaturated activation function for positive values of $z$ (its gradient has no maximum) and, therefore, it allows to avoid vanishing gradient issues for $z>0$. Nevertheless, it suffers from a problem known as the \textit{dying ReLUs}: during training, some neurons are effectively deactivated, meaning they stop outputting anything other than 0 (for $z<0$). To this purpose many variants exist. The ELU activation, for instance, takes on negative values when $z < 0$, which allows the unit to have an average output closer to 0. This helps alleviate the vanishing gradient problem, as discussed earlier. Second, it has a nonzero gradient for $z < 0$, which avoids the dying units issue. Finally, the function is smooth everywhere, including z = 0, which helps speed up gradient descent.\\

\begin{figure}[h]
	\centering
	\includegraphics[width=0.2\textwidth]{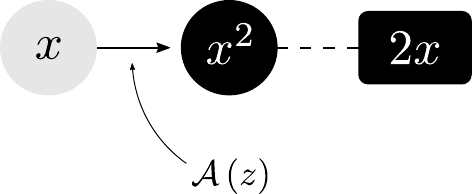}
	\caption{ANN which takes as input $x$ and returns (a) $\mathcal{O}= x^2$ and (b) its derivative with respect to the input, i.e., $\nabla_{\mathcal{I}} \mathcal{O} = 2x$, with one hidden layer whose activation function is $\mathcal{A}$.}
	\label{f:x_square}
\end{figure}

When dealing with \textsf{TANN}, \textit{second-order} vanishing gradients can appear. This is a new concept and, in order to illustrate it, we will use a simle example. Assume an ANN which takes as input some $\mathcal{I}=x$ and returns (a) $\mathcal{O}_1=x^2$ and (b) its derivative with respect to the input, i.e., $\mathcal{O}_2 = \nabla_{\mathcal{I}} \mathcal{O}_1 = 2x$ (see Figure \ref{f:x_square}). Let consider one hidden layer, with activation function $\mathcal{A}$ and $N_n$ nodes. The activation function of the single output layer, which returns $x^2$, is assumed to be linear. In this case, the output (a) is given by 
\begin{equation}
\begin{split}
\mathcal{O}_1=& \:p^{(\text{o})} =\mathcal{A}^{(\text{o})}\left( z_k^{(\text{o})}\right) \\
\mathcal{O}_1=&\:\sum_j w_j^{(\text{o})} p_j^{(1)} + b^{(\text{o})}\\
\mathcal{O}_1=&\:\sum_j w_j^{(\text{o})} \mathcal{A}\left( w_j^{(1)} i +b_j^{(1)}\right) + b^{(\text{o})}.
\end{split}
\end{equation}
The derivatives of the outputs with respect to the inputs can be easily computed, in this simple example, by taking advantage of the automatic (numerical) differentiation \cite{baydin2017automatic}. Output (b) is hence computed by the ANN as
\begin{equation}
\begin{split}
\mathcal{O}_2 = \nabla_{\mathcal{I}} \mathcal{O}_1 = \frac{\partial \mathcal{O}_1}{\partial \mathcal{I}} = &\sum_j \dfrac{\partial p^{(\text{o})}}{\partial z^{(\text{o})}} \dfrac{\partial z^{(\text{o})}}{\partial p_j^{(1)}} \dfrac{\partial p_j^{(1)}}{\partial z_j^{(1)}} \dfrac{\partial z_j^{(1)}}{\partial \mathcal{I}}  \\
\frac{\partial \mathcal{O}_1}{\partial \mathcal{I}} =& \sum_j w_j^{(\text{o})} w_j^{(1)} \mathcal{A}'\left(z^{(1)}_j\right).
\end{split}
\end{equation}

Consider the following loss function
\begin{equation*}
\mathcal{L} = \text{w}_o \mathcal{L}_o + \text{w}_{\nabla_{\mathcal{I}} \mathcal{O}} \mathcal{L}_{\nabla_{\mathcal{I}} \mathcal{O}},
\end{equation*}
where $\mathcal{L}_o$ and $\mathcal{L}_{\nabla_{\mathcal{I}} \mathcal{O}}$ are the loss functions corresponding to output $\mathcal{O}$ and $\nabla_{\mathcal{I}} \mathcal{O}$, respectively. Regularized weights, $\text{w}_o$ and $\text{w}_{\nabla_{\mathcal{I}}}$, can be used to obtain comparable order of magnitude of the two loss functions. During training, weights and biases are updated according to Eq. (\ref{eq:update}) where the computed gradients are
\begin{subequations}
	\begin{align}
	\label{eq:update_x2_a}
	\dfrac{\partial \mathcal{L}}{\partial w_j^{(\text{o})}} = & \:\mathcal{A} 	\mathcal{L}'_o  + w_j^{(1)} \mathcal{A}' \mathcal{L}'_{\nabla_{\mathcal{I}} \mathcal{O}}\\
	\label{eq:update_x2_b}
	\dfrac{\partial \mathcal{L}}{\partial w_j^{(\text{1})}} = & \: i w_j^{(\text{o})} \mathcal{A}'  \mathcal{L}'_o + \left(w_j^{(\text{o})} \mathcal{A}' + i w_j^{(1)} \mathcal{A}''\right)\mathcal{L}'_{\nabla_{\mathcal{I}} \mathcal{O}}\\
	\label{eq:update_x2_c}
	\dfrac{\partial \mathcal{L}}{\partial b^{(\text{o})}} = & \:\mathcal{L}'_o \\
	\label{eq:update_x2_d}
	\dfrac{\partial \mathcal{L}}{\partial b_j^{(1)}} = & \:w_j^{(\text{o})}  \mathcal{A}' \mathcal{L}'_o+ w_j^{(\text{o})} w_j^{(1)} \mathcal{A}'' \mathcal{L}'_{\nabla_{\mathcal{I}} \mathcal{O}}.
\end{align}
\end{subequations}
It follows, from relations (\ref{eq:update_x2_b}) and (\ref{eq:update_x2_d}), that the gradient descent algorithm needs the computation of both first- and second-order gradients of the activation function $\mathcal{A}$. This particular result is a direct consequence of the minimization of the error between the gradient of the outputs with respect to the inputs, i.e. $\mathcal{O}_2 = \nabla_{\mathcal{I}} \mathcal{O}_1$, and the corresponding  benchmark values, $2x$. This is what we call  second-order vanishing gradient problem. It is tantamount to the first-order variant, but it involves the second derivatives (and not only the first) of the activation functions in an ANN. With reference to Figure \ref{f:activation}, none of the depicted, classical activation functions is suitable for such class of problems. Consequently, care must be taken in selecting activation functions that do not have second-order vanishing gradients. 

\subsubsection{Understanding second-order vanishing gradient}
\label{par:second_order_grad}
In the following, we investigate the performance and influence of different activation functions on the computational time to train an ANN with input $\mathcal{I}$, \textit{primary} output $\mathcal{O}_1$, and \textit{secondary} output $\mathcal{O}_2 = \nabla_{\mathcal{I}} \mathcal{O}_1$. Consider the above discussed example with $\mathcal{I}=x$, $\mathcal{O}_1 = x^2$, and $\mathcal{O}_2 = 2x$. The ANN has one hidden layer, with $N_n=6$ nodes, and activation functions as reported in Table ~\ref{tab:activations}. The output layer has linear activation and null bias. The absolute error is selected as loss function for both $\mathcal{O}_1$ and $\mathcal{O}_2$. Training is performed on 1000 samples, normalized between -1 and 1. A very small value for the learning rate is selected, i.e., $\epsilon=10^{-5}$ in order to facilitate the gradient descent algorithm in reaching small values of the loss function. We use \textit{early-stopping}. In other words, training is stopped as the error of a validation set (500 samples) starts to increase while the learning error still decreases \cite{geron2019hands}. The validation set is used to avoid over-fitting of the training data. 
\begin{table}[hbt]
\setlength{\tabcolsep}{.5pc}
\newlength{\digitwidth} \settowidth{\digitwidth}{\rm 0}
\catcode`?=\active \def?{\kern\digitwidth}
\caption{Set of activation functions considered to investigate the performance of the network with outputs $\mathcal{O}= x^2$ and $\nabla_{\mathcal{I}} \mathcal{O}=2x$, with $\mathcal{I}=x$, in the framework of first- and second-order vanishing gradients.}
\label{tab:activations}
\medskip
\centering
\begin{adjustbox}{max width=0.48\textwidth}
\begin{tabular}{@{}l@{\extracolsep{\fill}}ccccc}
\toprule
			Function $\qquad$&  $z$ range & $\mathcal{A}(z)$&  $\mathcal{A}'(z)$ & $\mathcal{A}''(z)$\\
			\midrule
			$\text{ReLU}_z$& $z<0$& $0$ & $0$ & $0$ \\
			& $z\geq 0$& $z$ & $1$ & $0$\\
			\midrule
			$\text{ReLU}_{0.5z^2 + z}$& $z<0$& $0$ & $0$ & $0$\\
			& $z\geq 0$& $0.5z^2 + z$ & $z+1$ & $1$\\
			\midrule
			$\text{ReLU}_{z^2}$& $z<0$& $0$& $0$ & $0$\\
			& $z\geq 0$& $z^2$& $2z$ & $2$\\
			\midrule
			$\text{ELU}_{\text{e}}$ & $\forall z$& $\text{e}^z-1$ & $\text{e}^z$ & $\text{e}^z$\\
			\midrule
			$\text{ELU}_z$ & $z<0$ & $\text{e}^z-1$& $\text{e}^z$ & $\text{e}^z$\\
			& $z\geq 0$& $z$ & $1$ & $0$\\
			\midrule
			$\text{ELU}_{0.5z^2 +z}$& $z<0$& $\text{e}^z-1$ & $\text{e}^z$ & $\text{e}^z$\\
			&  $z\geq 0$& $0.5z^2 +z$ &  $z+1$ & $1$\\
			\midrule
			$\text{ELU}_{z^2}$& $z<0$& $\text{e}^z-1$ & $\text{e}^z$ & $\text{e}^z$\\
			&$z\geq 0$& $z^2$ & $2z$ & $2$\\
			\midrule
			$\text{ELU}_{z^4}$& $z<0$& $\text{e}^z-1$ & $\text{e}^z$ & $\text{e}^z$\\
			&$z\geq 0$& $z^4$ & $4z^3$ & $12z^2$\\
			\midrule
			$\text{ELU}_{z^4+0.5z^2+z}$& $z<0$& $\text{e}^z-1$ & $\text{e}^z$ & $\text{e}^z$\\
			&$z\geq 0$& $z^4+0.5z^2+z$ & $4z^3 + z + 1$ & $12z^2 + 1$\\
			\bottomrule
\end{tabular}
\end{adjustbox}
\end{table}

\begin{figure*}[ht]
\centering
\begin{subfigure}[b]{0.49\textwidth}
  \centering
  \includegraphics[height=0.4\textheight]{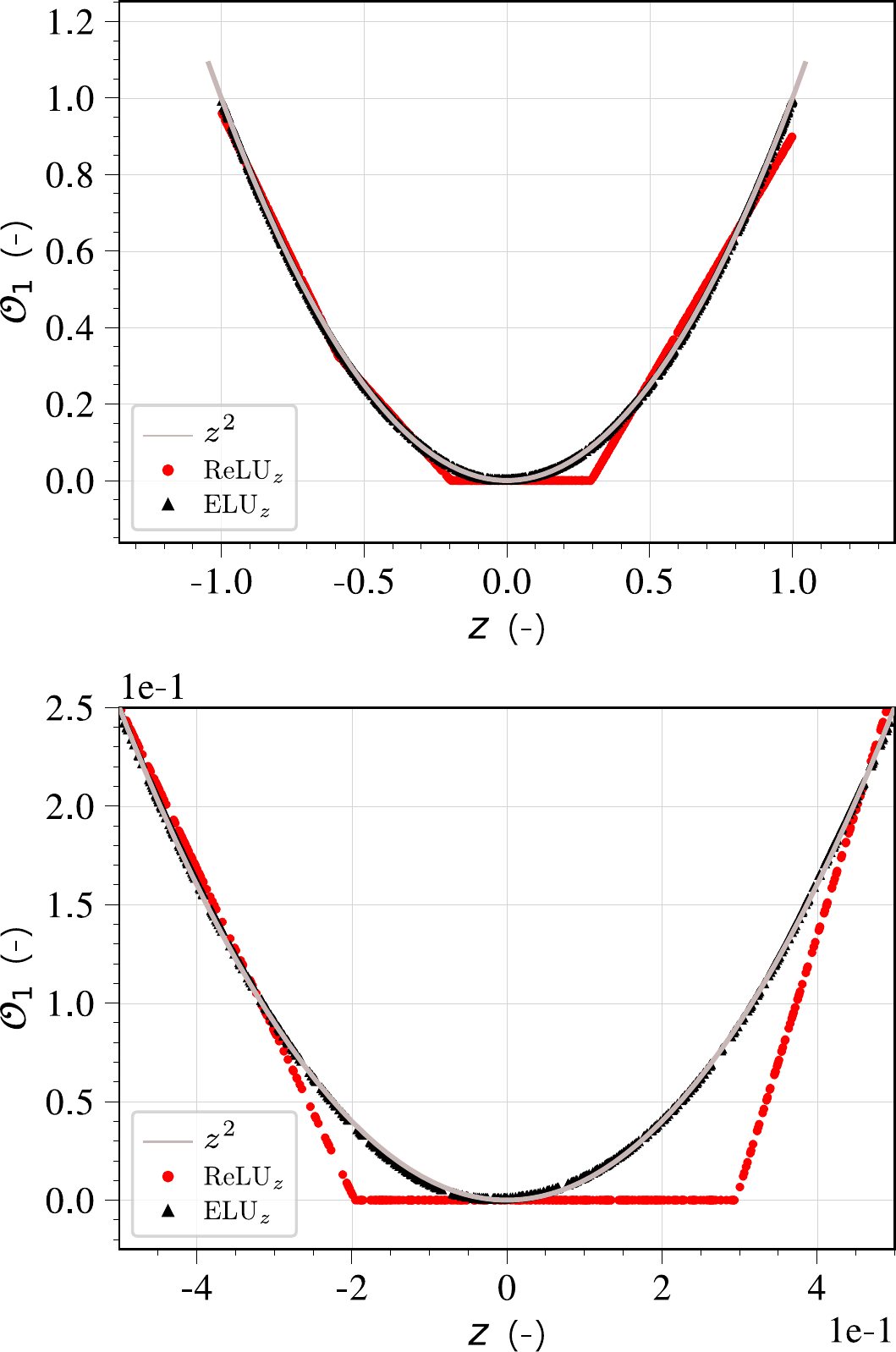}
	\caption{\footnotesize $x^2$ predictions, $\mathcal{O}_1$, using ReLU and $\text{ELU}_z$.}
	\label{f:x2a}
\end{subfigure}
\begin{subfigure}[b]{0.49\textwidth}
  \centering
  \includegraphics[height=0.4\textheight]{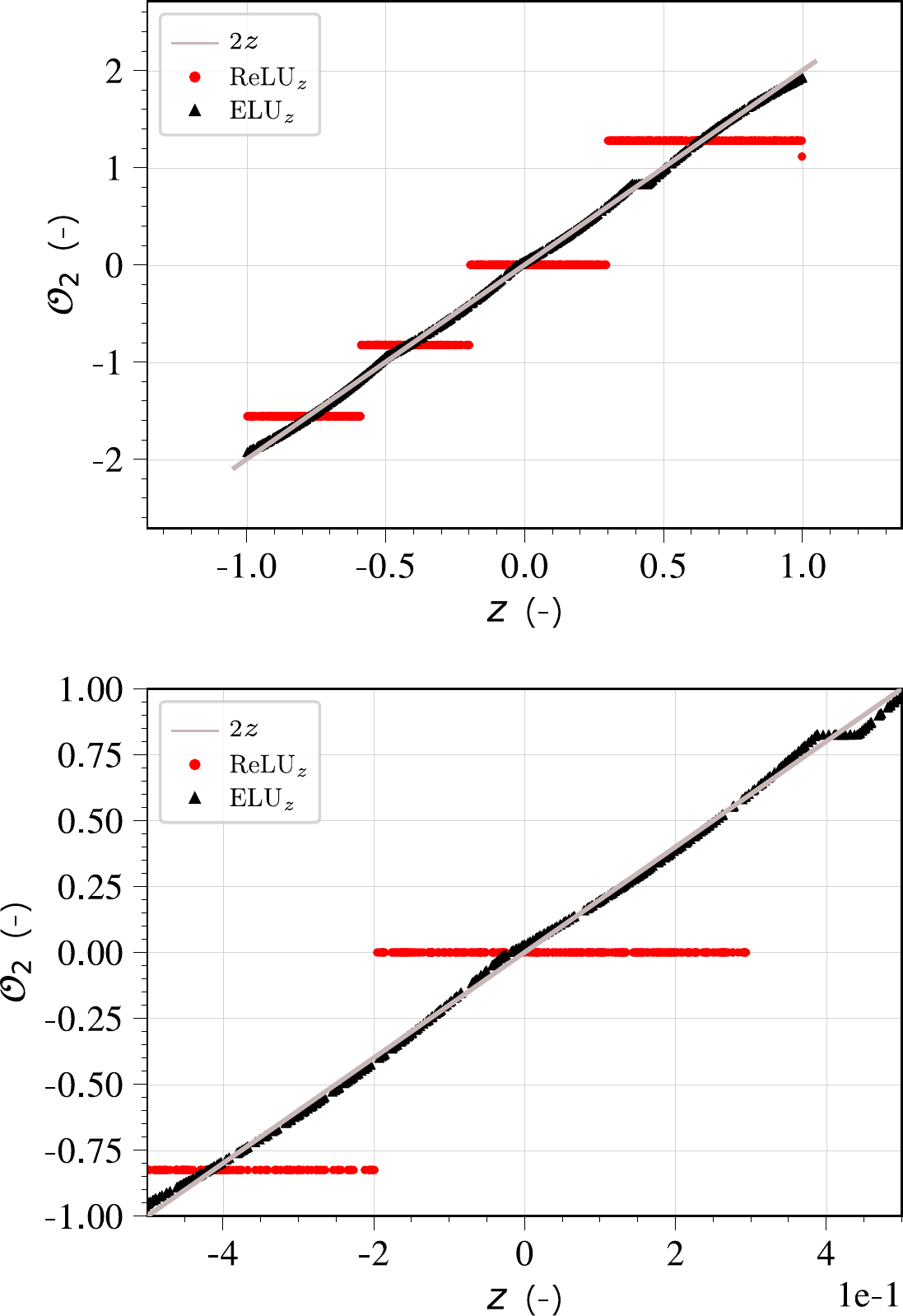}
	\caption{\footnotesize $2x$ predictions, $\mathcal{O}_2$, using ReLU and $\text{ELU}_z$.}
	\label{f:2xa}
\end{subfigure}\\
\vspace{0.5cm}

\begin{subfigure}[b]{0.49\textwidth}
  \centering
  \includegraphics[height=0.4\textheight]{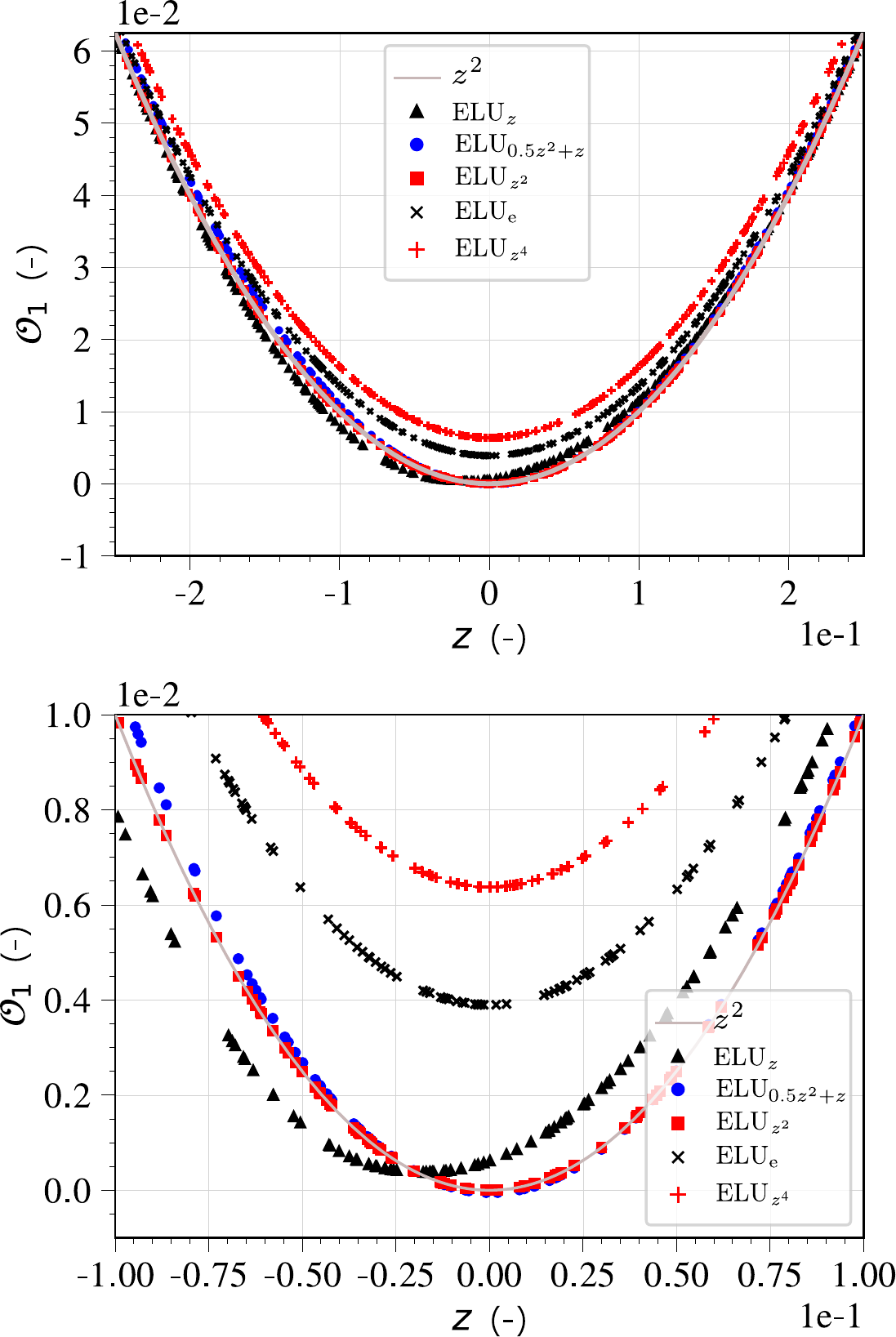}
	\caption{\footnotesize $x^2$ predictions, $\mathcal{O}_1$, using $\text{ELU}_z$, $\text{ELU}_{0.5z^2 + z}$, $\text{ELU}_{z^2}$, $\text{ELU}_{\text{e}}$, and $\text{ELU}_{z^4}$.}
	\label{f:x2b}
\end{subfigure}
\begin{subfigure}[b]{0.49\textwidth}
  \centering
  \includegraphics[height=0.4\textheight]{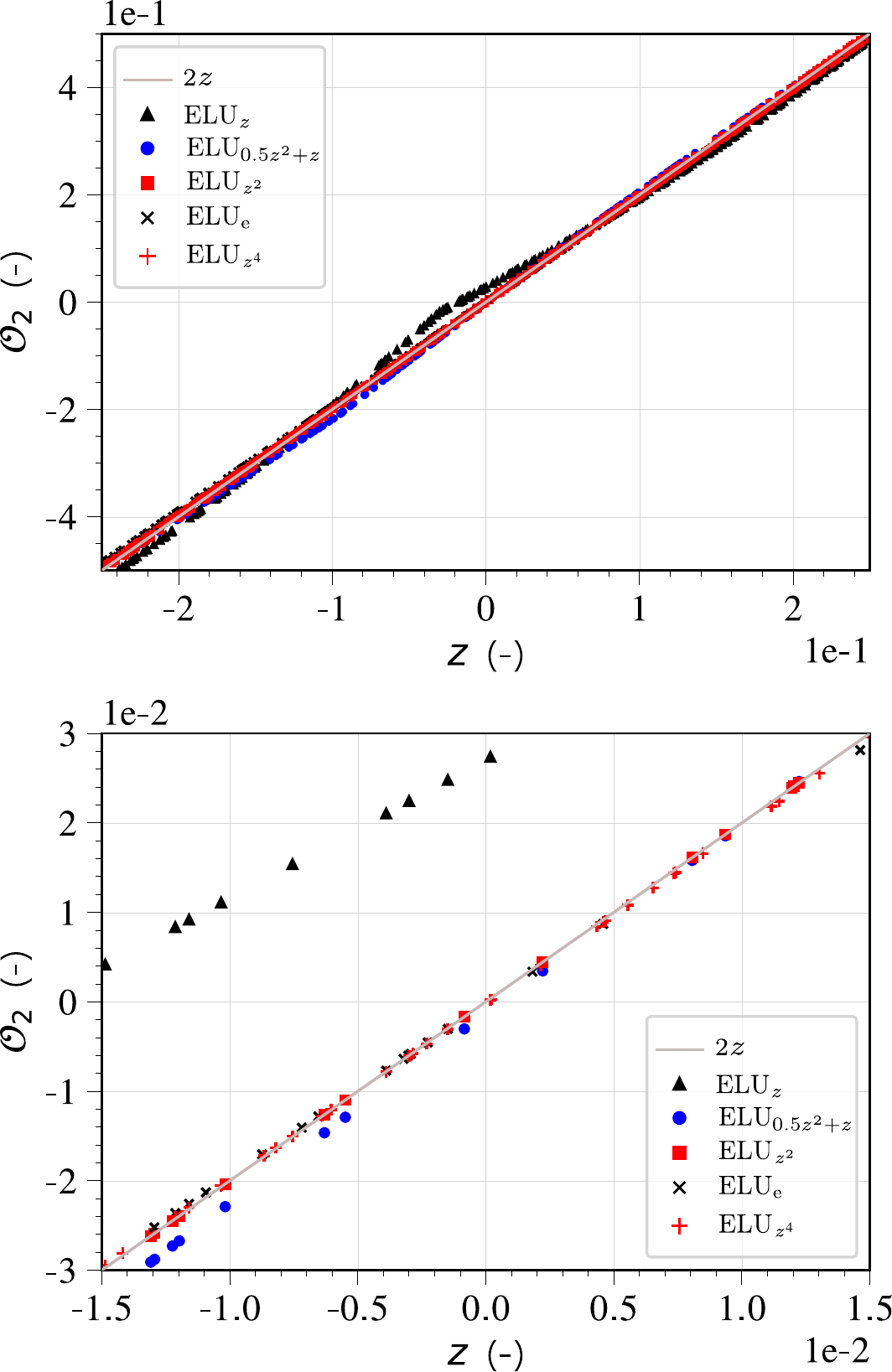}
	\caption{\footnotesize $2x$ predictions, $\mathcal{O}_2$, using $\text{ELU}_z$, $\text{ELU}_{0.5z^2 + z}$, $\text{ELU}_{z^2}$, $\text{ELU}_{\text{e}}$, and $\text{ELU}_{z^4}$.}
	\label{f:2xb}
\end{subfigure}
\caption{Comparison of different activation functions for the prediction of the primary output, $x^2$ (a), and secondary output, $2x$ (b). From top to bottom the range of $z$ decreases from larger to smaller values, to observe the behavior at $z\approx 0$.}
	\label{f:x2_2x}
\end{figure*}

For each tested activation function, Table \ref{tab:activations_comparison} shows the Mean Absolute Error (MAE) calculated using a set of new, unseen  data (500 samples) of input-output predictions for $x^2$ and $2x$. The advancement of training is quantified herein as the number of epochs, i.e., the number with which the training algorithm works with the training data-set \cite{geron2019hands}. Activation functions with quadratic terms, or of higher degree, perform very well, compared to their linear equivalents. RELU$_{z^2}$, ELU$_{z^2}$ outperform as their shape is very similar to the input-output regression they are trained to learn. Nevertheless, it is worth noticing that training fails when activation functions with vanishing second gradient are used (e.g. RELU$_z$ and ELU$_z$). Figure \ref{f:x2_2x} compares the ANN predictions for a selection of activation functions with the analytical (exact) results. Whilst RELU$_z$ is clearly inadequate, ELU$_z$ predictions overall agree with the analytical values. This is due to the fact that the ANN takes advantage of the exponential term, for negative $z$ and thus successfully manage to satisfy both $\mathcal{O}$ and $\nabla_{\mathcal{I}}\mathcal{O}$. Additional hidden layers may improve the performance of the network. It can be further noticed that activation function of high degree, e.g. $\text{ELU}_{\text{e}}$, $\text{ELU}_{z^4}$, and $\text{ELU}_{z^4+0.5z^2+z}$, even if successful, require a large number of epochs.

\begin{table}[hbt]
\caption{Activation functions and performance with unseen data.}
\label{tab:activations_comparison}
\medskip
\centering
\begin{adjustbox}{max width=0.48\textwidth}
\begin{tabular}{@{}l@{\extracolsep{\fill}}cccc}
\toprule
			Activation function $\mathcal{A}$ & $\mathcal{L}$ & $\mathcal{L}_{\mathcal{O}}$ & $\mathcal{L}_{\nabla_{\mathcal{I}} \mathcal{O}}$  & no. epochs\\
			& $(10^{-4})$&$(10^{-4})$ & $(10^{-4}) $ & (-) \\
			\midrule
			$\text{ReLU}_z$ & 1521.2 & 205.98 &1315.18 & 920\\
			$\text{ReLU}_{0.5z^2 + z}$ & 762.4& 93.58 &668.85 & 8054\\
			$\text{ReLU}_{z^2}$ & 0.061 & 0.0241 &0.0371& 148\\
			\midrule
			$\text{ELU}_{\text{e}}$ & 127.2& 26.83 & 100.38 & 19477\\
			$\text{ELU}_z$ & 108.56 & 12.12 & 96.44 & 17280\\
			$\text{ELU}_{0.5z^2 + z}$ & 65.5& 10.91 & 54.63 & 12178 \\
			$\text{ELU}_{z^2}$ & 0.13 & 0.067 & 0.067 & 88\\
			$\text{ELU}_{z^4}$ & 65.36 & 33.75& 31.61 & 20051\\
			$\text{ELU}_{z^4 + 0.5z^2 + z}$ &12.94 & 1.81 &11.13 & 9683\\
			\bottomrule
\end{tabular}
\end{adjustbox}
\end{table} 

\cleardoublepage

\subsection{Architecture of Thermodynamics-based ANN}
\label{par:tann}
\begin{figure*}[ht]
\centering
\begin{subfigure}[b]{0.49\textwidth}
  \centering
  \includegraphics[width=\linewidth]{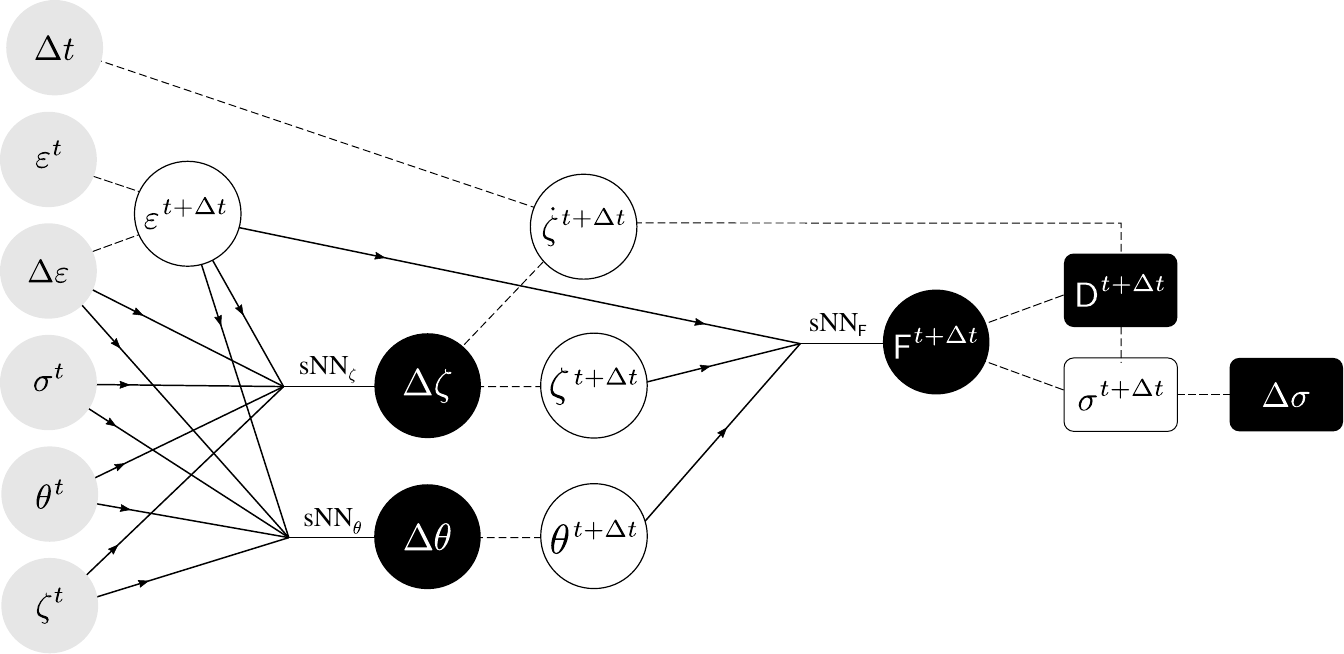}
	\caption{\footnotesize non-isothermal processes.}
	\label{f:full_TbNN_1}
\end{subfigure}
\begin{subfigure}[b]{0.49\textwidth}
  \centering
  \includegraphics[width=\linewidth]{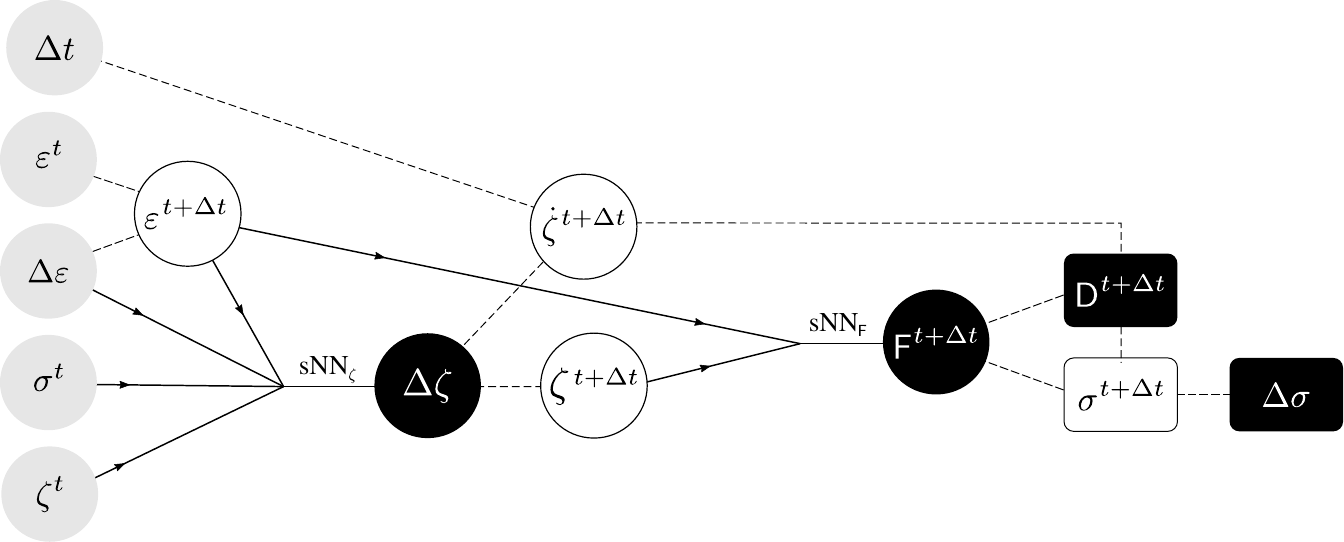}
	\caption{\footnotesize isothermal processes.}
	\label{f:full_TbNN_2}
\end{subfigure}
\caption{Architecture of \textsf{TANN}: general case (a) and for isothermal processes (b). Inputs are highlighted in gray (\tikzcircle[lightgray, fill=lightgray]{3pt}); outputs in black, (\tikzcircle{3pt}) for direct ANN predictions and (\tikzrect{0.1pt}) for derived outputs; and intermediate quantities (definitions) are in white (\tikzcircle[black, fill=white]{3pt}) and  (\tikzemptyrect{0.1pt}). Relationships obtained from definitions are represented with dashed lines, while arrows denote ANNs.}
	\label{f:full_TbNN}
\end{figure*}

Herein we detail the architecture and the internal steps/definitions \textsf{TANN} is relying on. The architecture is detailed in Figure \ref{f:full_TbNN}. The input vector is $\mathcal{I} = (\varepsilon^t, \Delta\varepsilon, \sigma^t, \theta^t, \zeta_i^t, \Delta t)$, the primary and secondary outputs are $\mathcal{O}=(\Delta \zeta_i, \Delta \theta, \textsf{F}^{t+\Delta t})$ and $\nabla_{\mathcal{I}}\mathcal{O} = (\Delta \sigma, \textsf{D}^{t+\Delta t})$, respectively. \textsf{\textsf{TANN}} involves the following steps:
\begin{itemize}[noitemsep,topsep=0pt]
	\item[1.] computation of the updated strain (definition):
	$\varepsilon^{t+\Delta t}:= \varepsilon^t +\Delta \varepsilon$
	\item[2.] prediction of the kinematic variables and temperature increments with two sub-ANNs: 
	\[\Delta \zeta=\textsf{sNN}_{\zeta}@\left(\varepsilon^{t+\Delta t},\Delta \varepsilon^t,\sigma^t,\theta^t,\zeta^t\right)\]
	and
	\[\Delta \theta=\textsf{sNN}_{\theta}@\left(\varepsilon^{t+\Delta t}, \Delta \varepsilon, \sigma^t, \theta^t, \zeta^t\right)\]
	\item[3.] computation of 
	\begin{itemize}[noitemsep,topsep=0pt]
		\item[$(a)$] the updated kinematic variables rates (backward finite difference approximation):
		 
		$\dot{\zeta}^{t+1} \approx \frac{\Delta \zeta}{\Delta t}$
 
		\item[$(b)$] the updated kinematic variables (definition):
		$\zeta^{t+1}:= \zeta^t + \Delta \zeta^t$ 

		\item[$(c)$] the updated temperature (definition):
		$\theta^{t+1}:= \theta^t + \Delta \theta$
 
	\end{itemize}
	
	\item[4.] prediction of the updated energy potential: 
	\[\text{\textsf{F}}^{t+\Delta t} = \textsf{sNN}_{\textsf{F}}@\{\varepsilon^{t+\Delta t} \quad \zeta^{t+\Delta t} \quad \theta^{t+\Delta t}\}\]

	\item[5.] computation of the updated dissipation (definition, Eq. (\ref{eq:energy_expr})):
	$\text{\textsf{D}}^{t+\Delta t}:= - \frac{\partial \text{\textsf{F}}^{t+\Delta t}}{\partial \zeta^{t+\Delta t}} \cdot \dot{\zeta}^{t+\Delta t}$

	\item[6.] computation of
	\begin{itemize}[noitemsep,topsep=0pt]
		\item[$(a)$] the updated stress (definition, Eq. (\ref{eq:energy_expr})):
					$\sigma^{t+\Delta t}:= \frac{\partial \text{\textsf{F}}^{t+\Delta t}}	 {\partial \varepsilon^{t+\Delta t}}$

		\item[$(b)$] the stress increment (definition):
			$\Delta \sigma:= \sigma^{t+\Delta t} -\sigma^t$
	\end{itemize}
\end{itemize}

\textsf{TANN} is thus composed of three sub-ANNs; $\textsf{sNN}_{\zeta}$ predicts the internal variables increment, $\textsf{sNN}_{\theta}$ predicts the temperature increment (note that in case of the isothermal conditions, this component can be removed from the architecture, see Fig. \ref{f:full_TbNN_2}), and $\textsf{sNN}_{\textsf{F}}$ predicts the Helmholtz free-energy. The main output, the increment in stress, is computed according to expression (\ref{eq:energy_expr}), which stems from thermodynamic requirements. By virtue of the fact that the entire constitutive response of a material can be derived from definition of only two potential functions, the model is able to predict the stress increment from the knowledge of the energy potential (and the internal variables $\zeta_i$). It is worth noticing that, differently from common approaches (cf. Sect. \ref{sec:ann}), the sub-network $\textsf{sNN}_{\textsf{F}}$ is required to learn a scalar quantity$-$that is, the Helmholtz free-energy potential. This offer compelling advantages. When dealing with ANNs, the curse of dimensions (increasing effort in training and large amount of training data required) is an important issue when the studied problem passes to higher dimensions, see e.g. \cite{bessa2017framework}. Passing from 1D to 3D, for instance, increases the number of variables the ANN need to learn. For stresses, from one single scalar value, in 1D, we pass to a vector with six-components, in 3D. The computational effort is thus not trivial. Nevertheless, \textsf{TANN} is, in principle, less affected by these issues as the two potentials, on which the entire set of predictions relies on, are scalar functions.\\
The computation of dissipation, from expression (\ref{eq:energy_expr}), plays a double role. First, it assures thermodynamic consistency of the predictions of \textsf{TANN} (first law). Second, it brings the information to distinguish between reversible and irreversible processes, e.g. elasticity from plasticity/damage, etc., and it is trained to be positive or zero (second law).\\
It is worth noticing that further improvements of the performance of \textsf{TANN}s may be obtained, as suggested in the work of Karpatne et al. \cite{karpatne2017physics}, by adding a physical inconsistency term to the loss functions (e.g., with respect to dissipation).

\section{Generation of data}
\label{sec:generation}
We present the procedure used to generate material data \textsf{TANN} is trained with. Herein, data are obtained by numerical integration of an incremental form of the constitutive relations. To this purpose, we assume the Ziegler's orthogonality condition (see paragraph \ref{par:hyperplasticity} and \cite{ziegler2012introduction, houlsby2000thermomechanical, houlsby2007principles}), which, in general, it is not a strict requirement.  Nevertheless, it is worth noticing that this restriction applies only on the generated data, and not on the ANN class here proposed. More precisely, \textsf{TANN} architecture still holds even for materials for which the Ziegler's normality condition does not apply. We shall recall that the aim is to demonstrate the advantages of thermodynamics-based neural networks with respect to classical approaches. Hence the restrictions, imposed by the orthogonality hypothesis for the generation of data, are expected not to affect the comparisons presented in Section \ref{sec:application}.
\subsection{Incremental formulation}
\label{par:hyperplasticity}
Following the hyperplasticity framework proposed in \cite{einav2007coupled}, the thermo-mechanical, non-linear, incremental constitutive relation for strain-rate independent materials, undergoing infinitesimal strains, is here derived in the framework of isothermal processes ($\theta = \text{cost}$). By differentiating the energy expressions (\ref{eq:energy_expr}) and rearranging the terms, we obtain the following non-linear incremental relations
\begin{subequations}
	\begin{align}
	\label{eq:incr_sg}
	\dot{\sigma} = \partial_{\varepsilon \varepsilon}\text{\textsf{F}} \cdot \varepsilon + \sum_k \partial_{\varepsilon \zeta_k}\text{\textsf{F}}\cdot \dot{\zeta}_k,\\
	\label{eq:incr_chi}
	-\dot{\chi}_i = \partial_{\zeta_i \varepsilon}\text{\textsf{F}} \cdot \varepsilon + \sum_k \partial_{\zeta_i \zeta_k}\text{\textsf{F}}\cdot \dot{\zeta}_k.
	\end{align}
\end{subequations}
where the following notation is adopted
\begin{equation*}
\partial_{\varepsilon \varepsilon}\text{\textsf{F}} = \dfrac{\partial^2 \text{\textsf{F}}}{\partial \varepsilon_{ij} \partial \varepsilon_{kl}},\quad
\partial_{\varepsilon \zeta_k}\text{\textsf{F}} = \dfrac{\partial^2 \text{\textsf{F}}}{\partial \varepsilon_{ij} \partial \zeta_k}, \quad \partial_{\zeta_i \zeta_k}\text{\textsf{F}} = \dfrac{\partial^2 \text{\textsf{F}}}{\partial \zeta_{i} \partial \zeta_k}.
\end{equation*}
Further, introducing the thermodynamic dissipative stresses $\mathcal{X}^{\dagger}= (X_1, \ldots, X_N)$ and assuming the Ziegler's orthogonality condition (\cite{ziegler2012introduction}), the following non-linear, incremental constitutive relation can be found
\begin{equation}
\dot{\Xi} = \begin{cases} 
      \mathcal{M}|_{y=0} \;\cdot \dot{\varepsilon}  & \text{if } y = 0 \\
      \mathcal{M}|_{y<0} \;\cdot \dot{\varepsilon}  & \text{else }
   \end{cases}
\label{eq:incremental}
\end{equation}
with
\begin{equation}
\dot{\Xi}=
\begin{bmatrix} 
\dot{\sigma}\\
-\dot{X}_i\\
\dot{\zeta_i}\\
\lambda
\end{bmatrix},
\quad
\mathcal{M}|_{y=0}
= 
\begin{bmatrix} 
\partial_{\varepsilon \varepsilon}\text{\textsf{F}} - \sum_k \partial_{\varepsilon \zeta_k}\text{\textsf{F}} \cdot \left(\frac{\mathcal{C}_{\varepsilon}}{B}\cdot\frac{\partial y}{\partial X_k}\right)\\
\partial_{\zeta_i \varepsilon}\text{\textsf{F}} - \sum_k \partial_{\zeta_i \zeta_k}\text{\textsf{F}} \cdot \left( \frac{\mathcal{C}_{\varepsilon}}{B} \cdot\frac{\partial y}{\partial X_k}\right)\\
-\dfrac{\mathcal{C}_{\varepsilon}}{B}\cdot\frac{\partial y}{\partial X_i}\\
-\dfrac{\mathcal{C}_{\varepsilon}}{B}
\end{bmatrix},
\quad \text{and} \quad
\mathcal{M}|_{y<0}
= 
\begin{bmatrix} 
\partial_{\varepsilon \varepsilon}\text{\textsf{F}}\\
\partial_{\zeta_i \varepsilon}\text{\textsf{F}} \\
\emptyset\\
\emptyset
\end{bmatrix},
\label{eq:elasticity_iso}
\end{equation}
and $\cdot$ denotes the contraction of adjacent indices.
In the above relations (\ref{eq:incremental}-\ref{eq:elasticity_iso}), whose derivation is presented in Appendix A, $y=\tilde{y}(\varepsilon, \mathcal{Z}, \mathcal{X}^{\dagger})$ is the yield function, $\emptyset$ denotes a quantity (scalar or tensorial, depending on the dimensionality of the internal variable set) equal to zero,
\begin{equation*}
\mathcal{C}_{\varepsilon} = \frac{\partial y}{\partial \varepsilon} - \sum_{i=1}^N \frac{\partial y}{\partial X_i} \cdot \partial_{\zeta_i \varepsilon}\text{\textsf{F}},
\end{equation*}
\begin{equation*}
B = \sum_{i=1}^N  \frac{\partial y}{\partial \zeta_i} \cdot \frac{\partial y}{\partial X_i} - \sum_{i=1}^N  \frac{\partial y}{\partial X_i} \left( \sum_{k=1}^N  \partial_{\zeta_k \varepsilon}\text{\textsf{F}} \cdot \frac{\partial y}{\partial X_k}\right).
\end{equation*}

\subsection{Data generation}
\label{par:data_generation}
Data are generated in a Python environment \cite{mckinney2011pandas}, where SymPy \cite{meurer2017sympy} and SciPy \cite{virtanen2020scipy} libraries are used for symbolic calculations and numerical integration. Data are generated by identifying an initial state for the material at time $t$,
\[\text{state at time } t: \qquad
\Xi^t = \begin{bmatrix} 
\sigma^t\\
-X_i^t\\
\zeta_i^t\\
0 \end{bmatrix}
\quad \text{and} \quad  \varepsilon^t,\]
and a given strain increment $\dot{\varepsilon}^t$, assuming constant and unitary time increment $\Delta t=1$ ($\dot{\varepsilon}^t = \Delta \varepsilon^t$). Numerical integration of the ordinary differential equations (\ref{eq:incremental}) is performed with an explicit solver \cite{bogacki19893} to obtain the state at the new time $t+\Delta t$, i.e., 
\[\text{state at time } t+\Delta t: \qquad
\Xi^{t+\Delta t} = \begin{bmatrix} 
\sigma^{t+\Delta t}\\
-X_i^{t+\Delta t}\\
\zeta_i^{t+\Delta t}\\
\lambda^{t+\Delta t} \end{bmatrix}
\]
The training data play a crucial role for both the accuracy of the predictions and the generalization with respect to the ANN state variables, e.g., strain increments. The generalization capability of a network is here defined as the ability to make predictions for loading paths different from those used in the training operation. Nevertheless, a significant dependency on the ANN state variables is usually observed. This may result in a poor network generalization. In \citet{lefik2003artificial}, an improvement of the generalization capability of the ANN is proposed. Artificial sub-sets of data, with zero strain increments, are added in the set of training data to force the network in learning that to zero input increments correspond zero output increments.\\
In the available literature, strain-stress loading paths are commonly used in training. If recursive neural networks are used, feeding them with history variables (loading paths) is the only possible solution (see e.g. \cite{mozaffar2019deep}). Nevertheless, ANNs do not necessary need the data-sets to be (historical) paths.

\begin{figure*}[ht]
\centering
\begin{subfigure}[b]{0.3\textwidth}
  \centering
  \includegraphics[width=.95\linewidth]{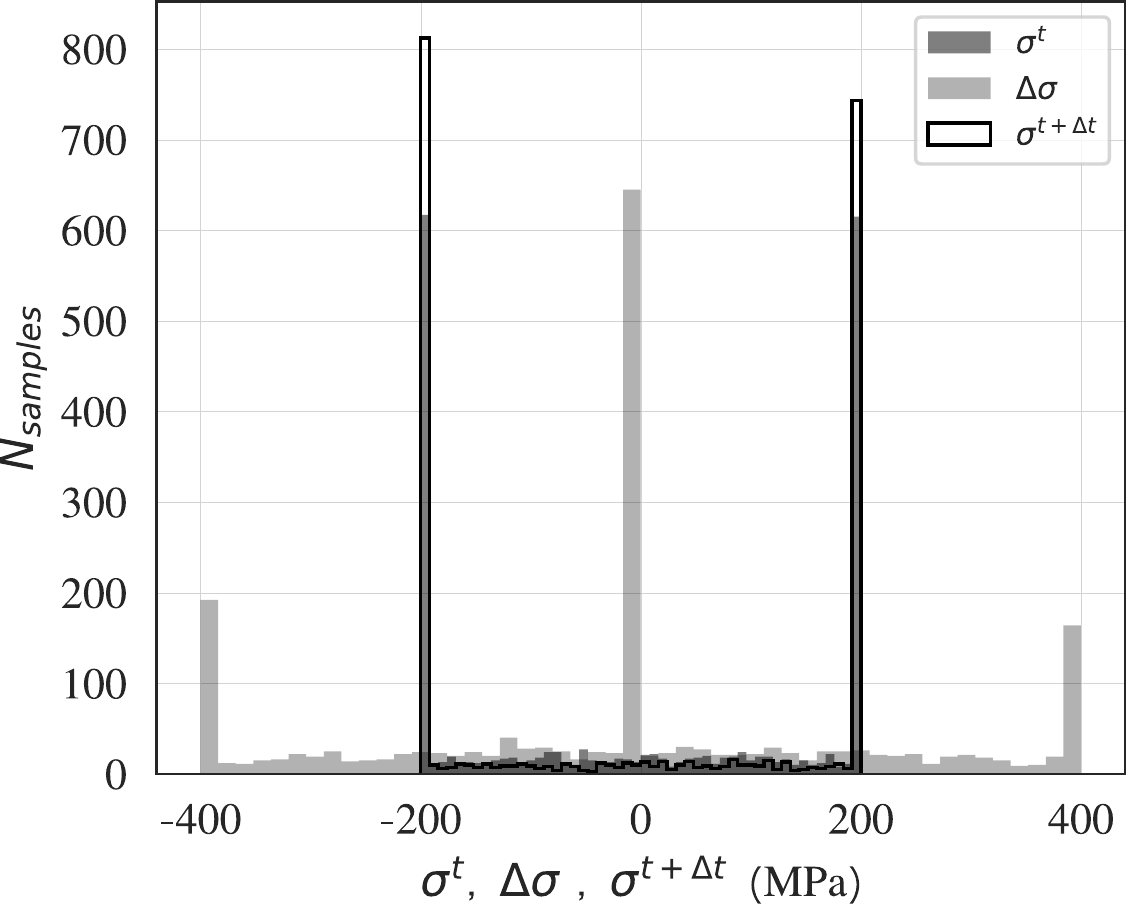}
	\caption{\footnotesize Cauchy stress, $\sigma^t$, $\sigma^{t+\Delta t}$, and increment $\Delta \sigma$.}
\end{subfigure} \hspace{0.05cm}
\begin{subfigure}[b]{0.3\textwidth}
  \centering
  \includegraphics[width=.95\linewidth]{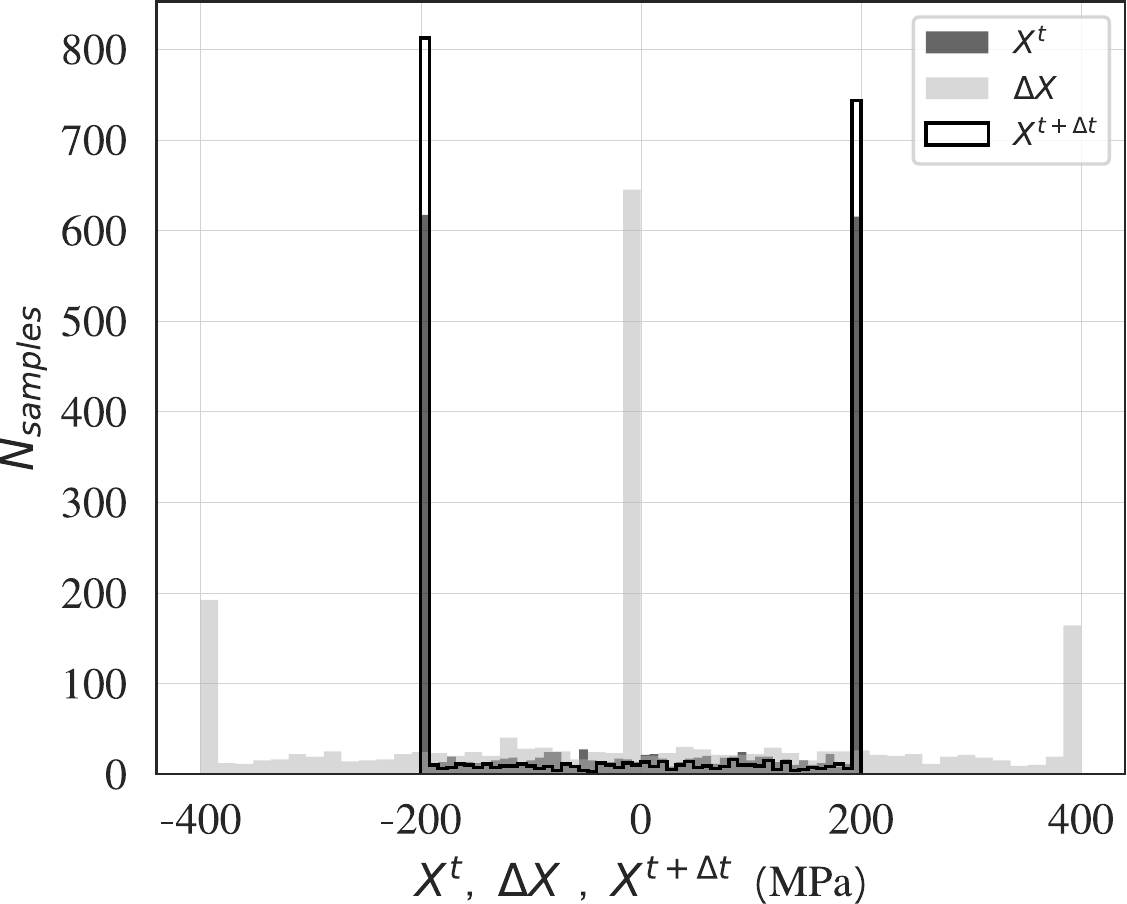}
	\caption{\footnotesize thermodynamic stress, $X^t$, $X^{t+\Delta t}$, and increment $\Delta X$.}
\end{subfigure} \hspace{0.05cm}
\begin{subfigure}[b]{0.3\textwidth}
  \centering
  \includegraphics[width=.95\linewidth]{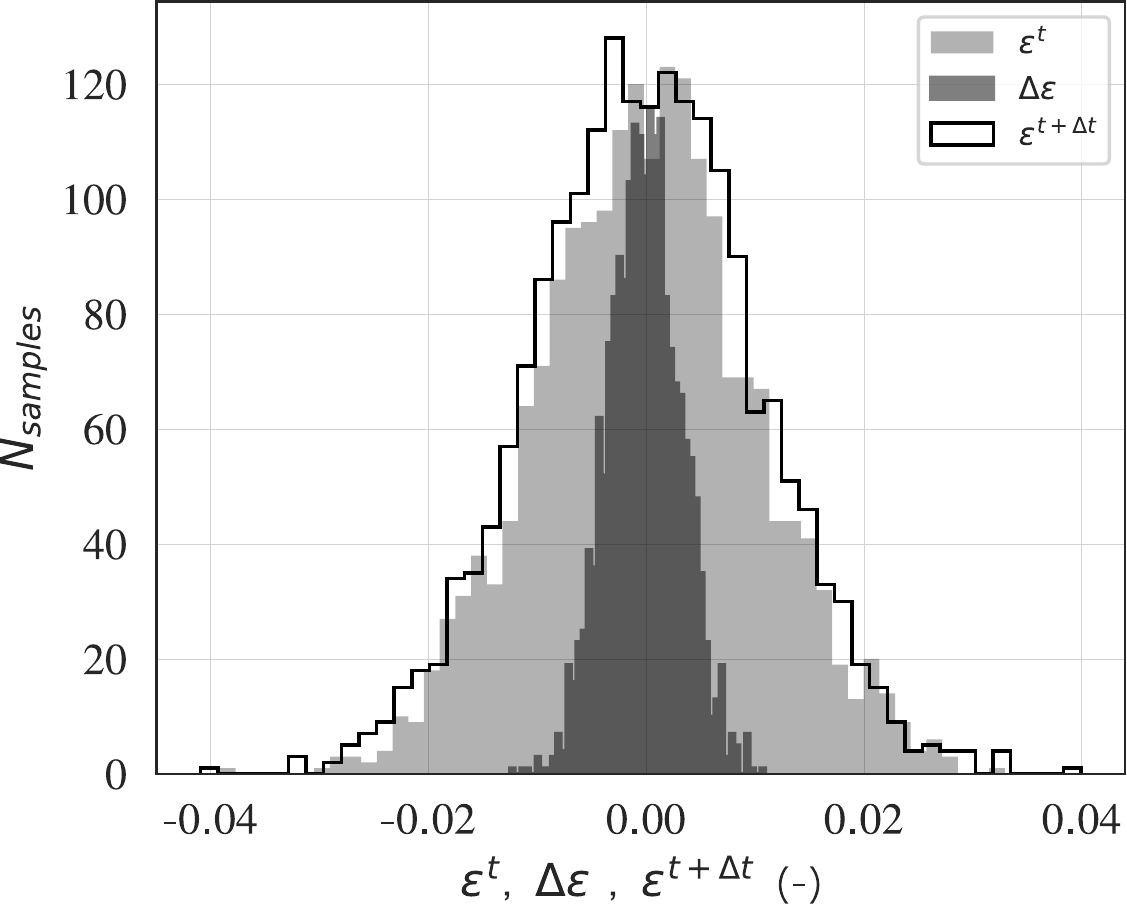}
	\caption{\footnotesize total strain, $\varepsilon^t$, $\varepsilon^{t+\Delta t}$, and increment $\Delta \varepsilon$.}
\end{subfigure}
\begin{subfigure}[t]{0.3\textwidth}
  \centering
  \includegraphics[width=.95\linewidth]{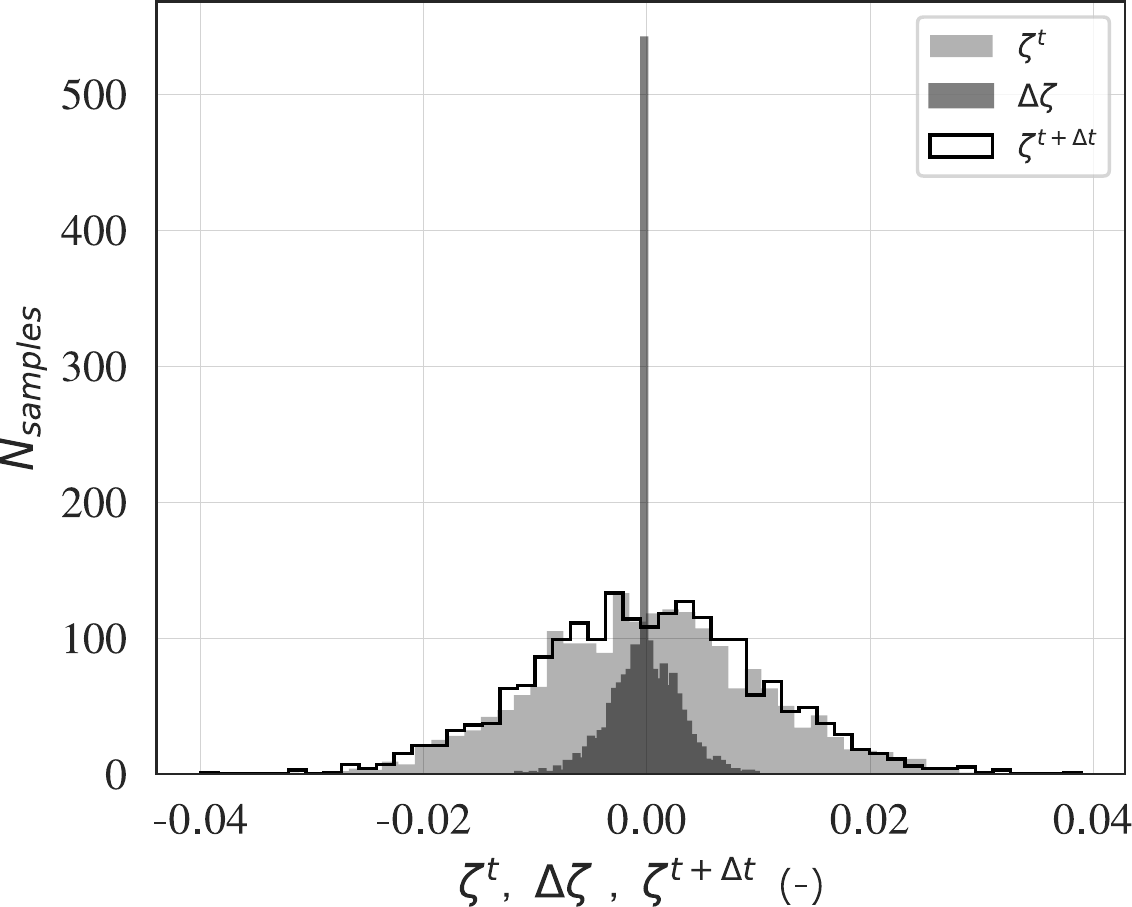}
	\caption{\footnotesize internal variable (inelastic strain), $\zeta^t$, $\zeta^{t+\Delta t}$, and increment $\Delta \zeta$.}
\end{subfigure}\hspace{0.05cm}
\begin{subfigure}[t]{0.3\textwidth}
  \centering
  \includegraphics[width=.95\linewidth]{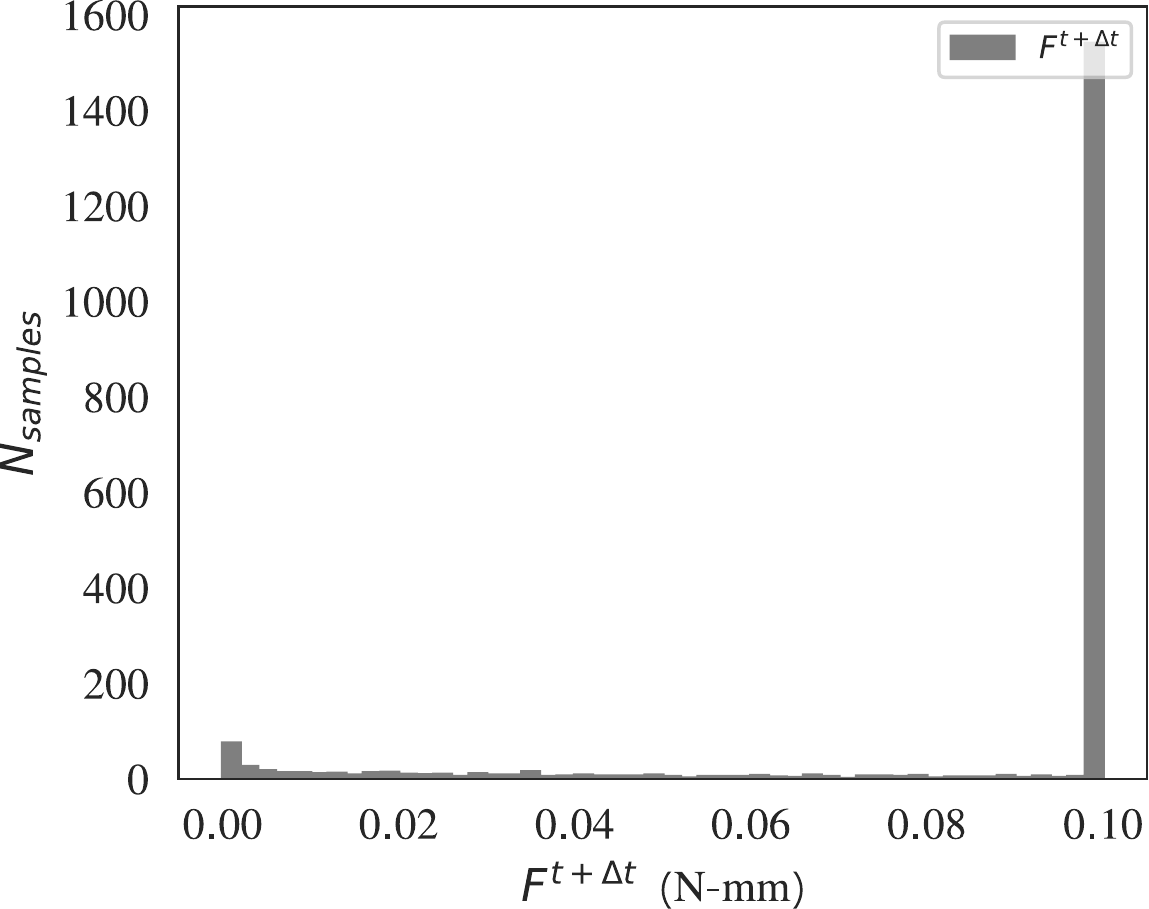}
	\caption{\footnotesize Helmholtz free-energy $F^{t+\Delta t}$.}
\end{subfigure}\hspace{0.05cm}
\begin{subfigure}[t]{0.3\textwidth}
  \centering
  \includegraphics[width=.95\linewidth]{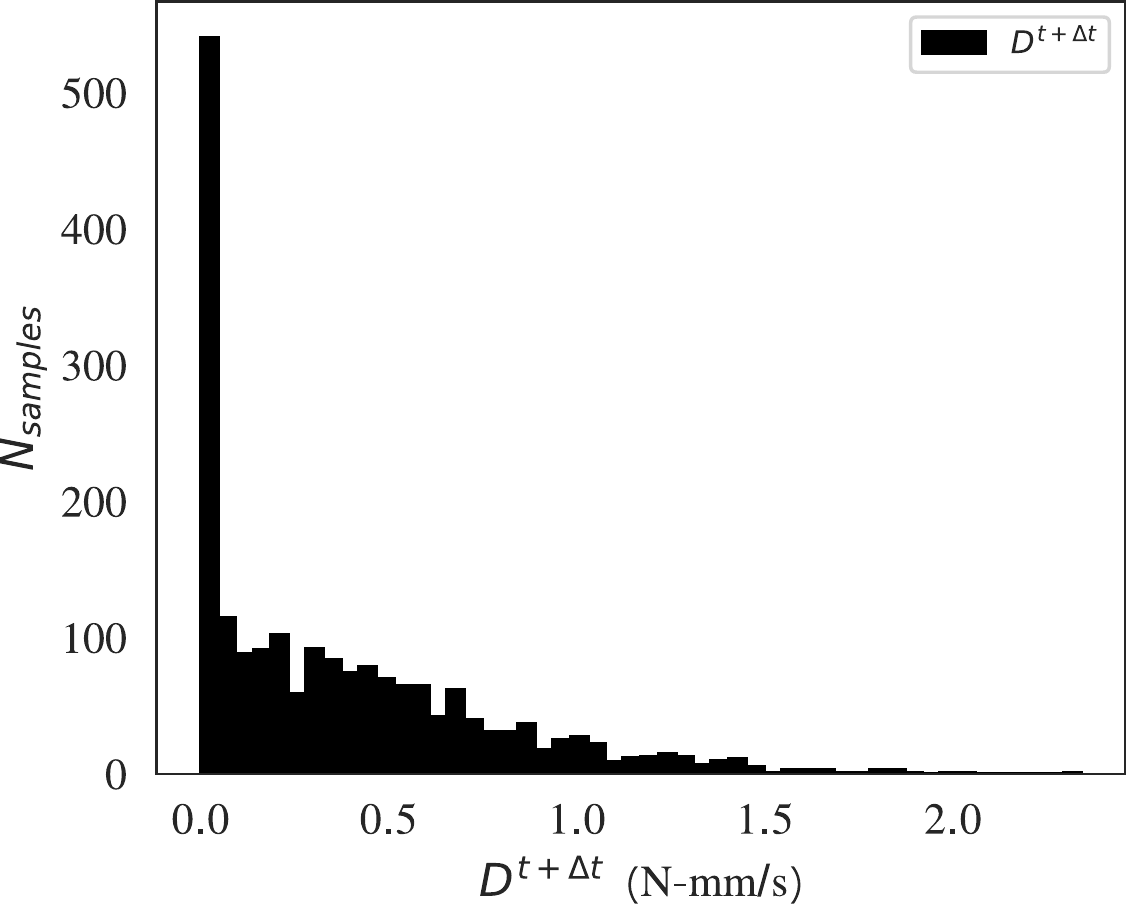}
	\caption{\footnotesize dissipation rate $D^{t+\Delta t}$.}
\end{subfigure}
	\caption{Sampling for one of the studied applications (paragraph \ref{par:1D}). 1D elasto-plastic material with Young's modulus $E=200$ GPa and yield strength $\sigma_y=200$ MPa. Standard deviations $\sigma^t_{\text{max}} = X^t_{\text{max}} = 2 \sigma_y$, $\varepsilon^t_{\text{max}} = \zeta^t_{\text{max}} = 10^{-2}$, and $\Delta \varepsilon^t_{\text{max}} = 10^{-3}$.}
	\label{f:distribution_data}
\end{figure*}

Herein, we generate data randomly, and not following prescribed loading paths. Conversely, this allows us to (1) improve the representativeness of the material data and (2) improve the generalization of the network on the strain increments. The initial state, $\Xi^t$ and $\varepsilon^t$, and the strain increment, $\Delta \varepsilon^t$, are randomly generated from standard distributions with mean value equal to zero and standard deviation equal to $\Xi_{\text{max}}$, $\varepsilon^t_{\text{max}}$, and $\Delta \varepsilon^t_{\text{max}}$, respectively. The Cauchy and thermodynamic stresses, $\sigma^t$ and $X_i^t$, as well as the internal variables $\zeta_i^t$ are then calculated to satisfy the constraint $y^t\leq 0$. This incremental procedure is repeated for $N_{\text{samples}}$, resulting in a set of $N_{\text{samples}}$ ordered pairs $\{\Xi^t, \varepsilon^t, \Delta \varepsilon^t; \Xi^{t+\Delta t}\}$, from which the corresponding energy potential and dissipation rate at time $t+\Delta t$ are evaluated.\\
The choice of the standard deviations $\Xi^t_{\text{max}}$, $\varepsilon^t_{\text{max}}$, and $\Delta \varepsilon^t_{\text{max}}$ depends on the investigated problem. As it follows, they are selected in a way such that $50\div60\%$ of data samples, $N_{\text{samples}}$, lie on the yield surface, i.e., $y^t=0$. Figure \ref{f:distribution_data} depicts the sampling for one of the investigated applications (see paragraph \ref{par:1D}). 

\section{Applications}
\label{sec:application}
Herein we use \textsf{TANN}s to the modeling of multi-dimensional elasto-plastic materials and demonstrate the wide applicability and effectiveness of \textsf{TANN}s. It is worth noticing that, even though the applications here investigated consist of elasto-plastic materials, the proposed class of ANN can be successfully applied (without any modification) to materials with different or more complex behavior, accounting e.g. for damage and/or other non-linearities (in the framework of strain-rate independent processes). In paragraph \ref{par:1D}, a one-dimensional motivating example, a spring-slider model, is studied. Then, the more general cases of three-dimensional elasto-plasticity (paragraph \ref{par:3D_VM}), accounting for perfect-plasticity, hardening and softening behaviors, are presented.\\
As it follows, the hyper-parameters (i.e., number of hidden layers, neurons, activation functions, etc.) of the networks are selected to give the best predictions, while requiring minimum number of hidden layers and nodes per layer. This is accomplished by comparing the learning error on the set of test patterns, per each trial choice of the hyper-parameters. In each training process, we use the commonly used technique of early-stopping (see par. \ref{par:second_order_grad}): the iterative update of weights and biases is stopped as the test error starts to increase while the learning error still decreases (indicator of an over-fitting  of the training data).\\
Throughout this Section relatively simple deep feed-forward neural networks architectures are used (with, at maximum, two hidden layers) and no additional regularization techniques are employed (e.g., L1/L2 penalties, dropout, etc.). Each numerical example is accompanied with a detailed discussion about the network architecture.
\subsection{1D elasto-plasticity with kinematic hardening}
\label{par:1D}
The Helmholtz free-energy potential, dissipation rate, and yield function that define the elasto-plastic 1D model with kinematic hardening (1D spring-slider, \cite{houlsby2007principles}) are:
\begin{equation}
\begin{split}
\textsf{F} = &\frac{E}{2} \left(\varepsilon - \zeta \right)^2 + \frac{H}{2} \zeta^2, \quad \textsf{D} = k |\dot{\zeta}|,\\
\text{and} \quad
y = &\frac{|\sigma - H \zeta|}{k} -1 \leq 0,
\end{split}
\label{eq:def1D}
\end{equation}
with $H$ being the kinematic hardening/softening-parameter and $k$ being the yield strength (slider threshold). The internal variable, $\zeta$, represents herein the plastic deformation. Table \ref{tab:1D_mat} displays the choice of the material parameters, selected to represent a steel-like material with either (1) perfect-plastic, (2) hardening, or (3) softening behavior.   

\begin{table}[hbt]
\catcode`?=\active \def?{\kern\digitwidth}
\caption{Material parameters for 1D elasto-plastic materials.}
\label{tab:1D_mat}
\medskip
\centering
\begin{adjustbox}{max width=0.3\textwidth}
\begin{tabular}{@{}@{\extracolsep{\fill}}cccc}
\toprule
case & $E$ & $k$ & $H$\\
& (GPa) & (MPa) & (GPa)\\
\midrule
1D-1& 200 & 200 & 0\\
1D-2& 200 & 200 & 10\\
1D-3& 200 & 200 & -10\\
\bottomrule
\end{tabular}
\end{adjustbox}
\end{table}

\subsubsection{Training}
\begin{figure}[h]
	\centering
	\includegraphics[width=0.4\textwidth]{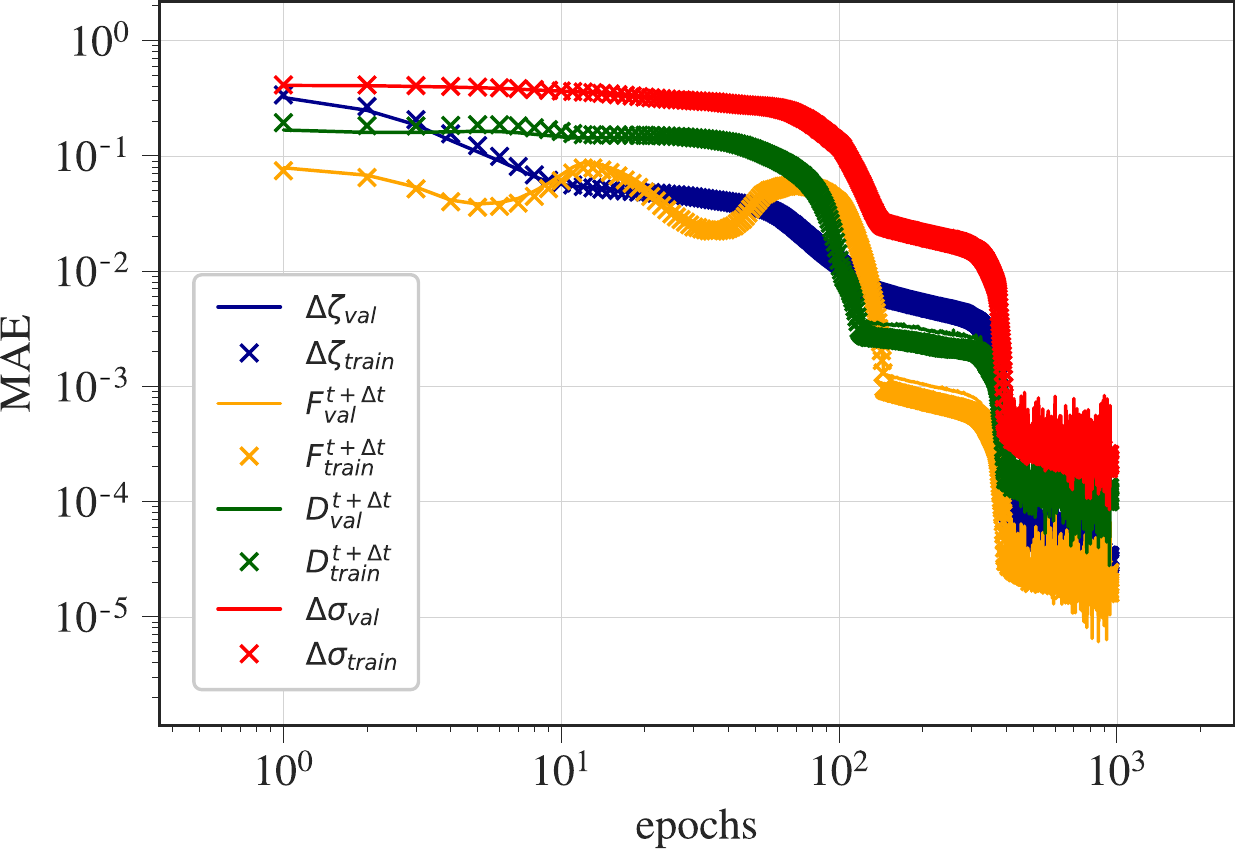}
	\caption{Errors of the predictions of \textsf{TANN} (loss functions), as the training is being performed, evaluated with respect to the training (train) and validation (val) sets. Weights and biases update are computed only on the training set.}
	\label{f:1D_training}
\end{figure}

According to the procedure detailed in Section \ref{sec:generation}, 2000 data (random increments at random states) are generated, for each (material) case, with the procedure detailed in Section \ref{sec:generation}. Training is performed with 50 \% of them (i.e., 1000). A validation set of 500 samples (validation data) is used to avoid over-fitting together with early stopping rules \cite{chen1995universal}. The performance of the predictions, at the end of the training, is evaluated on a set of 500 samples (test data). The sampling for material case 1D-1 is shown in Figure \ref{f:distribution_data}, while the samples distribution for cases 1D-2 and 1D-3 are presented in SM. Adam optimizer with Nesterov's acceleration gradient \cite{dozat2016incorporating} is selected and a batch size of 10 samples is used. We use the Mean Absolute Error (MAE) as loss functions for each output in order to assure the same precision between data of low and high numerical values (cf. Mean Square Error). Regularized weights are used to have consistent order of magnitude of different quantities involved in the loss functions. The architecture of \textsf{TANN} for all 1D cases consists of one hidden layer with 6 neurons (and leaky ReLU activation function) for the predictions of $\Delta \zeta$ and one hidden layer with 9 neurons (activation $\text{ELU}_{z^2}$) to predict $\textsf{F}^{t+\Delta t}$. The output layers for both sub-networks have linear activation function and biases set to zero. The corresponding number of degrees of freedom, i.e., the number of the hyper-parameters, is 72. Higher number of hidden layers could be used as well, but this is out of the scope of our investigations. Figure \ref{f:1D_training} displays the loss functions of each output as the training is performed, i.e., in number of epochs. The early stopping rule assures convergence with MAEs of the same order of magnitude for the 4 outputs, $\Delta \zeta$, $\textsf{F}^{t+\Delta t}$, $\Delta \sigma$, and $\textsf{D}^{t+\Delta t}$. Similar behaviors in training and performance with respect to the set of test data are also found for cases 1D-2 and 1D-3. For this reason, they are not presented herein. 

\subsubsection{Predictions in recall mode}
\begin{figure*}[ht]
\centering
\begin{subfigure}[b]{0.245\textwidth}
  \centering
  \includegraphics[width=\linewidth]{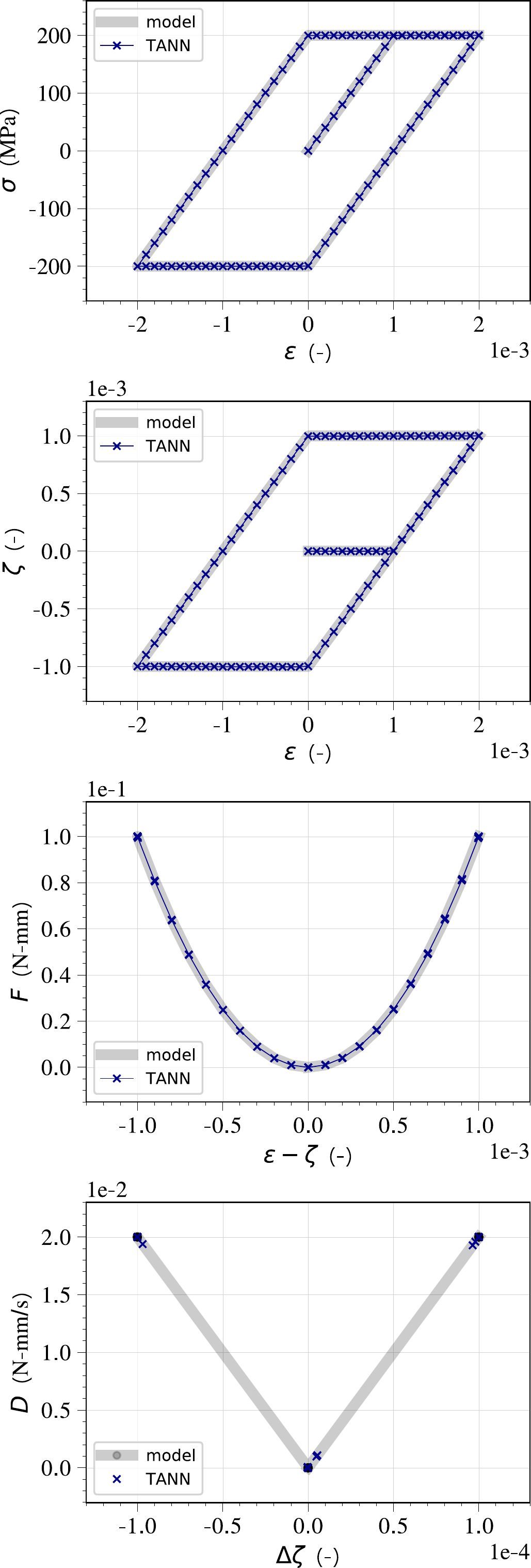}
	\caption{\footnotesize material case 1D-1.}
	\label{f:cyc_1D_ex1}
\end{subfigure}
\begin{subfigure}[b]{0.245\textwidth}
  \centering
  \includegraphics[width=\linewidth]{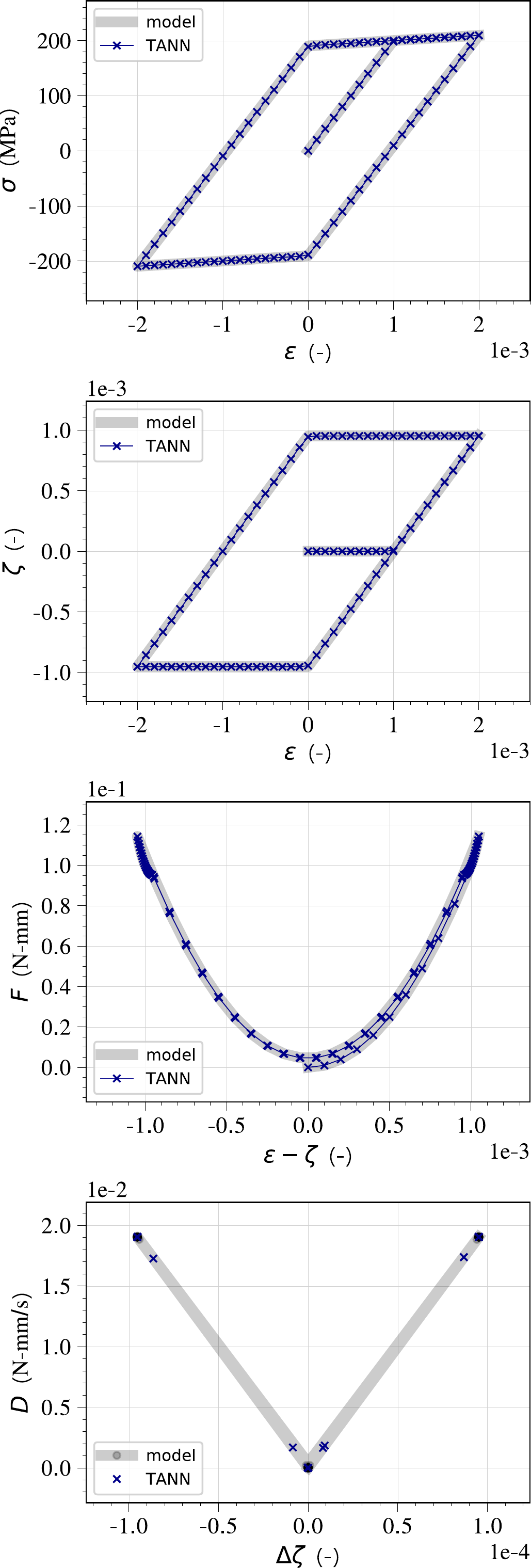}
	\caption{\footnotesize material case 1D-2.}
	\label{f:cyc_1D_ex2}
\end{subfigure}
\begin{subfigure}[b]{0.245\textwidth}
  \centering
  \includegraphics[width=\linewidth]{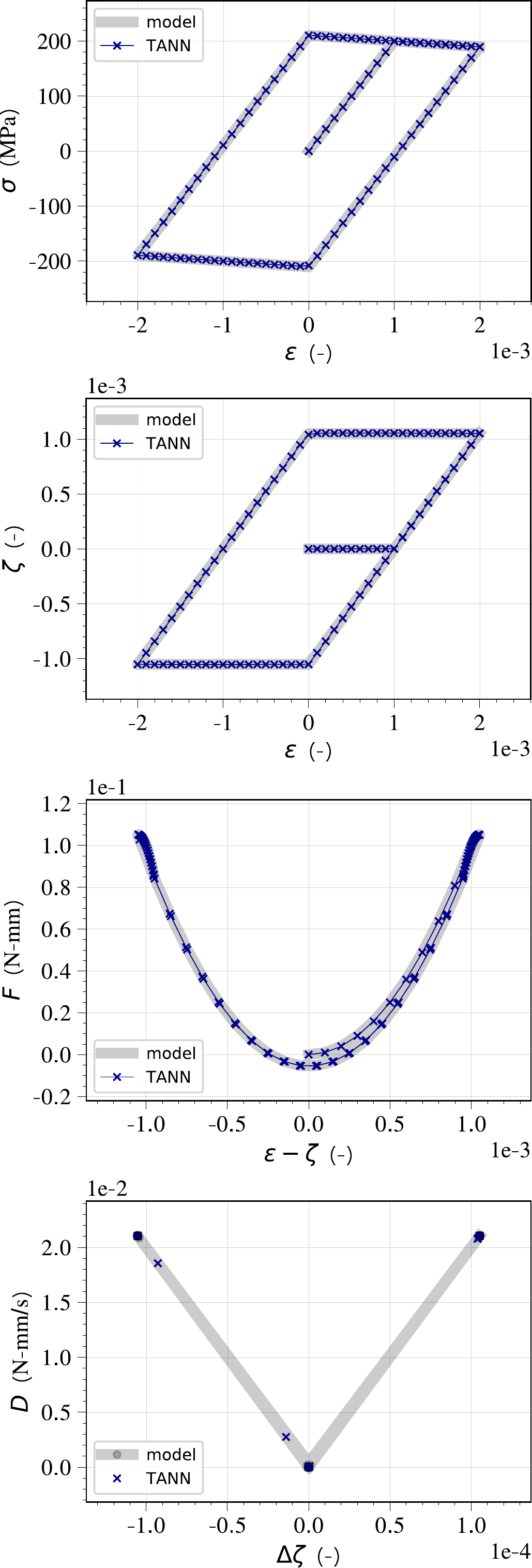}
	\caption{\footnotesize material case 1D-3.}
	\label{f:cyc_1D_ex3}
\end{subfigure}
\caption{Predictions of \textsf{TANN} due to cyclic loading, compared with the target constitutive model for cases (a) 1D-1, (b) 1D-2, and (c) 1D-3 (see Tab. \ref{tab:1D_mat}), with strain increments $\Delta {\varepsilon}=10^{-4}$.}
\label{f:cyc_1D_ex}
\end{figure*}

Once the neural networks have been trained, we use them in recall mode to predict the stress increment for a given strain, strain increment, and possibly other variables, and we compare the predictions with the corresponding targets. The results of the numerical integration scheme presented in Section \ref{sec:generation} are here considered as the exact solution of the material response. In particular, starting from an initial configuration, we make cyclic (or random) increments of the strain, $\Delta \varepsilon$. \textsf{TANN} hence predicts the corresponding increments, $\{\Delta \zeta, \Delta \sigma\}$, which will be transformed into the inputs in the successive call, as well as the energy and dissipation rates, $\{\textsf{F}^{t+\Delta t}, \textsf{D}^{t+\Delta t}\}$. This procedure is applied recursively. The neural network is so self-fed. Figures \ref{f:cyc_1D_ex1}, \ref{f:cyc_1D_ex2}, and \ref{f:cyc_1D_ex3} illustrate$-$for cases 1D-1, 1D-2, and 1D-3, respectively$-$the predictions of \textsf{TANN} for cyclic paths with strain increments $\Delta \varepsilon^n = \Delta \varepsilon \text{ sgn}\left(\cos \frac{n \pi}{2N} \right)$, where $n=1,2, \ldots$, $N= \varepsilon_{\text{max}}/\Delta {\varepsilon}$, $\varepsilon_{\text{max}}=2\times 10^{-3}$, and $\Delta {\varepsilon}=10^{-4}$.
\textsf{TANN} is found to successfully predict all quantities of interest. Moreover, and most important, the architecture and the training of the network allows to obtain thermodynamically consistent results. The first law of thermodynamics is automatically satisfied as a result of the structure of \textsf{TANN} and the predicted dissipation rate is always positive. Indeed, even if the second principle of thermodynamics is not explicitly assured by the \textsf{TANN} architecture, the fact that the training has been performed with consistent material data (i.e., positive dissipation rate) results automatically in the fulfillment of the second principle. Moreover, the linear dependency of dissipation with respect to $\dot{\zeta}$, property that stems from the strain rate independent material formulation, is also automatically recovered by \textsf{TANN}.
 
\subsubsection{Generalization of the network}
\label{par:generalization1D}
\begin{figure}[ht]
\centering
\begin{subfigure}{0.245\textwidth}
  \centering
  \includegraphics[height=0.78\textheight]{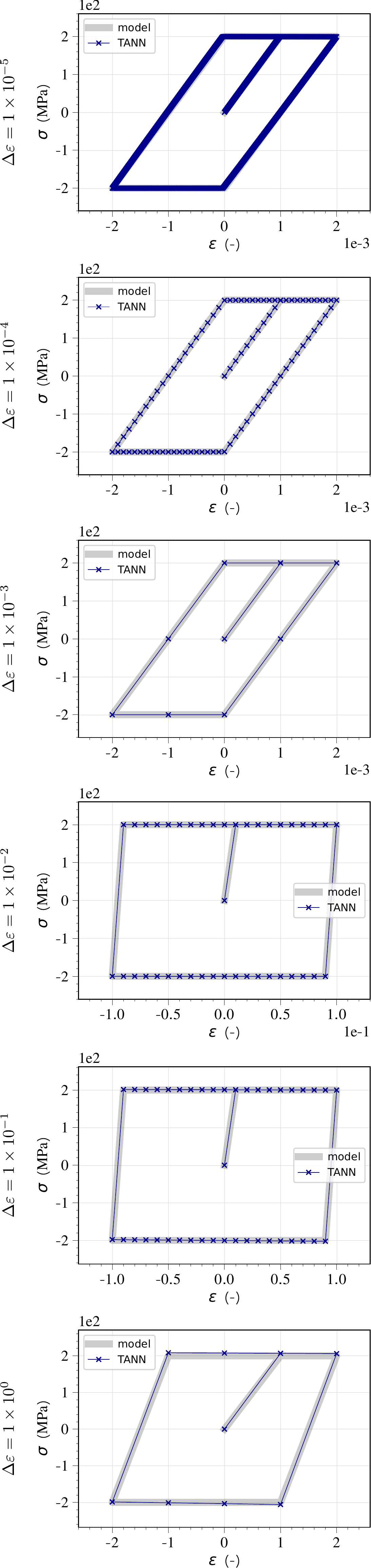}  
  \caption{\scriptsize $\sigma$ prediction.}
  \label{f:cyc_1D_pft_TANNa}
\end{subfigure}
\begin{subfigure}{0.245\textwidth}
  \centering
  \includegraphics[height=0.78\textheight]{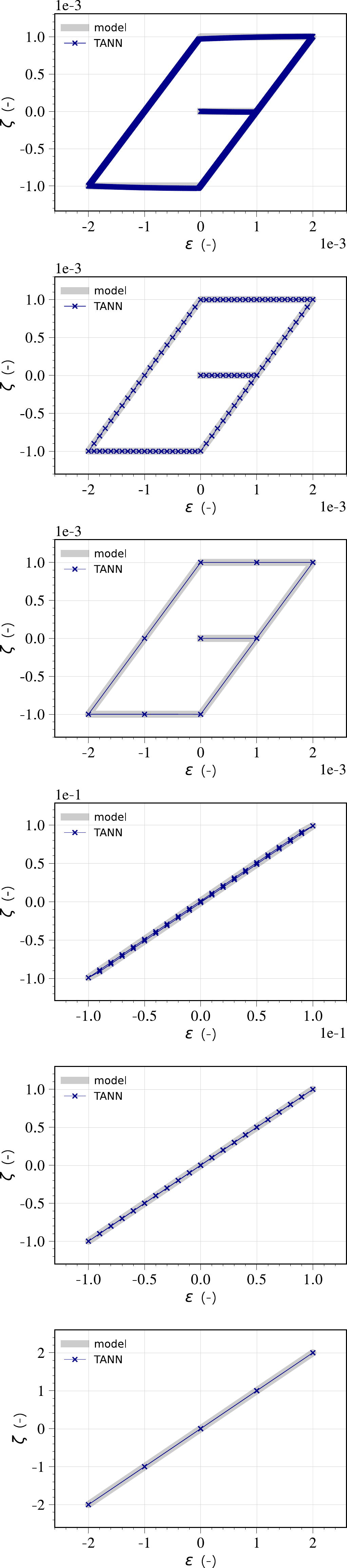} 
  \caption{\scriptsize $\zeta$ prediction.}
  \label{f:cyc_1D_pft_TANNb}
\end{subfigure}
\begin{subfigure}{0.245\textwidth}
  \centering
  \includegraphics[height=0.78\textheight]{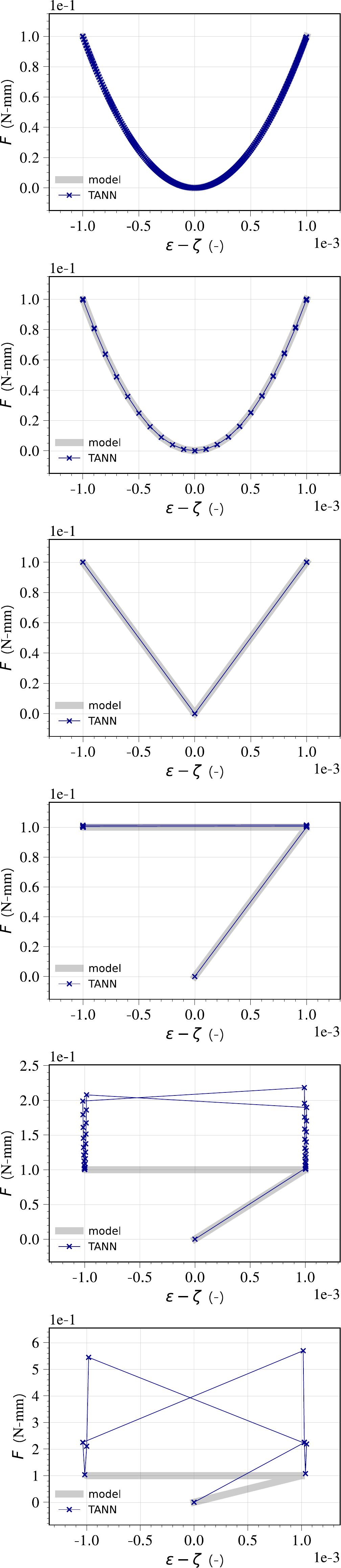}  
  \caption{\scriptsize $\textsf{F}$ prediction.}
  \label{f:cyc_1D_pft_TANNc}
\end{subfigure}
\begin{subfigure}{0.245\textwidth}
  \centering
  \includegraphics[height=0.78\textheight]{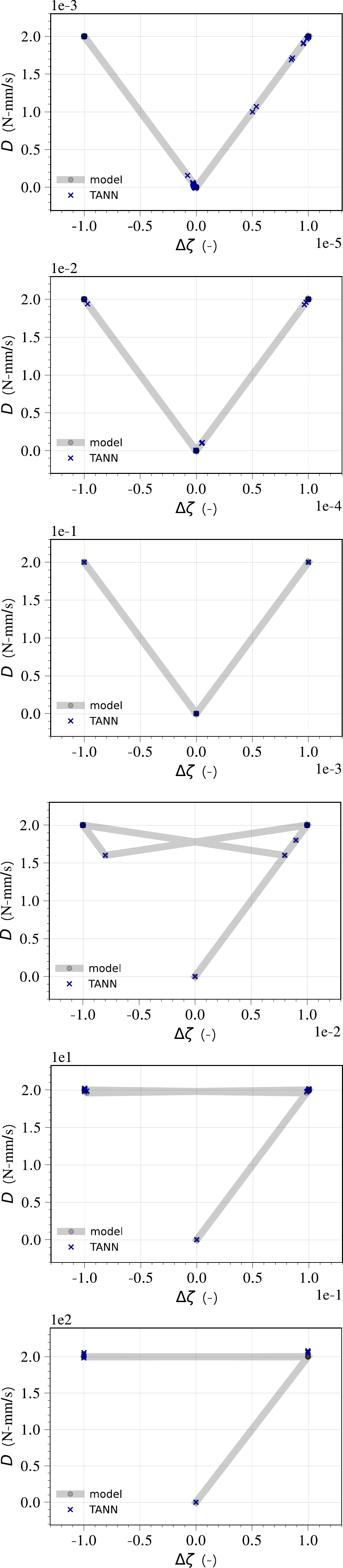} 
  \caption{\scriptsize $\textsf{D}$ prediction.}
  \label{f:cyc_1D_pft_TANNd}
\end{subfigure}
\caption{Sensitivity on the inputs for material case 1D-1, with strain increments $\Delta \varepsilon = \Delta {\varepsilon} \text{ sgn}\left(\cos \frac{n \pi}{2N} \right)-$with $N= \varepsilon_{\text{max}}/\Delta {\varepsilon}$, $\varepsilon_{\text{max}}=2\times10^{-3} \div 2$, and $\Delta {\varepsilon}$ varying from $\times 10^{-5}$ (top) to $1$ (bottom). Each column displays the response (from left to right) in term of $\Delta \sigma$ (a), $\Delta \zeta$ (b), $\textsf{F}^{t+\Delta t}$ (c), and $\textsf{D}^{t+\Delta t}$ (d). Each row represents the prediction at different $\Delta \varepsilon$.}
\label{f:cyc_1D_pft_res}
\end{figure}

Herein we investigate the generalization capability of \textsf{TANN} (i.e., the ability to make predictions for loading paths different from those used in the training operation). This is achieved by feeding the trained network with input values that not necessarily belong to the training range. With reference to Figures \ref{f:distribution_data} and S1 and S2 (SM file), the training ranges of the inputs are represented in Table \ref{tab:1D_range}. Figure \ref{f:cyc_1D_pft_res} displays the predictions for a cycling loading path $\Delta \varepsilon^n = \Delta \varepsilon \text{ sgn}\left(\cos \frac{n \pi}{2N} \right)-$with $\Delta \varepsilon \in (10^{-5}, 1)$. We clearly see that for input variables outside the training range, the predictions of the network become less accurate. Nevertheless, the predictions are always thermodynamically consistent. Moreover, the quantities of primary interest, such as the stress, the internal state variable, and the energy are in extremely good agreement with the reference model. The same stands also for the dissipation rate. We notice, once more, that its values are always positive, even when the network is used for predictions beyond the training range. In the SM file we present the predictions for the plastic hardening and softening behaviors, respectively.\\

\begin{table*}[h]
\catcode`?=\active \def?{\kern\digitwidth}
\caption{Range of the value of inputs used for training, for material cases 1D-1, 1D-2, and 1D-3.}
\label{tab:1D_range}
\medskip
\centering
\begin{adjustbox}{max width=0.7\textwidth}
\begin{tabular}{@{}@{\extracolsep{\fill}}ccccc}
\toprule
case & $\max |\varepsilon^t|$, $\min |\varepsilon^t|$ & $\max |\Delta \varepsilon|$, $\min |\Delta \varepsilon|$ & $\max |\zeta^t|$, $\min |\zeta^t|$ & $\max |\sigma^t|$, $\min |\sigma^t|$\\
& (-) & (-) & (-) & (MPa) \\
\midrule
1D-1& $4\times 10^{-2}$, $1\times10^{-7}$ & $1.4\times 10^{-2}$, $1\times10^{-7}$ & $4\times 10^{-2}$, $6\times10^{-6}$ & $200$, $0.168$\\
1D-2& $4\times 10^{-2}$, $4.\times10^{-6}$ & $1.2\times 10^{-2}$, $1\times10^{-7}$ & $3.6\times 10^{-2}$, $6\times10^{-7}$ & $495$, $0.300$\\
1D-3& $4\times 10^{-2}$, $1\times10^{-7}$ & $1.4\times 10^{-2}$, $1\times10^{-7}$ & $3.5\times 10^{-2}$, $6\times10^{-6}$ & $505$, $0.060$\\
\bottomrule
\end{tabular}
\end{adjustbox}
\end{table*}

It is worth noticing that in all the cases, even for very large strain increments$-$for which the predictions of the network in terms of dissipation rate, energy potential, and (for some values) stress and internal variable differ from the target values$-$, \textsf{TANN} successfully predicts the Jacobian, i.e., $\frac{\partial \sigma}{\partial \epsilon}$, in very good agreement with the reference model. This is true even when the error in the stress prediction is not negligible. This is of particular importance for numerical simulations with implicit algorithms. Therefore, \textsf{TANN} can successfully replace complicated constitutive models or multiscale approaches, but considerably and safely decreasing the calculation cost, even when the requested increments are outside the training range.

\clearpage

\subsubsection{\textsf{TANN} vs ANN}
\label{par:TANNvsANN1D}
\begin{figure*}[ht]
\centering
\begin{subfigure}[b]{0.3\textwidth}
  \centering
  \includegraphics[height=0.15\textheight]{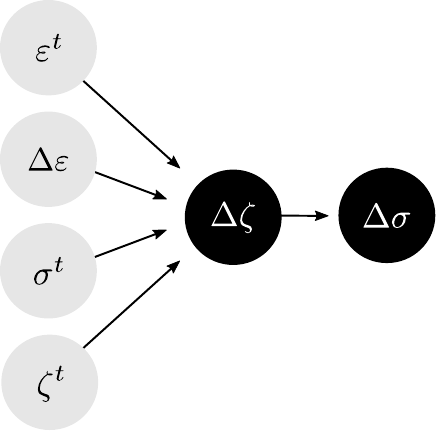}
	\caption{\footnotesize \textsf{ANN} scheme.}
	\label{f:ANN_arch_a}
\end{subfigure}
\hspace{1cm}
\begin{subfigure}[b]{0.5\textwidth}
  \centering
  \includegraphics[height=0.15\textheight]{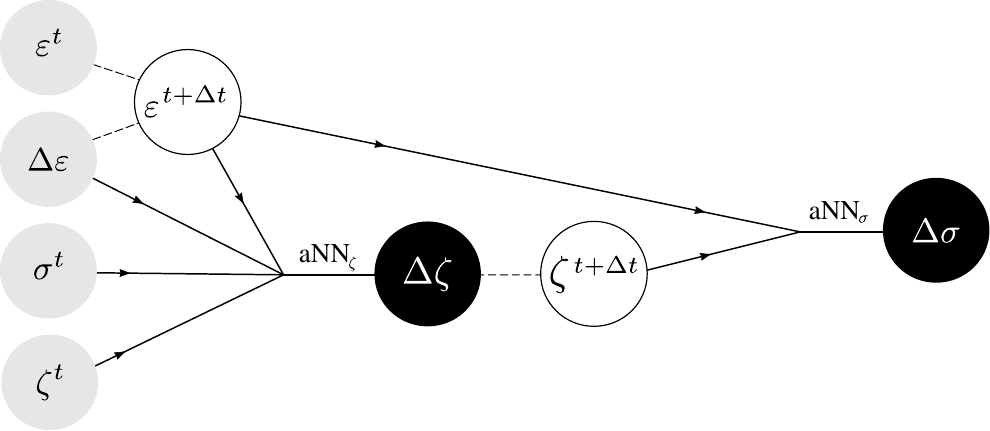}
	\caption{\footnotesize \textsf{ANN} architecture.}
	\label{f:ANN_arch_b}
\end{subfigure}
\caption{Schematic (a) and full architecture (b) of the network, not based on thermodynamics, \textsf{ANN}. Inputs are highlighted in gray (\tikzcircle[lightgray, fill=lightgray]{3pt}), outputs in black (\tikzcircle{3pt}).}
	\label{f:ANN_arch}
\end{figure*}

\begin{figure*}[ht]
\centering
\begin{subfigure}{0.4\textwidth}
  \centering
  \includegraphics[width=\textwidth]{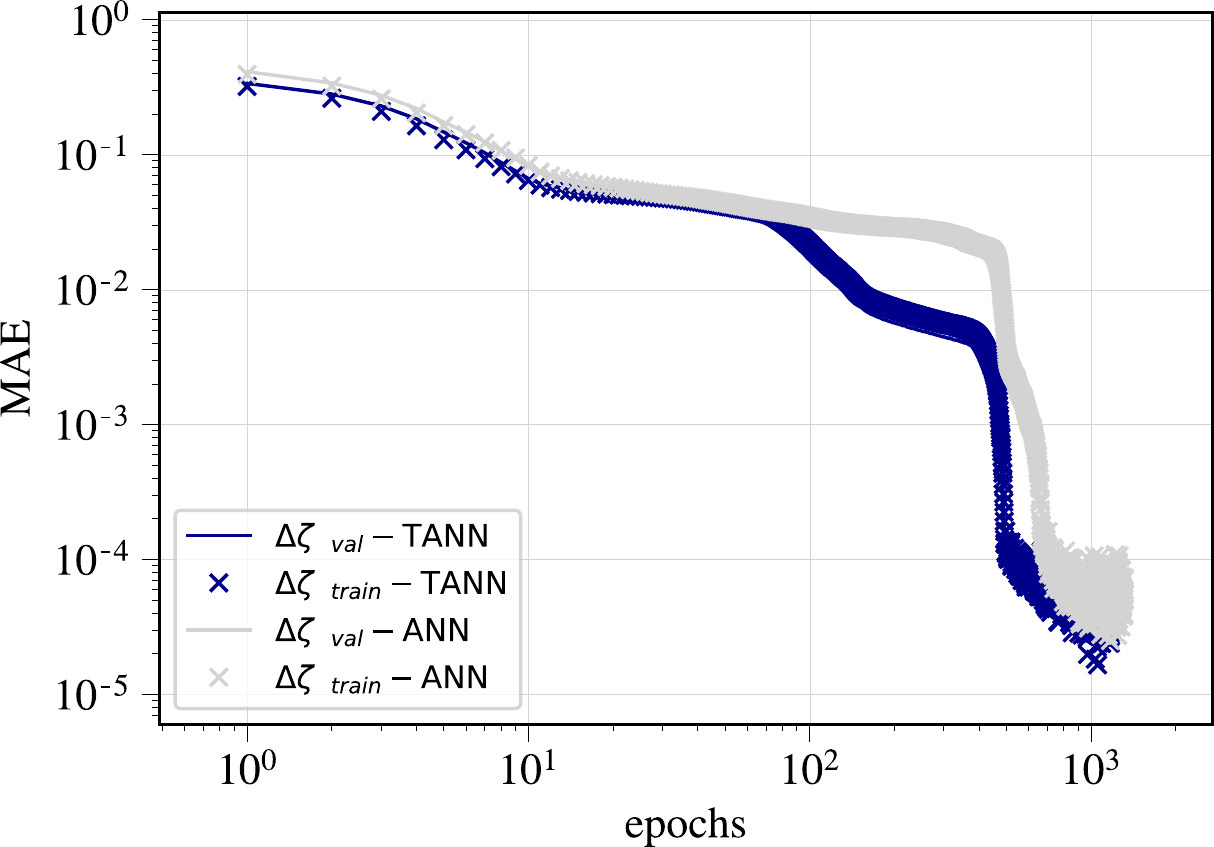}
  \caption{\footnotesize mean absolute error of $\Delta \zeta$ prediction}
\end{subfigure}
\begin{subfigure}{0.4\textwidth}
  \centering
  \includegraphics[width=\linewidth]{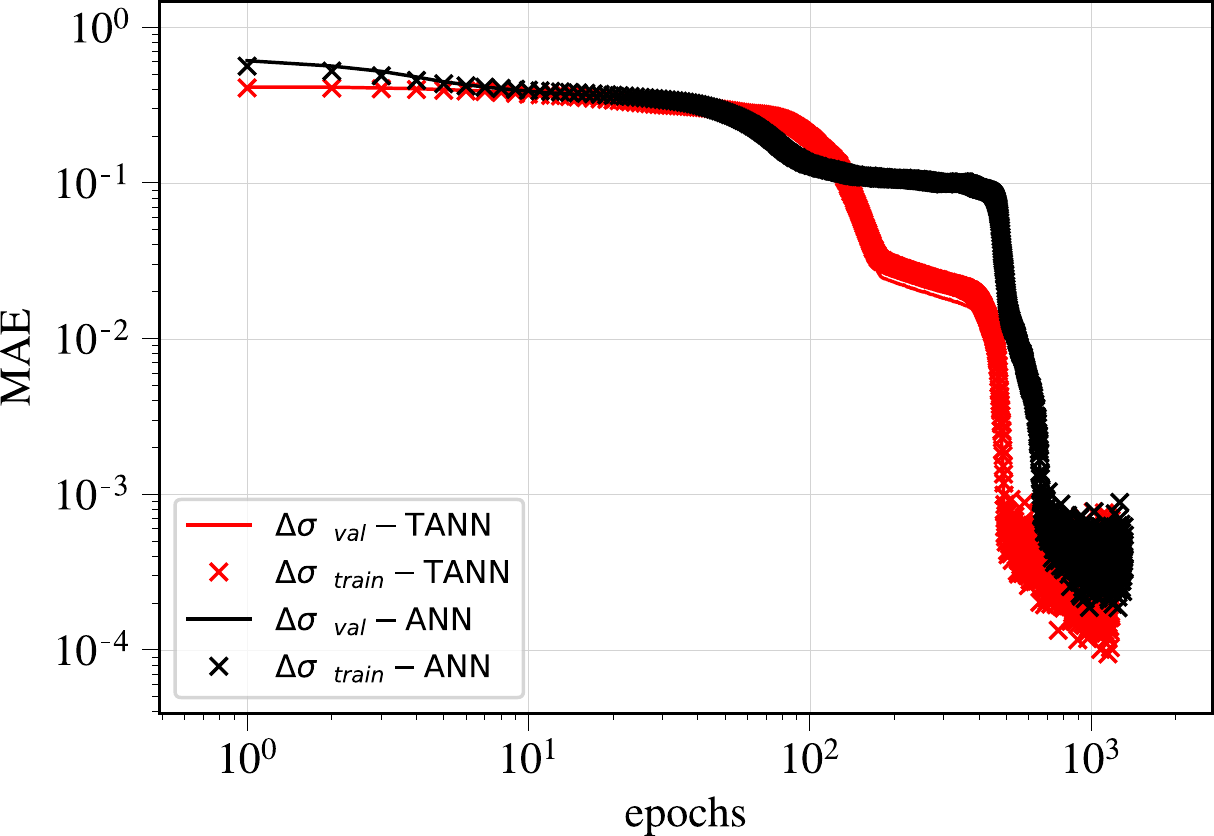}  
  \caption{\footnotesize mean absolute error of $\Delta \sigma_i$ prediction}
\end{subfigure}
	\caption{Training of \textsf{ANN} compared with \textsf{TANN} evaluated with respect to the training (train) and validation (val) sets.}
	\label{f:1D_training_ANN}
\end{figure*}

We compare herein the performance, in recall mode, of \textsf{TANN} with respect to the classical approach of ANN for constitutive modeling \cite{ghaboussi1991, lefik2003artificial}. Figure \ref{f:ANN_arch} displays the architecture of the network, \textsf{ANN}, with inputs $\mathcal{I}=(\varepsilon^t, \Delta \varepsilon, \sigma^t, \zeta^t)$ and output $\mathcal{O} = (\Delta \zeta, \Delta \sigma)$. As for \textsf{TANN}, the stress increment is derived by assuming ANN state variables (see Sect. \ref{sec:ann}) such that they coincide with the thermodynamic state variables, i.e., $\varepsilon^t$ and $\zeta^t$, and their increments, i.e., $\Delta \varepsilon$ and $\Delta \zeta$, as in \cite{IJET18938,YU20191}. \textsf{ANN} is thus composed of two sub-ANNs; $\textsf{aNN}_{\zeta}$ predicts the internal variables increment and $\textsf{sNN}_{\sigma}$ predicts the stress increment, i.e., $\Delta \sigma = \textsf{aNN}_{\sigma} (\varepsilon^{t+\Delta}, \Delta \varepsilon, \zeta^{t+\Delta t}, \Delta \zeta)$. The architecture of the network is selected to give the best performance while assuring the same amount of degrees of freedom, hyper-parameters, of \textsf{TANN}. Both sub-networks, $\textsf{aNN}_{\zeta}$ and $\textsf{aNN}_{\sigma}$, consist of one hidden layer, with 6 neurons each and leaky ReLU activation function. Same with \textsf{TANN}, the output layers have linear activation function and zero bias. Training is performed on the same set of samples that are used for the thermodynamics-based network. Figure \ref{f:1D_training_ANN} displays the error of the predictions of \textsf{ANN}, as training is performed, and compares it with \textsf{TANN}.

\begin{figure}[ht]
\centering
\begin{subfigure}[t]{0.245\textwidth}
  \centering
  \includegraphics[height=0.78\textheight]{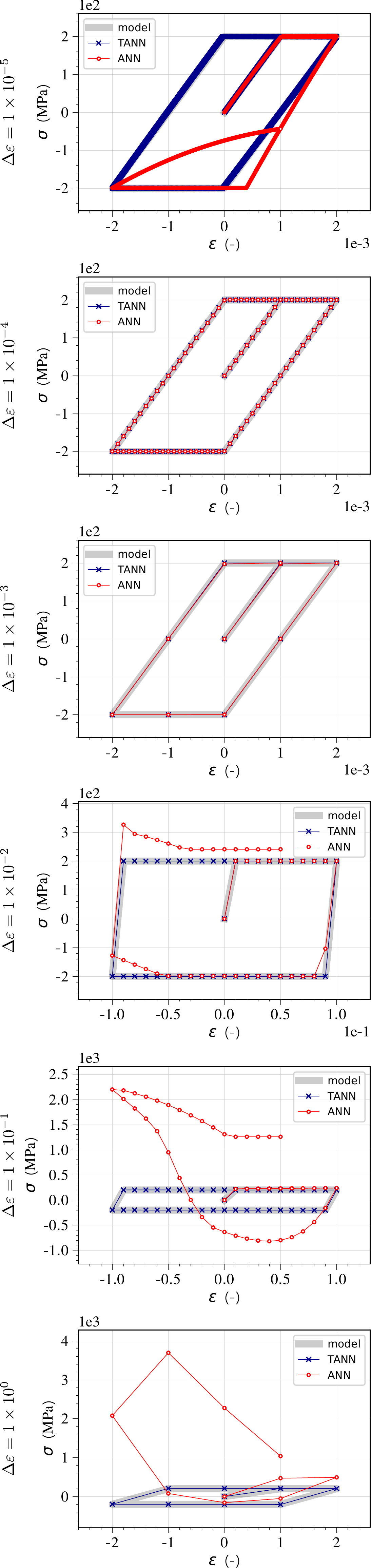}  
  \caption{\footnotesize $\sigma$ prediction.}
  \label{f:cyc_1D_pft_ANN_TANNa}
\end{subfigure}
\begin{subfigure}[t]{0.245\textwidth}
  \centering
  \includegraphics[height=0.78\textheight]{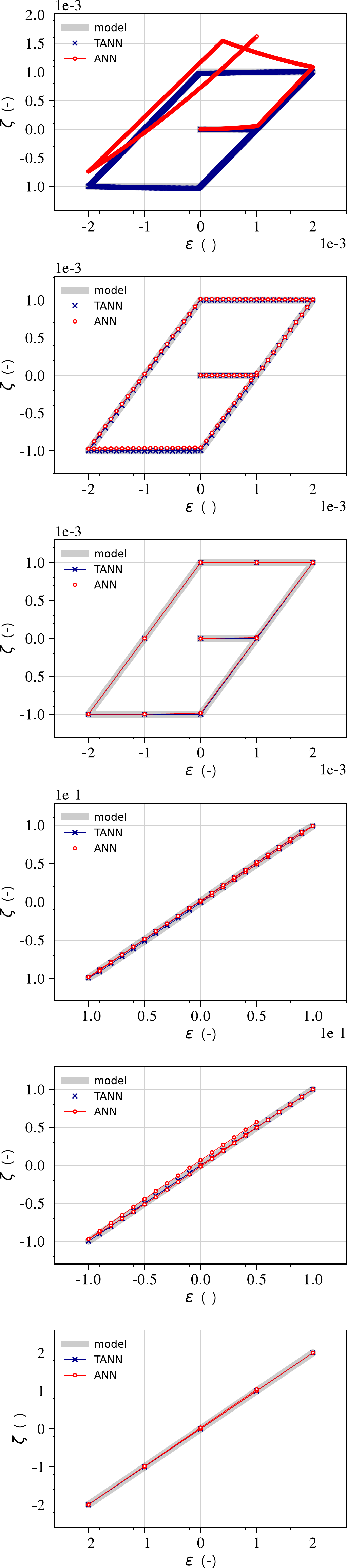} 
  \caption{\footnotesize $\zeta$ prediction.}
  \label{f:cyc_1D_pft_ANN_TANNb}
\end{subfigure}
\begin{subfigure}[t]{0.245\textwidth}
  \centering
  \includegraphics[height=0.78\textheight]{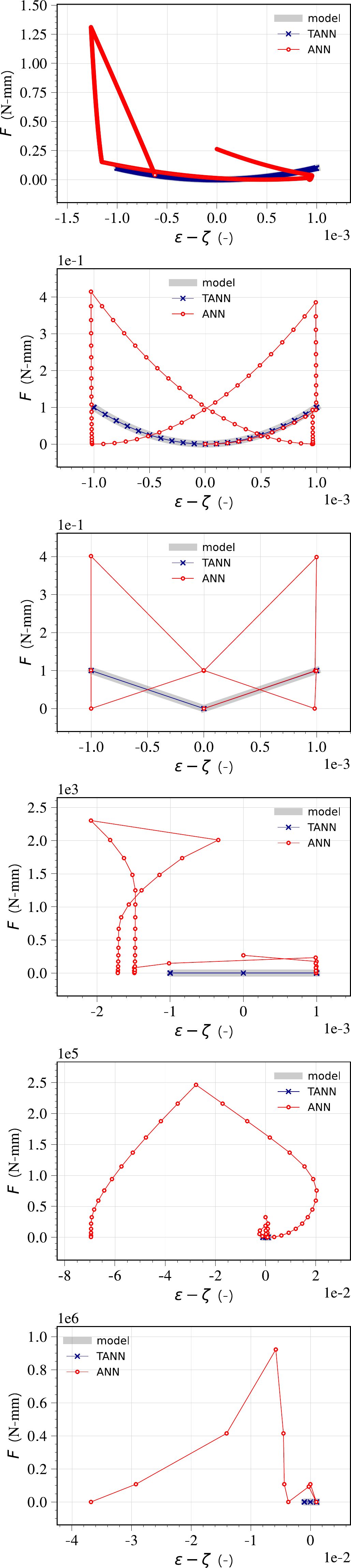}  
  \caption{\footnotesize $\textsf{F}$ prediction (\textsf{TANN}), Eq. (\ref{eq:def1D}) (\textsf{ANN}).}
  \label{f:cyc_1D_pft_ANN_TANNc}
\end{subfigure}
\begin{subfigure}[t]{0.245\textwidth}
  \centering
  \includegraphics[height=0.78\textheight]{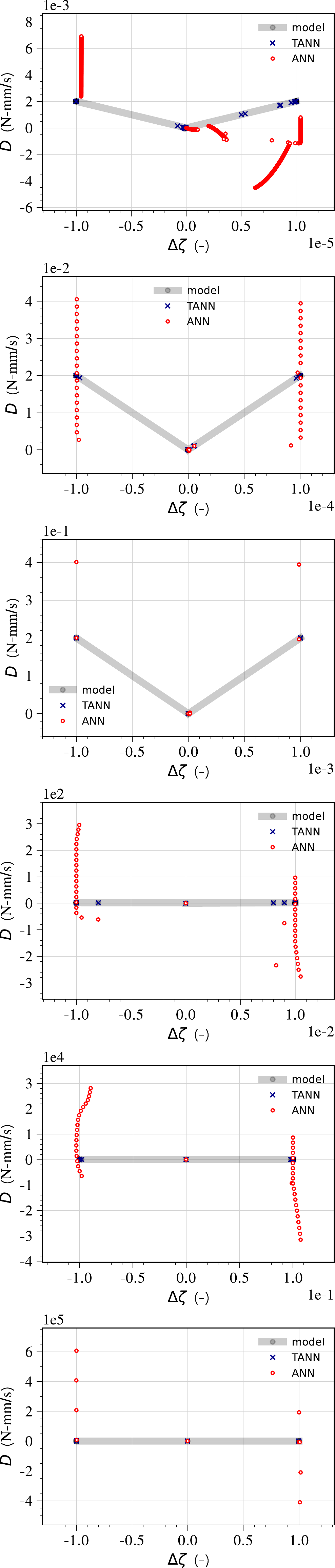} 
  \caption{\footnotesize $\textsf{D}$ prediction (\textsf{TANN}), Eq. (\ref{eq:def1D}) (\textsf{ANN}).}
  \label{f:cyc_1D_pft_ANN_TANNd}
\end{subfigure}
\caption{Comparison of the predictions of \textsf{TANN} and those of standard \textsf{ANN}, for material case 1D-1 (perfect plasticity). Each row represents the prediction at different $\Delta \varepsilon$.}
\label{f:cyc_1D_pft_ANN_TANN}
\end{figure}

We present in Figure \ref{f:cyc_1D_pft_ANN_TANN} the comparisons between the predictions of \textsf{TANN} and \textsf{ANN}, in terms of stress, for a cycling loading path $\Delta \varepsilon^n = \Delta \varepsilon \text{ sgn}\left(\cos \frac{n \pi}{2N} \right)$. The cases with strain hardening and strain softening are presented in the SM file. \textsf{TANN} is clearly superior in terms of (a) accuracy of the prediction and (b) generalization with respect to the inputs. Moreover, \textsf{ANN} predictions do not fulfill the principles of thermodynamics, even though the training of the network has been performed on consistent material data. This is clearly shown by computing from the predictions of \textsf{ANN} the increment of the Helmholtz free-energy and dissipation rate using the corresponding definitions, Eq. (\ref{eq:def1D}). Figures \ref{f:cyc_1D_pft_ANN_TANNc} and \ref{f:cyc_1D_pft_ANN_TANNd} display the computed quantities, $\textsf{F}$ and $\textsf{D}$, for material case 1D-1. The predictions of the standard ANN clearly do not respect the thermodynamics principles (both the first and second laws).

\begin{figure}[ht]
\centering
\begin{subfigure}{0.245\textwidth}
  \centering
  \includegraphics[height=0.65\textheight]{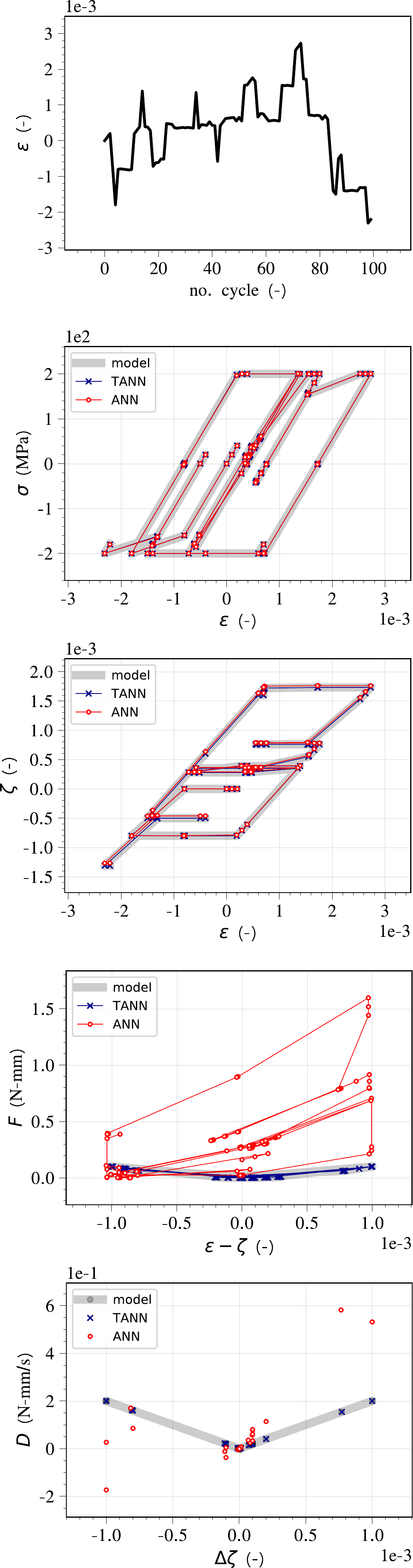}  
  \caption{\footnotesize material case 1D-1 (perfect plasticity).}
  \label{f:cyc_1D_rnd_ANN_TANN_pft}
\end{subfigure}
\begin{subfigure}{0.245\textwidth}
  \centering
  \includegraphics[height=0.65\textheight]{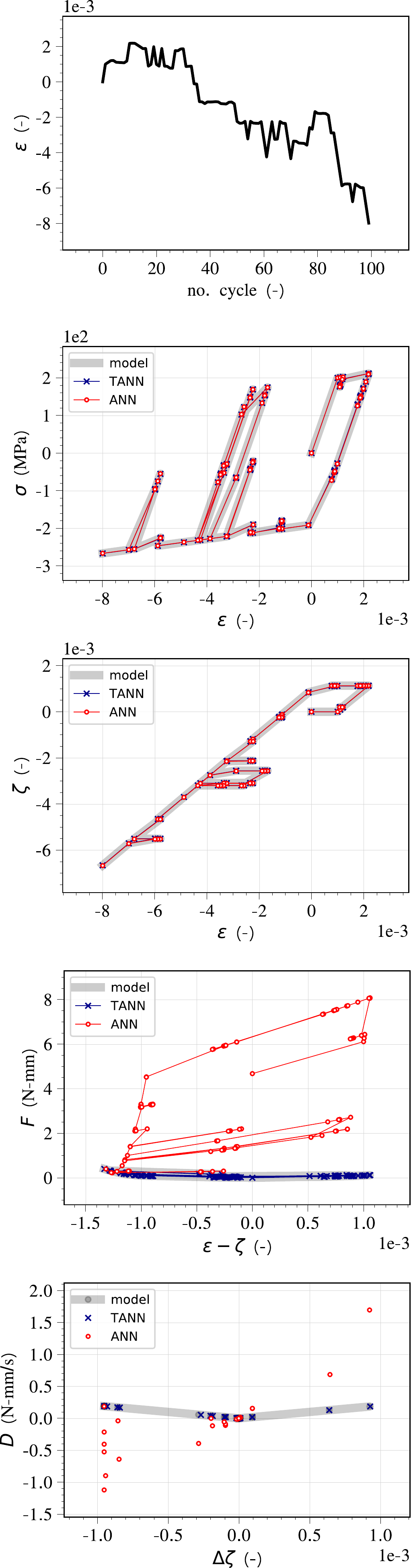} 
  \caption{\footnotesize material case 1D-2 (hardening).}
  \label{f:cyc_1D_rnd_ANN_TANN_hrd}
\end{subfigure}
\begin{subfigure}{0.245\textwidth}
  \centering
  \includegraphics[height=0.65\textheight]{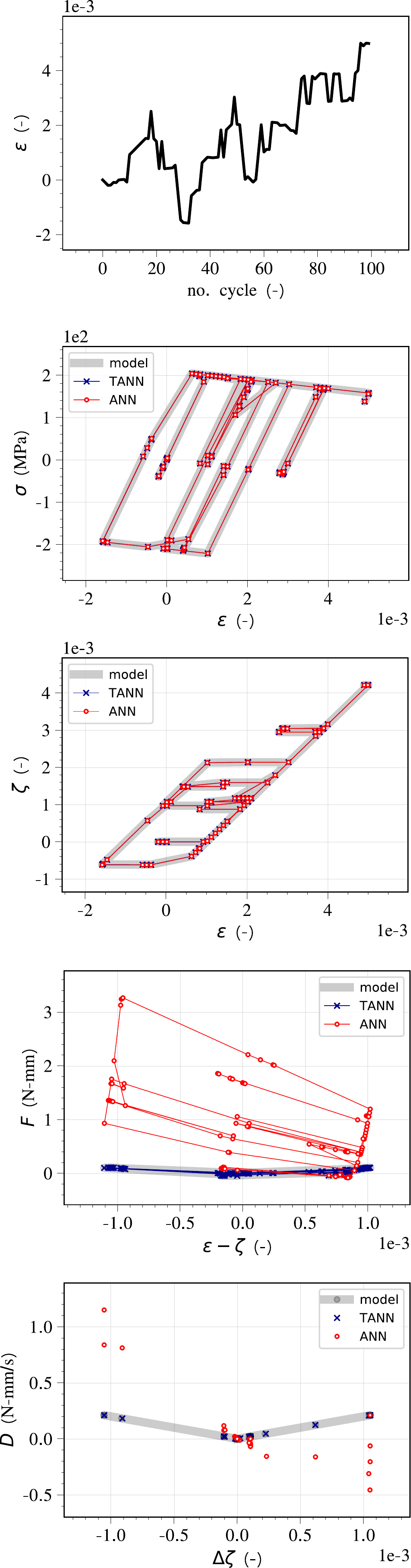}  
  \caption{\footnotesize material case 1D-3 (softening).}
  \label{f:cyc_1D_rnd_ANN_TANN_sft}
\end{subfigure}
\caption{Comparison of the predictions of \textsf{TANN} and standard \textsf{ANN} for a random loading path, for material cases 1D-1, perfect plasticity (a), 1D-2, hardening (b), and 1D-3, softening (c). The loading path is displayed at the first row, in each case.}
\label{f:cyc_1D_rnd_ANN_TANN}
\end{figure}

Figure \ref{f:cyc_1D_rnd_ANN_TANN} displays the predictions of both \textsf{TANN} and \textsf{ANN} for a random loading path, for a perfectly plastic behavior (Fig. \ref{f:cyc_1D_rnd_ANN_TANN_pft}), hardening (Fig. \ref{f:cyc_1D_rnd_ANN_TANN_hrd}), and softening (Fig. \ref{f:cyc_1D_rnd_ANN_TANN_sft}). Once more, the performance of the thermodynamics-based network, as well as its generalization capabilities, are significantly better than those of standard \textsf{ANN}. One could, of course, increase the number of layers and neurons in order to assure better predictions, but still there will not be guarantee that the predictions of standard \textsf{ANN} would be thermodynamically consistent. An increase of the number of samples used in the training operation may as well improve the predictions. This is not the case for \textsf{TANN}.
\clearpage

\subsection{3D elasto-plasticity}
\label{par:3D_VM}
In order to illustrate the performance of \textsf{TANN} in three dimensions we use the simple von Mises elasto-plastic model with kinematic hardening (and softening). The model can be derived from the following expressions of the energy potential and dissipation rate
\begin{equation}
\begin{split}
\textsf{F} = &\frac{9K}{2} \left(\varepsilon_p - \zeta_p \right)\cdot\left(\varepsilon_p - \zeta_p \right) +\\
 & +G \left(e - z \right)\cdot \left(e - z\right) + \frac{H}{2} z\cdot z,\\
\textsf{D} = & k \sqrt{2} \sqrt{\dot{z} \cdot \dot{z}},
\end{split}
\label{eq:def3D}
\end{equation}
where $k$ represents the elastic limit in simple shear; $K$ and $G$ are the bulk and shear moduli; $\varepsilon_p$ and $\zeta_p$ are, respectively, the mean total and plastic deformation; and $e$ and $z$ are, respectively, the total and plastic deviatoric strain tensors. The yield surface can be derived as shown in Appendix A \cite{houlsby2007principles} and is defined as
\begin{equation}
y = D - X' \cdot z = \sqrt{X' \cdot X'} -\sqrt{2}k \leq 0,
\end{equation} 
with $X'_{ij} = 2 G \left(e_{ij} - z_{ij}\right) + H z_{ij}$.

\begin{table}[hbt]
\catcode`?=\active \def?{\kern\digitwidth}
\caption{Material parameters for 3D elasto-plastic von Mises material.}
\label{tab:3D_mat}
\medskip
\centering
\begin{adjustbox}{max width=0.3\textwidth}
\begin{tabular}{@{}@{\extracolsep{\fill}}ccccc}
\toprule
case & $K$ & $G$ & $k$ & $H$\\
& (GPa) & (GPa) & (MPa) & (GPa)\\
\midrule
3D-1& 167 & 77 & 140 & 0\\
3D-2& 167 & 77 & 140 & 10\\
3D-3& 167 & 77 & 140 & -10\\
\bottomrule
\end{tabular}
\end{adjustbox}
\end{table}

\subsubsection{Training}
Data are generated as detailed in Section \ref{sec:generation}. A total of 6000 data with random increments of deformation are generated. In order to improve the performance of the network in recall mode, additional sampling with random uni-axial and bi-axial loading paths are also used. The samples are split into training (50\%), validation (25\%), and test (25\%) sets. The sampling in terms of the mean and deviatoric stresses, $p$ and $q$, and deformations, $\varepsilon_p$ and $e$, is presented in the SM file. For the sake of simplicity, stress and deformation are converted in the principal axes frame of reference.\\
The network architecture is adapted to the size of the inputs and outputs, with respect to the mono-dimensional case. In particular, the sub-network $\textsf{sNN}_{\zeta}$ consists of two hidden layers, with 48 neurons (leaky ReLU activation function), and three output layers, one per each (principal) component of (increment of) $\zeta$. The sub-network $\textsf{s-NN}_{\textsf{F}}$ has one hidden layer with 36 neurons (activation $\text{ELU}_{z^2}$). The output layers for both sub-networks have linear activation functions and biases set to zero. The resulting number of hyper-parameters is $\approx 3000$. The loss functions of each output as the training is performed, for material case 3D-1 (perfect plasticity), is presented in the SM file. Similar behaviors are also recovered for cases 3D-2 (hardening), 3D-3 (softening).

\subsubsection{Predictions in recall mode}
\begin{figure*}[ht]
\centering
\begin{subfigure}{0.245\textwidth}
  \centering
  \includegraphics[width=\linewidth]{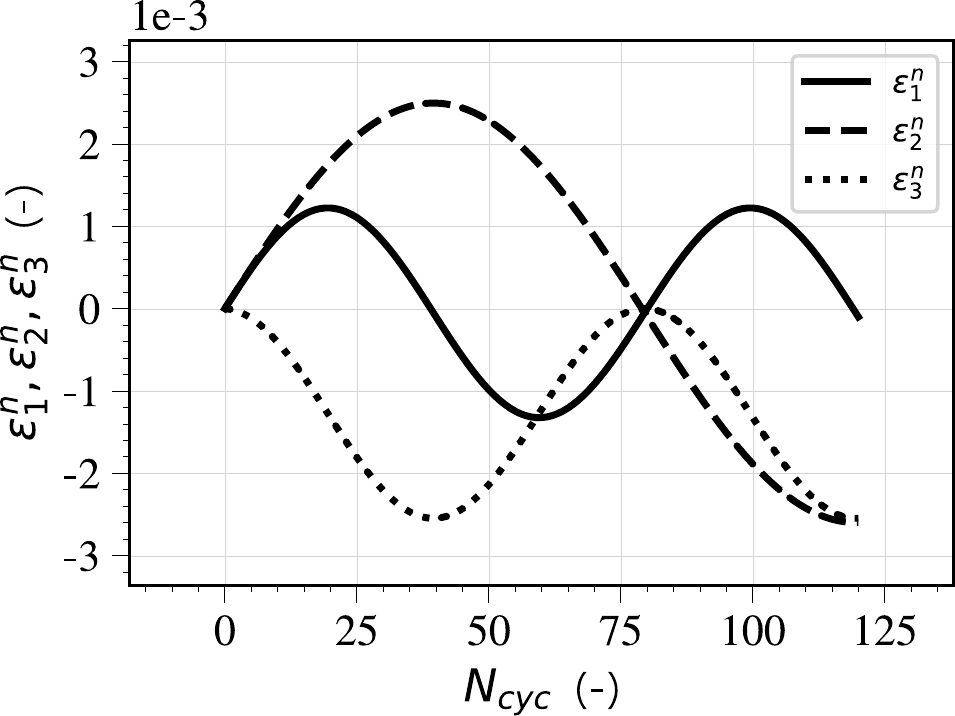}
  \caption{\footnotesize loading path.}
  \label{f:sin_TANN_prf_sub0}
\end{subfigure}
\medskip

\begin{subfigure}{0.735\textwidth}
  \centering
  \includegraphics[width=\linewidth]{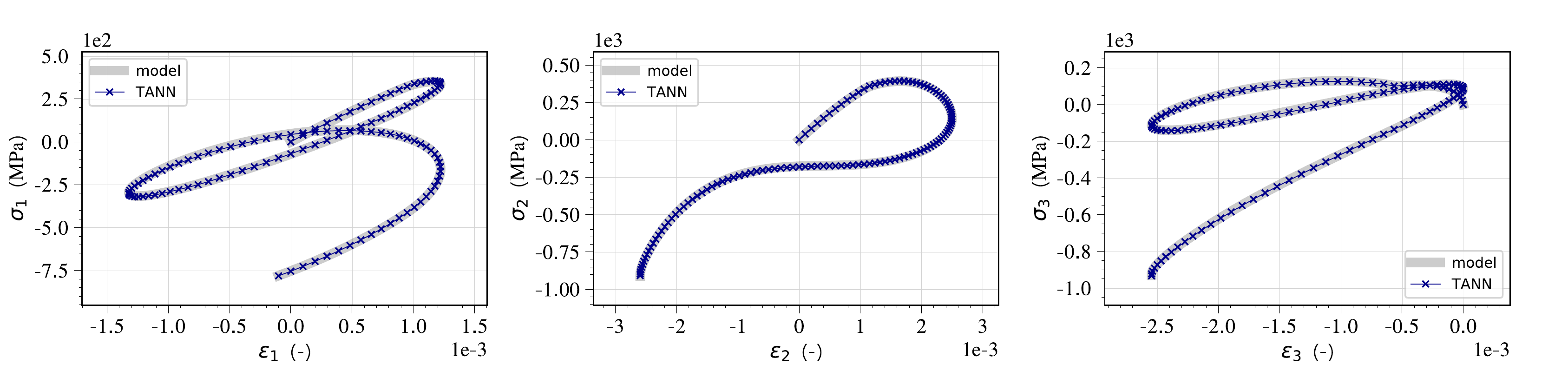}  
  \caption{\footnotesize $\sigma_i$ prediction.}
  \label{f:sin_TANN_prf_sub1}
\end{subfigure}
\medskip

\begin{subfigure}{0.735\textwidth}
  \centering
  \includegraphics[width=\linewidth]{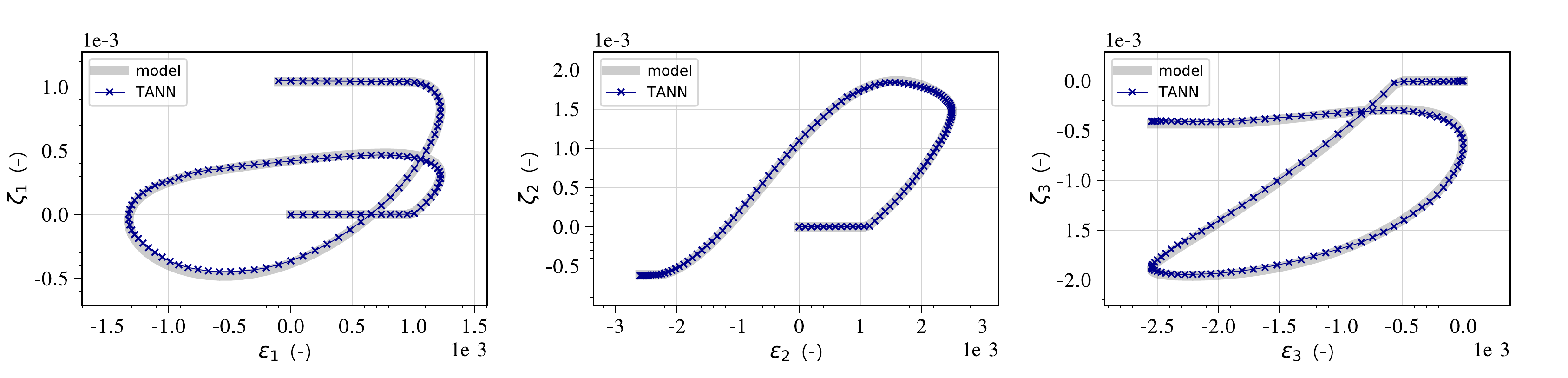}  
  \caption{\footnotesize $\zeta_i$ prediction.}
  \label{f:sin_TANN_prf_sub2}
\end{subfigure}
\medskip

\begin{subfigure}{0.49\textwidth}
  \centering
  \includegraphics[width=\linewidth]{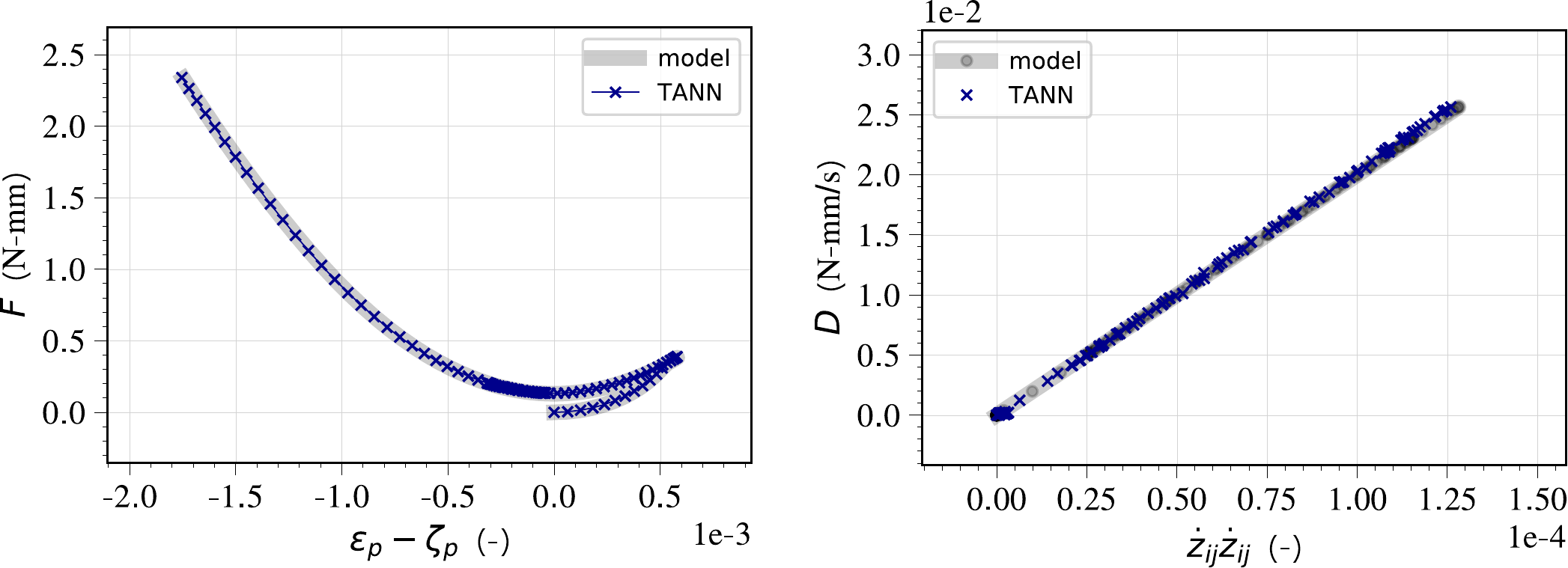}  
  \caption{\footnotesize energy (left) and dissipation rate (right) predictions.}
  \label{f:sin_TANN_prf_sub3}
\end{subfigure}
\caption{Predictions of \textsf{TANN} due to cyclic loading, compared with the target constitutive model, case 3D-1, perfect plasticity: loading path (\ref{f:sin_TANN_prf_sub0}), in terms of principal deformations; principal stress predictions ($\sigma_1, \sigma_2, \sigma_3$) (\ref{f:sin_TANN_prf_sub1}); plastic deformation predictions ($\zeta_1, \zeta_2, \zeta_3$) (\ref{f:sin_TANN_prf_sub2}); energy (\textsf{F}) and dissipation rate predictions (\textsf{D}) (\ref{f:sin_TANN_prf_sub3}).}
\label{f:sin_TANN_prf}
\end{figure*}

Once the network has been trained, it is used, in recall mode, to make predictions. We briefly present the performance of \textsf{TANN} in predicting the material response for a cyclic loading path. Figure \ref{f:sin_TANN_prf} depicts the comparison with the target material model for material case 3D-1. The predictions for the same loading path with material cases 3D-2 and 3D-3 are presented in the SM file. In all the cases, the network shows good performance.\\
Similarly with the 1D case, the generalization capabilities of the network are presented together with the comparison of the thermodynamics-based network with the standard ANN approach. 

\clearpage
\subsubsection{\textsf{TANN} vs standard ANN. Generalization of the network}
Herein we investigate the performance of \textsf{TANN} with respect to the classical approach of ANN, as well as the sensitivity with respect to the input variables range. Similarly to the comparisons in paragraph \ref{par:TANNvsANN1D}, we select a network with inputs $\mathcal{I}=(\varepsilon^t_i, \Delta \varepsilon_i, \sigma_i^t, \zeta_i^t)$ and output $\mathcal{O}=(\Delta \zeta_i, \Delta \sigma_i)$, with $i=1,2,3$ denoting the principal components. The architecture is selected to give the best performance, preserving the same number of hyper-parameters between \textsf{TANN} and standard \textsf{ANN}. The network, \textsf{ANN}, consists of the two sub-networks, $\textsf{aNN}_{\zeta}$ and $\textsf{aNN}_{\sigma}$, with two hidden layers, each one, leaky ReLU activation functions, and number of neurons per layer equal to 48. As for $\textsf{sNN}_{\zeta}$ and $\textsf{sNN}_{\sigma}$, in $\textsf{aNN}_{\zeta}$ and $\textsf{aNN}_{\sigma}$ three output layers (1 neuron each) are used, with linear activation functions and zero biases. In the SM file we present the comparison of the MAE of the network predictions with respect to the target values (training and validation data-sets).

\begin{figure*}[ht]
\centering
\begin{subfigure}{0.735\textwidth}
  \centering
  \includegraphics[width=\textwidth]{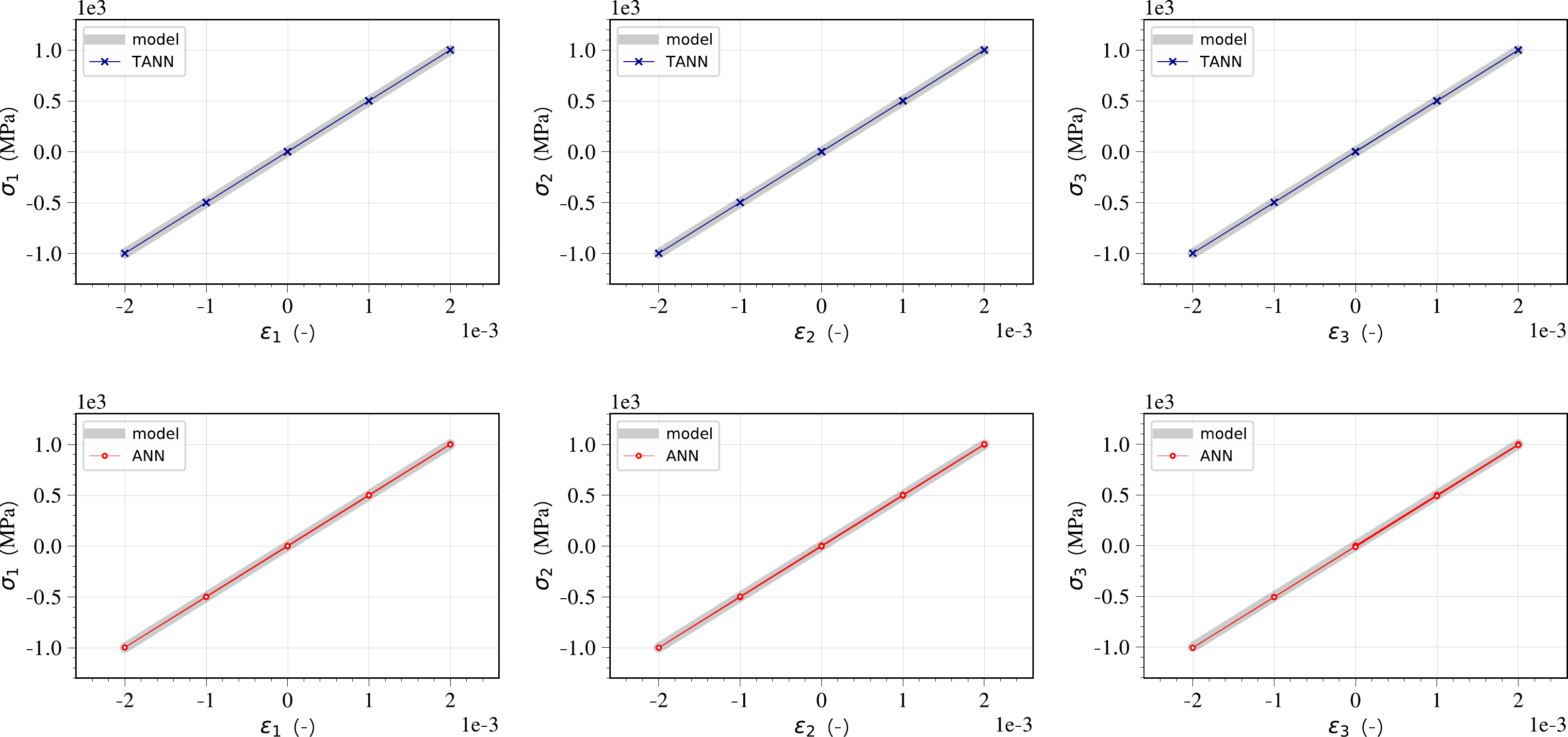}
  \caption{\footnotesize strain increment $\Delta {\varepsilon}=1\times 10^{-3}$.}
\end{subfigure}
\begin{subfigure}{0.735\textwidth}
 \centering
  \includegraphics[width=\textwidth]{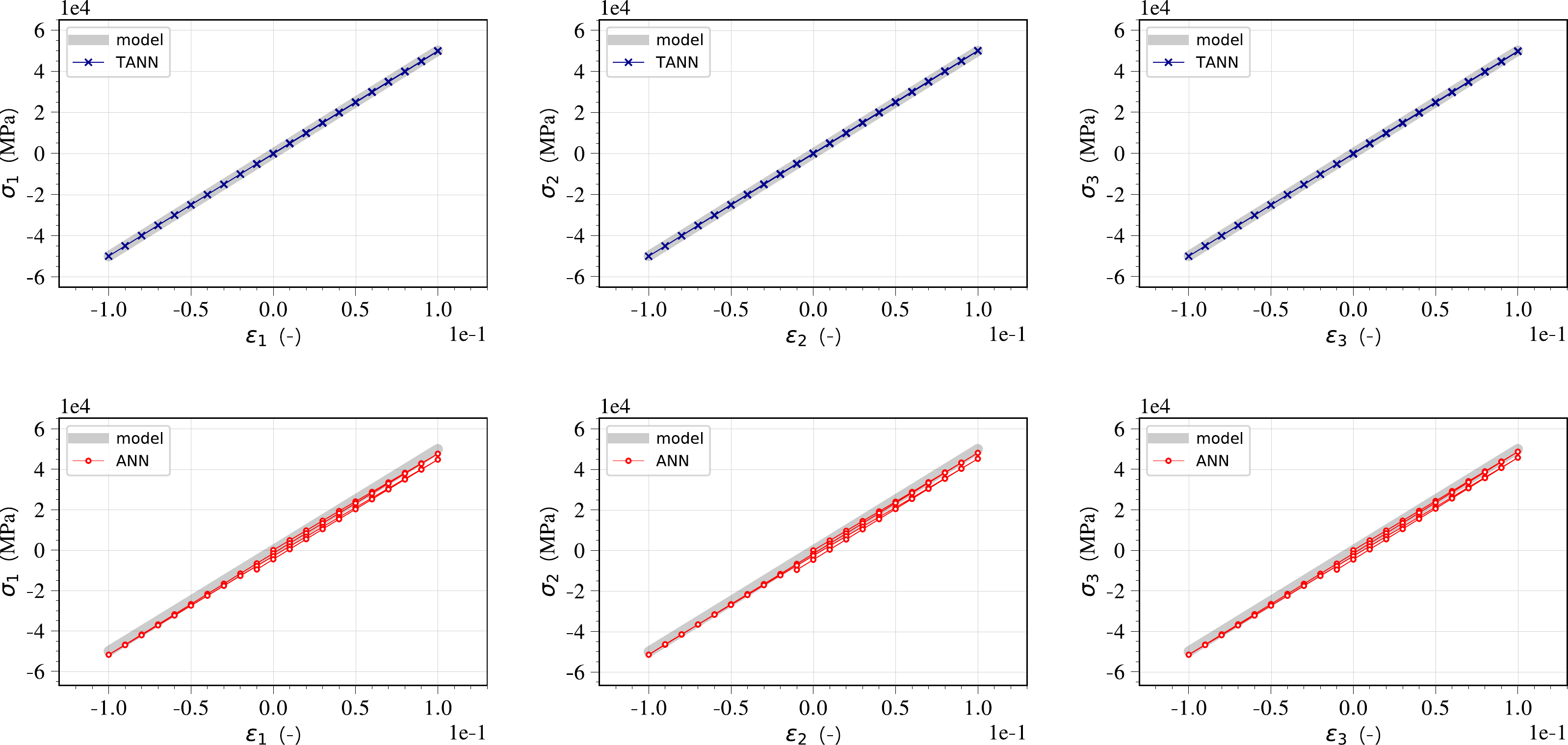}
  \caption{\footnotesize strain increment $\Delta {\varepsilon}=1\times 10^{-2}$.}
\end{subfigure}
\begin{subfigure}{0.735\textwidth}
  \centering
  \includegraphics[width=\textwidth]{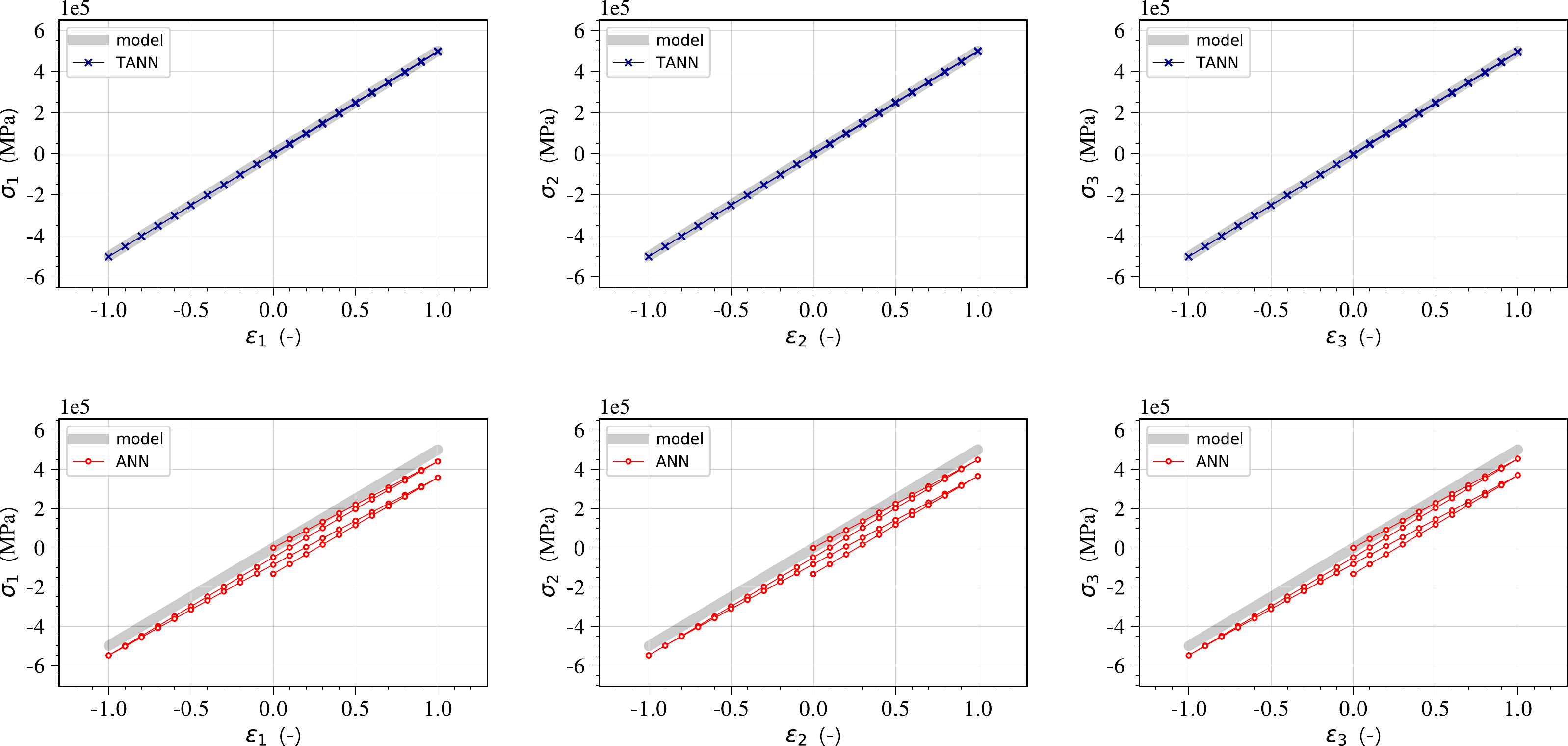}
  \caption{\footnotesize strain increment $\Delta {\varepsilon}=1\times 10^{-1}$.}
\end{subfigure}  
\caption{Comparison of the stress predictions of \textsf{TANN} and standard \textsf{ANN} with respect to the target values, for the cyclic, isotropic loading path $\Delta \varepsilon_1^n = \Delta \varepsilon_2^n = \Delta \varepsilon_3^n = \Delta {\varepsilon} \text{ sgn}\left(\cos \frac{n \pi}{2N} \right)-$with $N= \varepsilon_{\text{max}}/\Delta {\varepsilon}$, $\varepsilon_{\text{max}}=2\times10^{-3}$ (a), $\varepsilon_{\text{max}}=\times10^{-1}$ (b), and $\varepsilon_{\text{max}}=1$ (c), for material case 3D-1 (perfect plasticity). Each row represents the prediction at different $\Delta \varepsilon$ increments.}
\label{f:3D_TANN_vs_ANN_triEL_H0}
\end{figure*}

We first compare the performance of both networks, \textsf{TANN} and standard \textsf{ANN}, in predicting the material response for cyclic isotropic loading paths (material case 3D-1, cf. Table \ref{tab:3D_mat}). A linear elastic material response is expected and retrieved. Figure \ref{f:3D_TANN_vs_ANN_triEL_H0} displays the stress predictions of \textsf{TANN} and \textsf{ANN}, compared with the target values, for different strain increments. It is worth mentioning that the standard approach of \textsf{ANN} does not succeed in accurately predicting the elastic deformation range. Moreover, contrary to \textsf{TANN}, the stress predictions of standard \textsf{ANN}, depend strongly on the cyclic loading. As the network is used recursively, in recall mode, the stress predictions rapidly become less and less precise, due to error accumulation.\\

\begin{figure}[ht]
\centering
\begin{subfigure}[t]{0.245\textwidth}
  \centering
  \includegraphics[height=0.65\textheight]{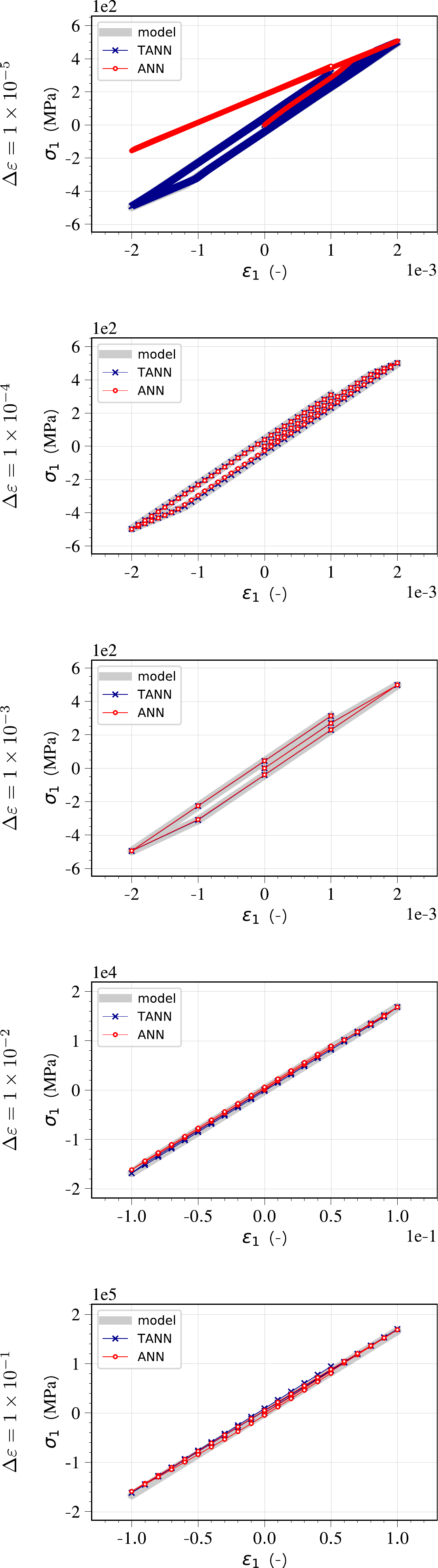}  
  \caption{\scriptsize $\sigma$ prediction}
  \label{f:cyc_3D_pft_TANNa_uni}
\end{subfigure}
\begin{subfigure}[t]{0.245\textwidth}
  \centering
  \includegraphics[height=0.65\textheight]{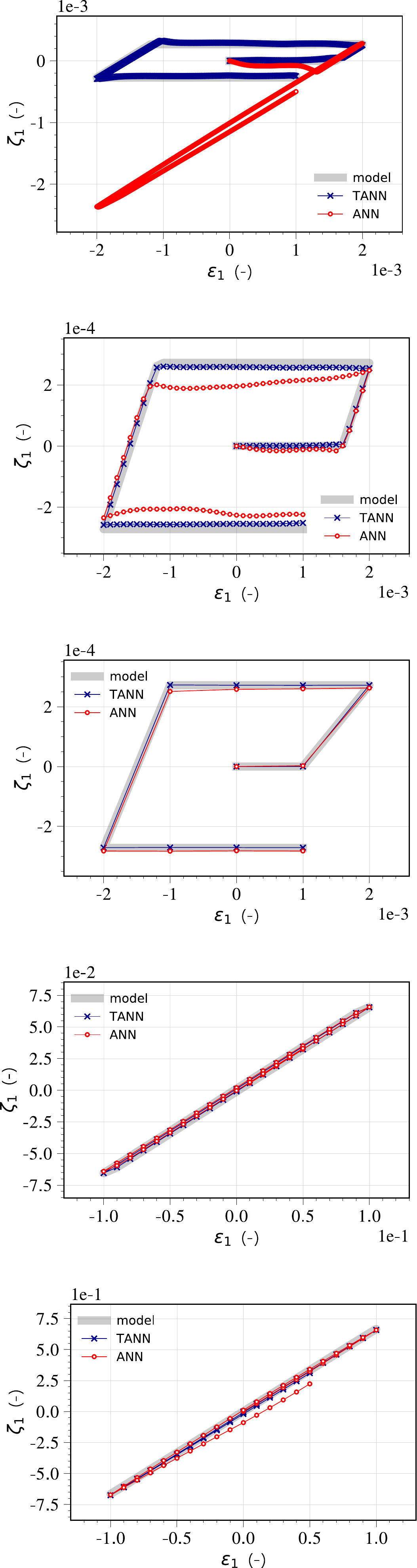}   
  \caption{\scriptsize $\zeta$ prediction}
  \label{f:cyc_3D_pft_TANNb_uni}
\end{subfigure}
\begin{subfigure}[t]{0.245\textwidth}
  \centering
  \includegraphics[height=0.65\textheight]{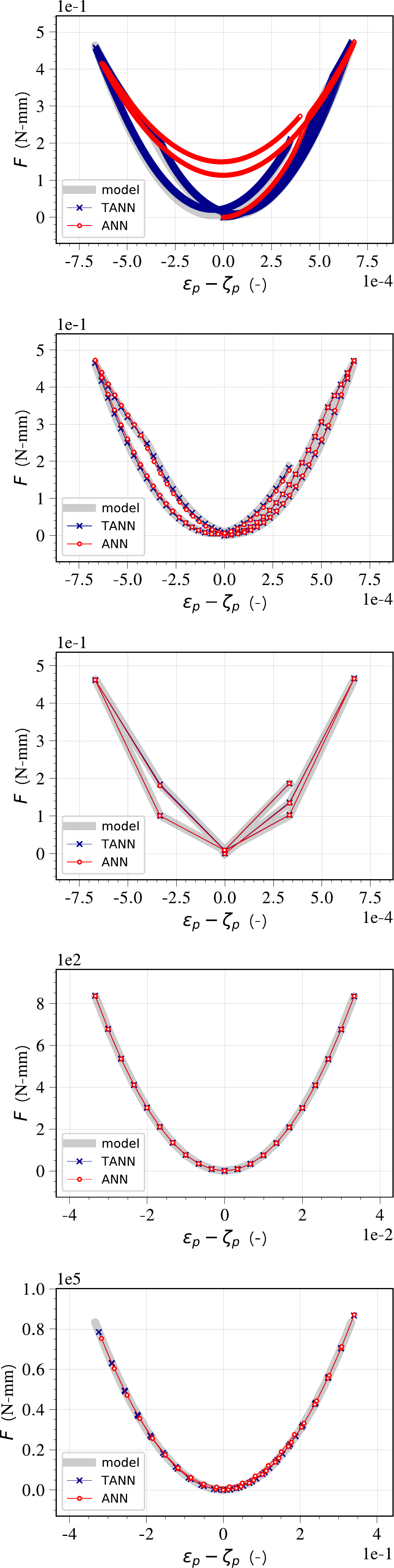}   
  \caption{\scriptsize $\textsf{F}$ prediction (\textsf{TANN}), Eq. (\ref{eq:def3D}) (\textsf{ANN}).}
  \label{f:cyc_3D_pft_TANNc_uni}
\end{subfigure}
\begin{subfigure}[t]{0.245\textwidth}
  \centering
  \includegraphics[height=0.65\textheight]{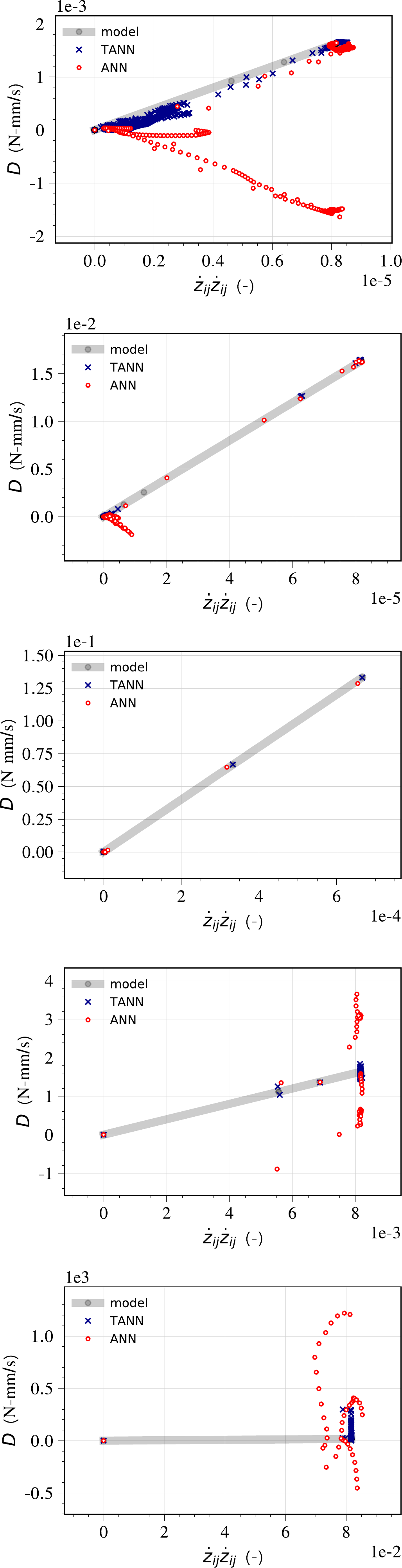}  
  \caption{\scriptsize $\textsf{D}$ prediction (\textsf{TANN}), Eq. (\ref{eq:def3D}) (\textsf{ANN}).}
  \label{f:cyc_3D_pft_TANNd_uni}
\end{subfigure}
\caption{Comparison of the predictions of \textsf{TANN} and standard \textsf{ANN} with respect to the target values, for the uni-axial cyclic loading path, Eq. (\ref{eq:uni_bi_tri}), for material case 3D-1 (perfect plasticity). Each row represents the prediction at different $\Delta \varepsilon$ increments.}
\label{f:3D_ANN_TANN_uni}
\end{figure}

\begin{figure*}[ht]
\centering
  \begin{subfigure}[t]{0.245\textwidth}
  \centering
  \includegraphics[height=0.65\textheight]{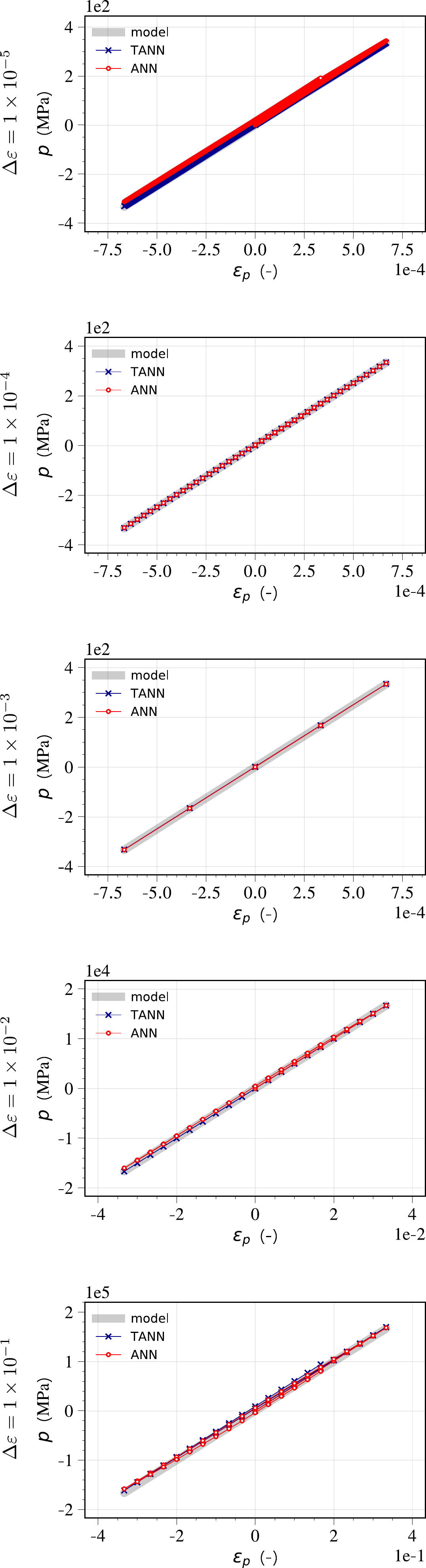}   
  \caption{\scriptsize $p$ computation.}
\end{subfigure}
\begin{subfigure}[t]{0.245\textwidth}
  \centering
  \includegraphics[height=0.65\textheight]{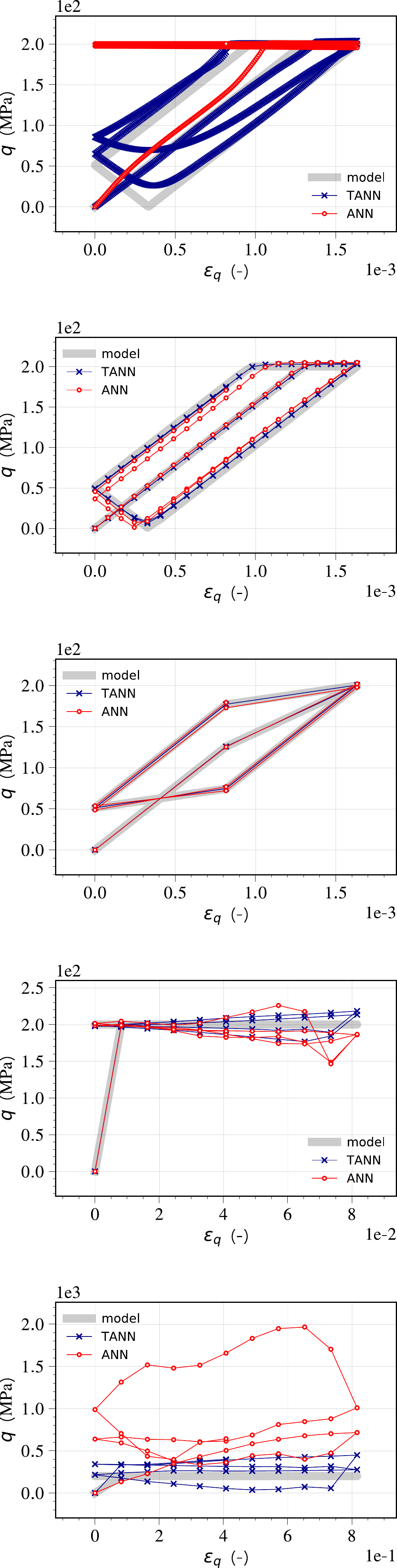}  
  \caption{\scriptsize $q$ computation.}
\end{subfigure}
\caption{Comparison of the stress predictions of \textsf{TANN} and \textsf{ANN} in terms of mean and deviatoric stress, $p$ (left) and $q$ (right), for the uni-axial loading path in Figure \ref{f:3D_ANN_TANN_uni}, Eq. (\ref{eq:uni_bi_tri}). Each row represents the prediction at different $\Delta \varepsilon$ increments.}
\label{f:3D_ANN_TANN_uni_pq}
\end{figure*}

The performance of both networks is further compared for the following three loading paths
\begin{equation}
	\begin{split}
	\text{uni-axial:} \qquad &\Delta \varepsilon_1^n = \Delta {\varepsilon} \text{ sgn}\left(\cos \frac{n \pi}{2N} \right), \quad \Delta \varepsilon_2^n = \Delta \varepsilon_3^n=0;\\
	\text{bi-axial:} \qquad &\Delta \varepsilon_1^n = \Delta {\varepsilon} \text{ sgn}\left(\cos \frac{n \pi}{2N} \right),\quad  \Delta \varepsilon_2^n = -\Delta {\varepsilon} \text{ sgn}\left(\cos \frac{n \pi}{4N} \right), \quad \Delta \varepsilon_3^n=0;\\
	\text{tri-axial:} \qquad &\Delta \varepsilon_1^n = \Delta {\varepsilon} \text{ sgn}\left(\cos \frac{n \pi}{2N} \right), \quad \Delta \varepsilon_2^n = \Delta \varepsilon_3^n=\Delta {\varepsilon} \text{ sgn}\left(\sin \frac{n \pi}{2N} \right),
	\end{split}
	\label{eq:uni_bi_tri}
\end{equation}
with $N= \frac{\varepsilon_{\text{max}}}{\Delta{\varepsilon}}$, $\varepsilon_{\text{max}}=2\times10^{-3}\div 2$, and $\Delta \varepsilon = 10^{-4} \div 1$.\\ 
In Figures \ref{f:3D_ANN_TANN_uni} and \ref{f:3D_ANN_TANN_uni_pq} the results obtained for a uni-axial loading scheme are presented, for different values of the strain increment, and material case 3D-1 (cf. Table \ref{tab:3D_mat}). Figures \ref{f:cyc_3D_pft_TANNa_uni} and \ref{f:cyc_3D_pft_TANNb_uni} display the material response in terms of the principal stress, $\sigma_1$, and inelastic strain, $\zeta_1$, over the principal strain, $\varepsilon_1$. Figures \ref{f:cyc_3D_pft_TANNc_uni} and \ref{f:cyc_3D_pft_TANNd_uni} compare the energy and dissipation rate predicted by \textsf{TANN} with those computed, with standard \textsf{ANN}, directly using the corresponding definitions for the free-energy and dissipation rate, Eq. (\ref{eq:def3D}). The predictions of \textsf{TANN} are in good agreement with the constitutive model, independently from the strain increment, which exceeds considerably the training range. Nevertheless, the performance of \textsf{ANN} is found to be strongly affected by the values of $\Delta \varepsilon$. Standard \textsf{ANN} performs poorly for strain increments smaller and larger than the ones at which it was trained ($\Delta \varepsilon = 1\times 10^3$). Furthermore, standard \textsf{ANN} predicts thermodynamically inconsistent outputs.\\
We emphasize that, even though for relatively large strain increments \textsf{TANN} predictions are less accurate (see e.g. Fig. \ref{f:cyc_3D_pft_TANNd_uni}), predictions remain thermodynamically consistent. Moreover, the network successfully learns the Jacobian, $\frac{\partial \sigma}{\partial \epsilon}$, contrary to standard \textsf{ANN}. As discussed for the 1D case, \textsf{TANN} hence offers compelling capabilities in replacing complicated constitutive models in numerical simulations, with smaller computational cost. Figure \ref{f:3D_ANN_TANN_uni_pq} illustrates the stress predictions, but in terms of the mean and deviatoric stress (computed from the principal stress predictions).\\

\begin{figure}[ht]
\centering
\begin{subfigure}[t]{0.49\textwidth}
  \centering
  \includegraphics[height=0.65\textheight]{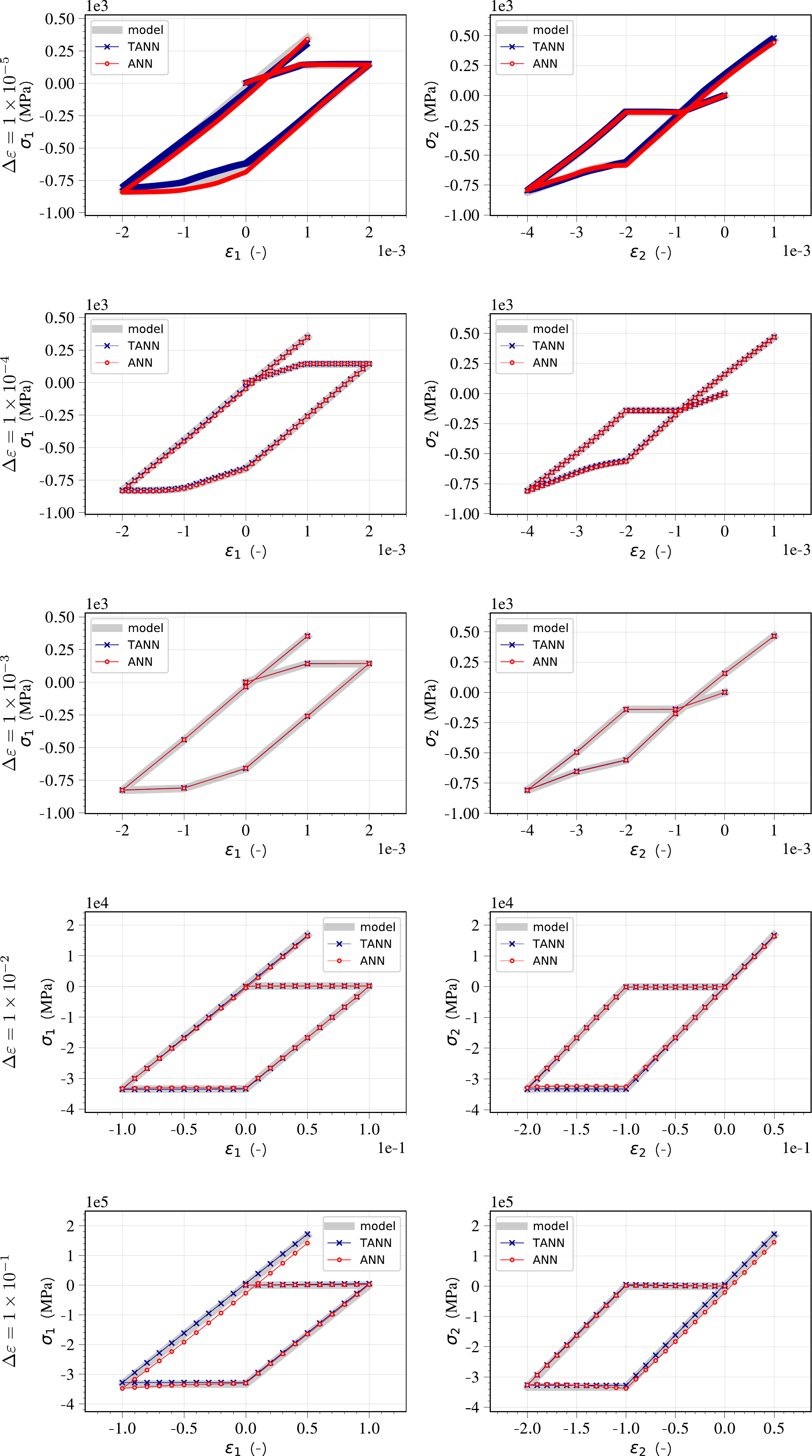}  
  \caption{\scriptsize $\sigma$ prediction.}
  \label{f:cyc_3D_pft_TANNa_bi}
\end{subfigure}
\begin{subfigure}[t]{0.49\textwidth}
  \centering
  \includegraphics[height=0.65\textheight]{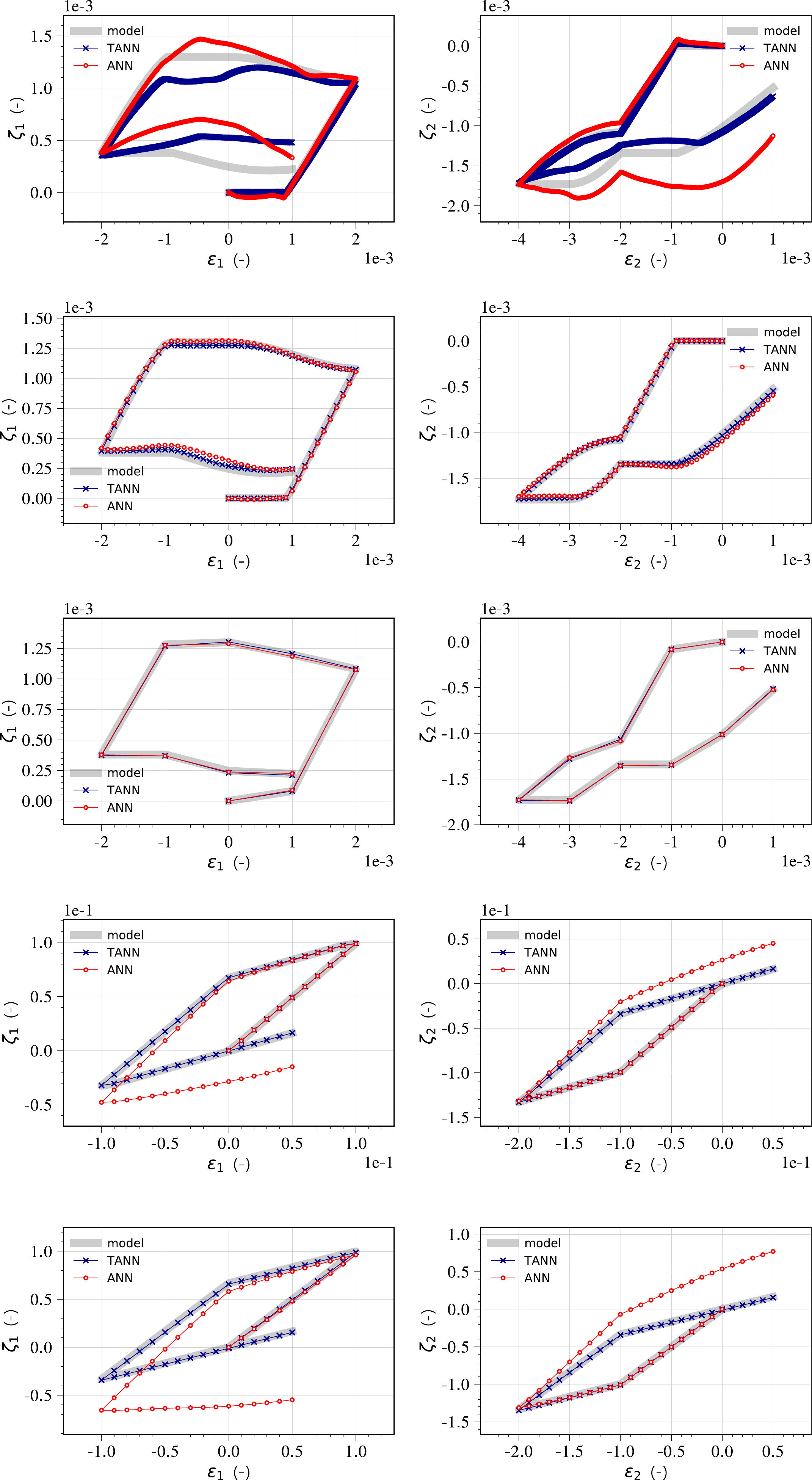}   
  \caption{\scriptsize $\zeta$ prediction.}
  \label{f:cyc_3D_pft_TANNb_bi}
\end{subfigure}
\caption{Comparison of the stress and internal variable predictions of \textsf{TANN} and standard \textsf{ANN} with respect to the target values, for the bi-axial cyclic loading path, Eq. (\ref{eq:uni_bi_tri}), for material case 3D-1 (perfect plasticity). Each row represents the prediction at different $\Delta \varepsilon$ increments.}
\label{f:3D_ANN_TANN_bi}
\end{figure}

\begin{figure}[ht]
\centering
\begin{subfigure}[b]{0.245\textwidth}
  \centering
  \includegraphics[height=0.65\textheight]{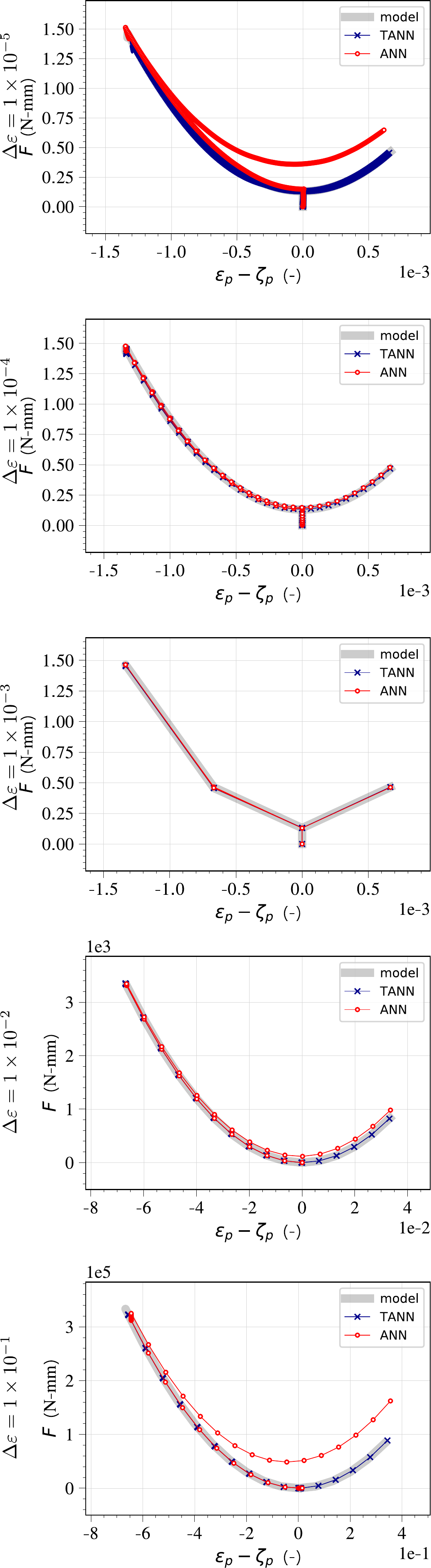}   
  \caption{\scriptsize $\textsf{F}$ prediction (\textsf{TANN}), Eq. (\ref{eq:def3D}) (\textsf{ANN}).}
  \label{f:cyc_3D_pft_TANNc_bi}
\end{subfigure} \hspace{0.25cm}
\begin{subfigure}[b]{0.245\textwidth}
  \centering
  \includegraphics[height=0.65\textheight]{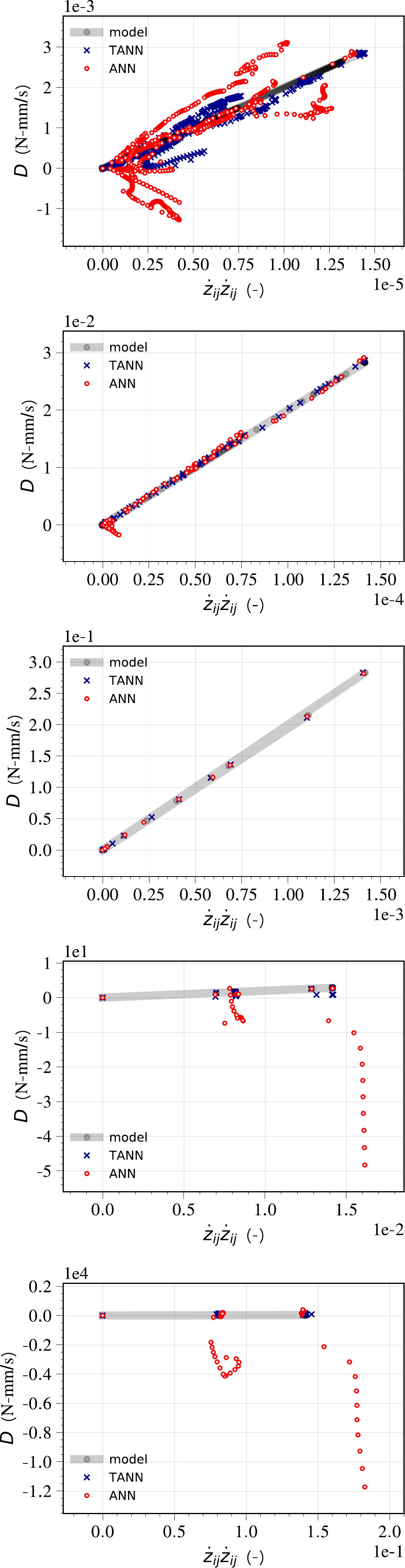}   
  \caption{\scriptsize $\textsf{D}$ prediction (\textsf{TANN}), Eq. (\ref{eq:def3D}) (\textsf{ANN}).}
  \label{f:cyc_3D_pft_TANNcd_bi}
\end{subfigure}
\caption{Comparison of the energy and dissipation rate predictions of \textsf{TANN} and computation according to Eq. (\ref{eq:def3D}) for standard \textsf{ANN} with respect to the target values, for the bi-axial cyclic loading path, Eq. (\ref{eq:uni_bi_tri}), for material case 3D-1 (perfect plasticity). Each row represents the prediction at different $\Delta \varepsilon$ increments.}
\label{f:3D_ANN_TANN_bi_second}
\end{figure}

\begin{figure}[ht]
  \centering
  \includegraphics[height=0.65\textheight]{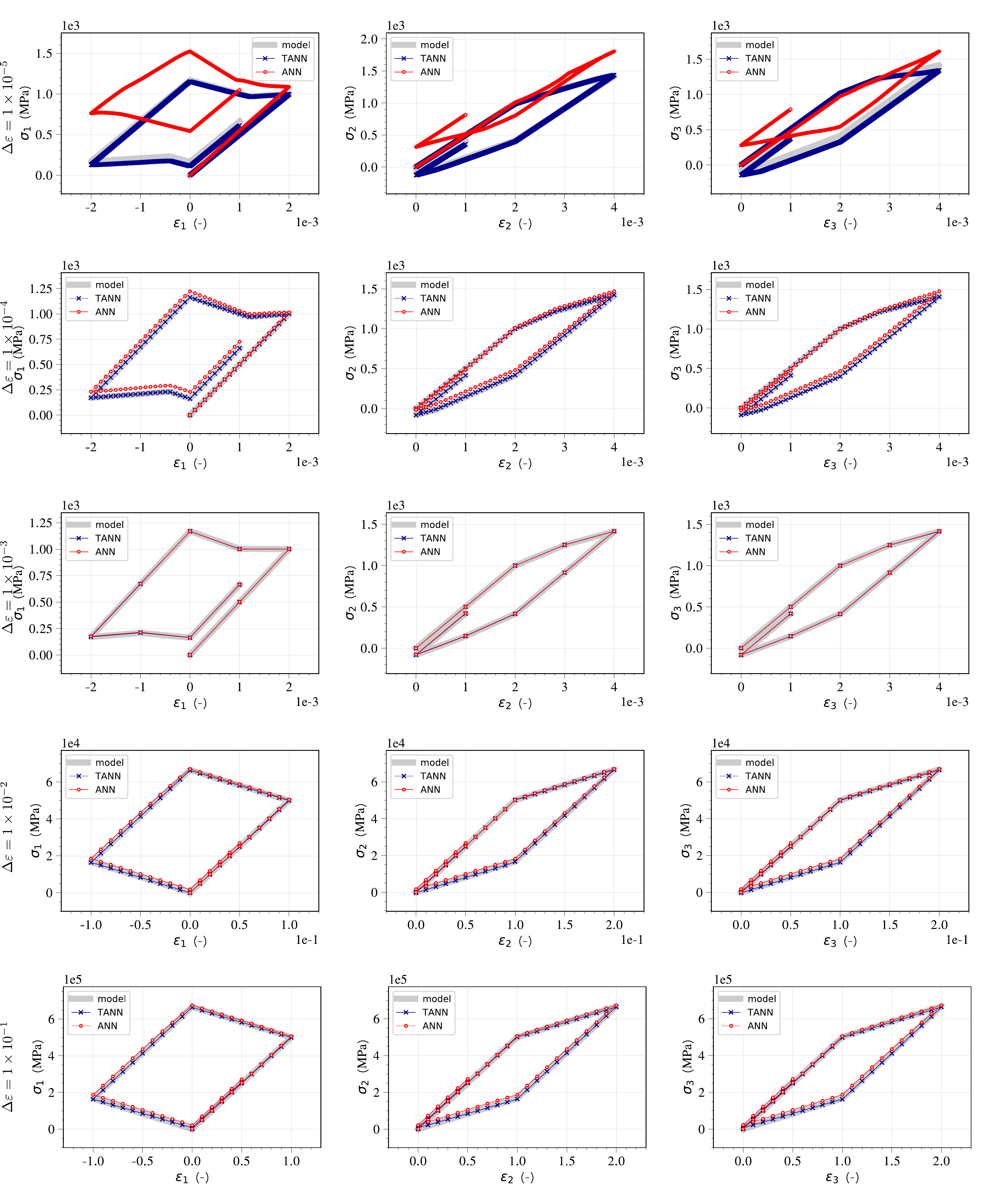}  
\caption{Comparison of the stress predictions of \textsf{TANN} and standard \textsf{ANN} with respect to the target values, for the tri-axial cyclic loading path, Eq. (\ref{eq:uni_bi_tri}), for material case 3D-1 (perfect plasticity). Each row represents the prediction at different $\Delta \varepsilon$ increments.}
\label{f:3D_ANN_TANN_tri_a}
\end{figure}

\begin{figure}[ht]
  \centering
  \includegraphics[height=0.65\textheight]{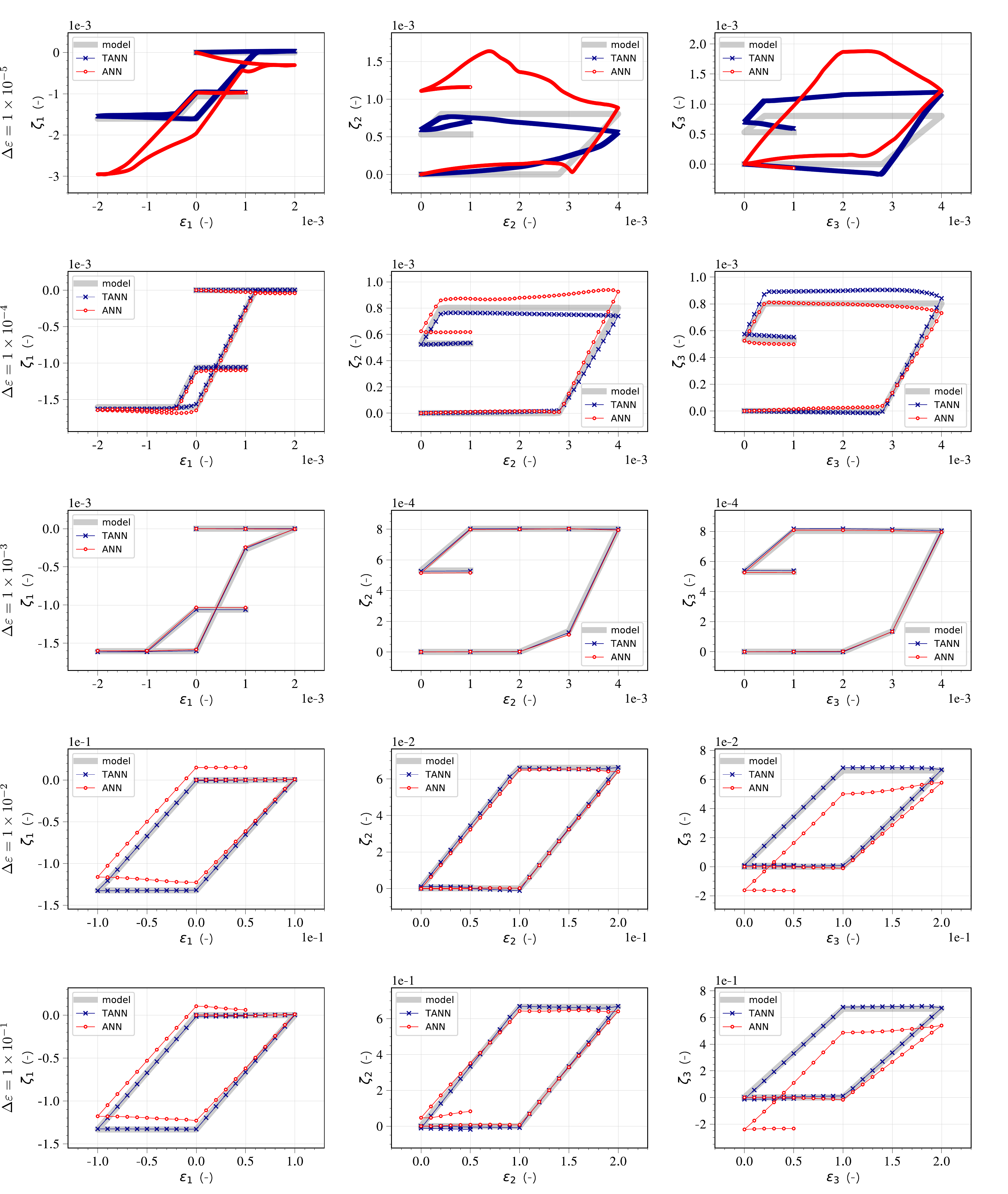}  
\caption{Comparison of the internal variable predictions of \textsf{TANN} and standard \textsf{ANN} with respect to the target values, for the tri-axial cyclic loading path, Eq. (\ref{eq:uni_bi_tri}), for material case 3D-1 (perfect plasticity). Each row represents the prediction at different $\Delta \varepsilon$ increments.}
\label{f:3D_ANN_TANN_tri_b}
\end{figure}

\begin{figure}[ht]
\centering
\begin{subfigure}[b]{0.245\textwidth}
  \centering
  \includegraphics[height=0.65\textheight]{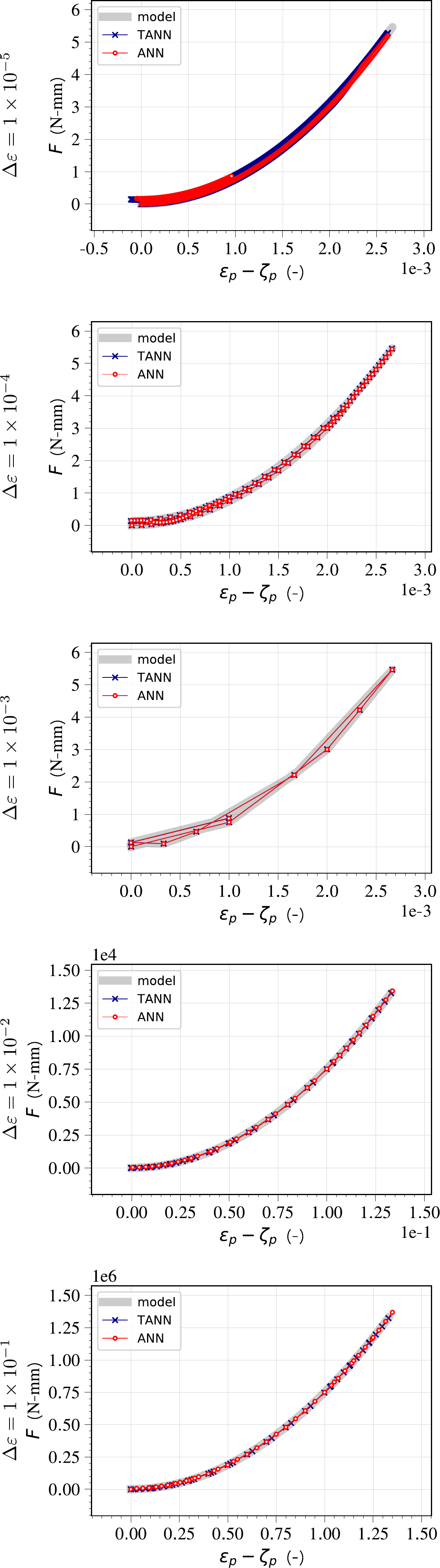}   
  \caption{\scriptsize $\textsf{F}$ prediction (\textsf{TANN}), Eq. (\ref{eq:def3D}) (\textsf{ANN}).}
  \label{f:cyc_3D_pft_TANN_tri_c}
\end{subfigure}\hspace{0.25cm}
\begin{subfigure}[b]{0.245\textwidth}
  \centering
  \includegraphics[height=0.65\textheight]{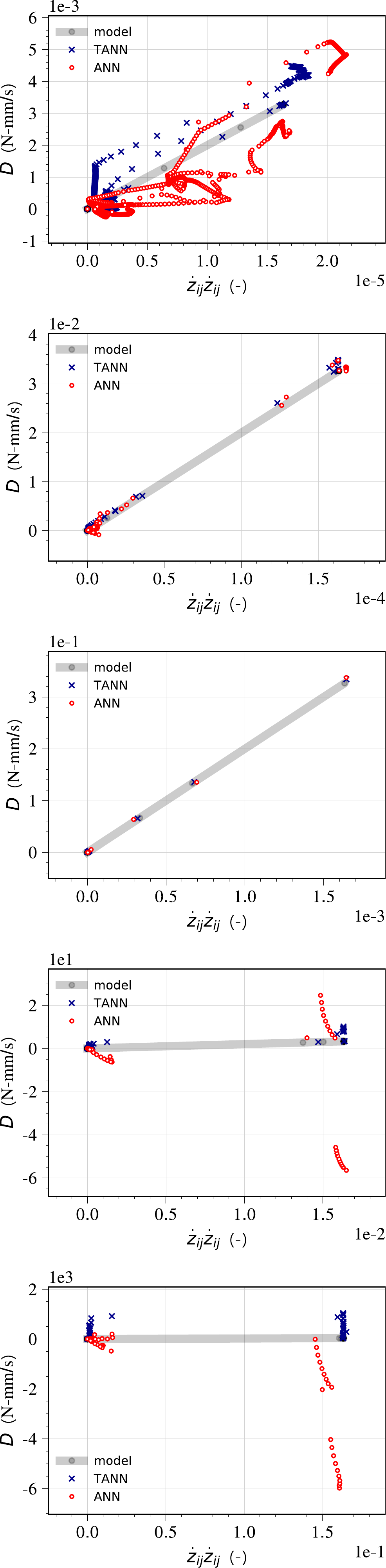}   
  \caption{\scriptsize $\textsf{D}$ prediction (\textsf{TANN}), Eq. (\ref{eq:def3D}) (\textsf{ANN}).}
  \label{f:cyc_3D_pft_TANN_tri_d}
\end{subfigure}
\caption{Comparison of the energy and dissipation rate predictions of \textsf{TANN} and computation according to Eq. (\ref{eq:def3D}) for standard \textsf{ANN} with respect to the target values, for the tri-axial cyclic loading path, Eq. (\ref{eq:uni_bi_tri}), for material case 3D-1 (perfect plasticity). Each row represents the prediction at different $\Delta \varepsilon$ increments.}
\label{f:3D_ANN_TANN_tri_cd}
\end{figure}

Figures \ref{f:3D_ANN_TANN_bi} and \ref{f:3D_ANN_TANN_bi_second} present the predictions of \textsf{TANN} and standard \textsf{ANN} for the bi-axial cyclic path (Eq. (\ref{eq:uni_bi_tri})), in terms of the principal stresses, for the material case 3D-1 (cf. Table \ref{tab:3D_mat}). Similarly, Figures \ref{f:3D_ANN_TANN_tri_a} to \ref{f:3D_ANN_TANN_tri_cd} show the predictions for the tri-axial loading path. The predictions obtained for the same loading path but for material cases 3D-2 (hardening) and 3D-3 (softening) are presented in the SM file. As mentioned above, \textsf{TANN} capabilities of generalizing the predictions with respect to the values assumed by the input variables are remarkably good. Standard \textsf{ANN} succeeds in correctly predicting the stress increments only in a reduced range of the strain increment values, close to the range of its training data. It is worth noticing that in the computation of the mean and deviatoric stress, from the principal stress components,\textsf{TANN} gives relatively higher errors, but still much lower when compared with those of \textsf{ANN}). 
The performance of \textsf{TANN} in predicting $p$ and $q$ can be further improved by small modifications to include these invariants as outputs. However, this kind of optimization exceeds the scope of the current work.\\

Similar to the 1D case, it should be stressed that the performance of \textsf{TANN} and standard \textsf{ANN} can be improved by increasing the dimension of the training data-set, the number of the hyperparameters (e.g. numbers of hidden layers, etc.). Nevertheless, the fundamental gap between the two approaches in assuring thermodynamically consistent quantities still persist.

\clearpage
\section{Concluding remarks}
A new class of artificial neural networks models to replace constitutive laws and predict the material response at the material point level was proposed. The two basic laws of thermodynamics were directly encoded in the architecture of the model, which we refer to as Thermodynamics-based Neural Network (TANN). Our approach was inspired by the so-called Physics-Informed Neural Networks (PINNs) \cite{raissi2019physics}, where the automatic differentiation was used to  perform the numerical calculation of the derivative of a neural network with respect to its inputs.

The numerical requirements regarding the mathematical class of appropriate activation functions to be used together with automatic differentiation were investigated. More specifically, the internal restrictions, derived from the first law of thermodynamics, require activation functions whose second gradient does not vanish. This new problem and its remedy was extensively explored and discussed in the manuscript.

TANN, relying on an incremental formulation and on the theoretical developments in \cite{houlsby2007principles}, posses the special feature that the entire constitutive response of a material can be derived from definition of only two scalar functions: the free-energy and the dissipation rate. This assures thermodynamically consistent predictions both for seen and unseen data. Differently from the standard ANN approaches, TANN does not have to identify, through learning, the underlying thermodynamic laws. Indeed, predictions of standard ANNs may be thermodynamically inconsistent, even though the training of the network has been performed on consistent material data. Being aware of physics, TANNs are found to be a robust approach. If the training data-sets are not thermodynamically consistent, the training operation of the network will be unsuccessful, contrary to the standard \textsf{ANN} approach (see SM).

For the cases here investigated, we showed that TANNs are characterized by high accuracy of the predictions, higher than those of standard approaches. The integration of thermodynamic principles inside the network renders TANN's ability of generalization (i.e., make predictions for loading paths different from those used in the training operation) remarkably good. Consequently, TANN is an excellent candidate for replacing constitutive calculations at Finite Element incremental formulations. Moreover, thanks to the implementation of the free-energy in the network predictions and its thermodynamical relation with the stresses, the Jacobian $\frac{\partial \Delta \sigma}{\Delta \varepsilon}$ at the material point level is better predicted even for increments far outside the training data-set range. As a result quadratic convergence in implicit formulations can be preserved, reducing the calculation cost.

Further extensions of \textsf{TANN} in a wide range of applications, for complex materials, are straightforwards, as the thermodynamics principles hold true for any known class of material, at any length (micro- and macro-scale).

\section*{Acknowledgments}
The author I.S. would like to acknowledge the support of the European Research Council (ERC) under the European Union Horizon 2020 research and innovation program (Grant agreement ID 757848 CoQuake).

\clearpage
\section*{Appendix A. Derivation of the incremental material formulation}
By differentiating the energy expressions (\ref{eq:energy_expr}) and rearranging the terms, we obtain the following non-linear incremental relations
\begin{subequations}
	\begin{align}
	\dot{\sigma} = \partial_{\varepsilon \varepsilon}\text{\textsf{F}} \cdot \varepsilon + \sum_k \partial_{\varepsilon \zeta_k}\text{\textsf{F}}\cdot \dot{\zeta}_k + \partial_{\varepsilon \theta} \text{\textsf{F}} \,\dot{\theta}\\
	-\dot{\chi}_i = \partial_{\zeta_i \varepsilon}\text{\textsf{F}} \cdot \varepsilon + \sum_k \partial_{\zeta_i \zeta_k}\text{\textsf{F}}\cdot \dot{\zeta}_k + \partial_{\zeta_i \theta} \text{\textsf{F}} \,\dot{\theta}\\
	-\dot{\text{\textsf{S}}} = \partial_{\theta \varepsilon}\text{\textsf{F}} \cdot \varepsilon + \sum_k \partial_{\theta \zeta_k}\text{\textsf{F}}\cdot \dot{\zeta}_k + \partial_{\theta \theta} \text{\textsf{F}} \,\dot{\theta},
\end{align}
\end{subequations}
where the following notation is adopted
\begin{eqnarray*}
\partial_{\varepsilon \varepsilon}\text{\textsf{F}} =& \dfrac{\partial^2 \text{\textsf{F}}}{\partial \varepsilon_{ij} \partial \varepsilon_{kl}}, \quad
\partial_{\varepsilon \zeta_k}\text{\textsf{F}} =& \dfrac{\partial^2 \text{\textsf{F}}}{\partial \varepsilon_{ij} \partial \zeta_k},\\
\partial_{\varepsilon \theta}\text{\textsf{F}} = &\dfrac{\partial^2 \text{\textsf{F}}}{\partial \varepsilon_{ij} \partial \theta},\quad
\partial_{\theta \theta}\text{\textsf{F}} =& \dfrac{\partial^2 \text{\textsf{F}}}{\partial  \theta^2}.
\end{eqnarray*}
We introduce the thermodynamic dissipative stresses $\mathcal{X}^{\dagger}= (X_1, \ldots, X_N)$ with 
\begin{equation}
X_i :=  \dfrac{\partial  \text{\textsf{D}}}{\partial \dot{\zeta}_i} \qquad \forall \: i\in [1,N].
\end{equation}
For a rate-independent material, the dissipation is a homogeneous first-order function in the internal variable rates $\dot{\zeta}_i$ \cite{houlsby2007principles}. This homogeneity can be expressed by the Euler's relation
\begin{equation}
\text{\textsf{D}} = \sum_{i=1}^N \dfrac{\partial  \text{\textsf{D}}}{\partial \dot{\zeta}_i} \cdot \dot{\zeta}_i= \sum_i X_i \cdot \dot{\zeta}_i,
\label{eqh:normality}
\end{equation}
which, together with (\ref{eq:chi}), implies
\begin{equation}
\sum_{i=1}^{N} \left(X_i-\chi_i \right) \cdot \dot{\zeta}_i = 0
\end{equation}
Ziegler's orthogonality condition \cite{ziegler2012introduction} is further assumed, i.e., $X_i=\chi_i$ $\forall \: i \in [1,N]$.\\
Being $\text{\textsf{D}}$ homogeneous first-order function in $\dot{\zeta}_i$, the Legendre transform, conjugate to $X_i$, is degenerate, that is equal to zero, and represents the yield function $y=\tilde{y}(\theta, \varepsilon, \mathcal{Z}, \mathcal{X}^{\dagger})$, i.e.
\begin{equation}
\lambda y = \sum_i X_i \cdot \dot{\zeta}_i - \text{\textsf{D}} = 0,
\end{equation}
where $\lambda$ is a non-negative multiplier. From the properties of Legendre transform, the following flow rules must hold
\begin{equation}
\dot{\zeta}_i = \lambda \dfrac{\partial y}{\partial X_i} \quad \forall \: i \in [1,N].
\label{eq:flowrule}
\end{equation}
Since $\lambda \geq 0$ and $\lambda y =0$, $y\leq 0$. If $y=0$, the following consistency equation is met
\begin{equation}
\dot{y} = \frac{\partial y}{\partial \varepsilon} \cdot \dot{\varepsilon} + \sum_{i=1}^N \frac{\partial y}{\partial \zeta_i} \cdot \dot{\zeta}_i +\sum_{i=1}^N \frac{\partial y}{\partial X_i}\cdot \dot{X}_i + \frac{\partial y}{\partial \theta} \; \dot{\theta} = 0.
\end{equation}
By further using the flow rules (\ref{eq:flowrule}) and Ziegler's normality condition, we obtain
\begin{equation}
\lambda = -\dfrac{\mathcal{C}_{\varepsilon}}{B}\cdot \dot{\varepsilon} - \dfrac{\mathcal{C}_{\theta}}{B}\cdot \dot{\theta},
\end{equation}
with
\begin{equation*}
\mathcal{C}_{\varepsilon} = \frac{\partial y}{\partial \varepsilon} - \sum_{i=1}^N \frac{\partial y}{\partial X_i} \cdot \partial_{\zeta_i \varepsilon}\text{\textsf{F}},
\end{equation*}
\begin{equation*}
\mathcal{C}_{\theta} = \frac{\partial y}{\partial \theta} - \sum_{i=1}^N \frac{\partial y}{\partial X_i} \cdot \partial_{\zeta_i \theta}\text{\textsf{F}},
\end{equation*}
and
\begin{equation*}
B = \sum_{i=1}^N  \frac{\partial y}{\partial \zeta_i} \cdot \frac{\partial y}{\partial X_i} - \sum_{i=1}^N  \frac{\partial y}{\partial X_i} \left( \sum_{k=1}^N  \partial_{\zeta_k \varepsilon}\text{\textsf{F}} \cdot \frac{\partial y}{\partial X_k}\right).
\end{equation*}
Finally, we arrive to the following, incremental non-linear formulation, for $y=0$,
\begin{equation}
\dot{\Xi} = \mathcal{M}|_{y=0} \;\dot{\xi},\quad  \text{ with } \quad  
\label{eq:incr_pl_}
\dot{\Xi}=
\begin{bmatrix} 
\dot{\sigma}\\
-\dot{X}_i\\
-\dot{\text{\textsf{S}}}\\
\dot{\zeta_i}\\
\lambda
\end{bmatrix},
\quad
\dot{\xi}=
\begin{bmatrix} 
\dot{\varepsilon}\\
\dot{\theta}
\end{bmatrix},\quad
\mathcal{M}|_{y=0}
= 
\begin{bmatrix} 
\textsf{M}_{\varepsilon \varepsilon}  & \textsf{M}_{\varepsilon \theta} \\
\textsf{M}_{\zeta \varepsilon} & \textsf{M}_{\zeta \theta}\\
\textsf{M}_{\theta \varepsilon} &\textsf{M}_{\theta \theta} \\
-\dfrac{\mathcal{C}_{\varepsilon}}{B}\cdot\frac{\partial y}{\partial X_i}& - \dfrac{\mathcal{C}_{\theta}}{B}\cdot \frac{\partial y}{\partial X_i}\\
-\dfrac{\mathcal{C}_{\varepsilon}}{B}\cdot & - \dfrac{\mathcal{C}_{\theta}}{B}
\end{bmatrix},
\end{equation}
and
\begin{eqnarray*}
\textsf{M}_{\varepsilon \varepsilon} =& \partial_{\varepsilon \varepsilon}\text{\textsf{F}} - \sum_k \partial_{\varepsilon \zeta_k}\text{\textsf{F}} \cdot \left(\frac{\mathcal{C}_{\varepsilon}}{B}\cdot\frac{\partial y}{\partial X_k}\right),\\
\textsf{M}_{\varepsilon \theta} =& \partial_{\varepsilon \theta}\text{\textsf{F}} - \sum_k \partial_{\varepsilon \zeta_k}\text{\textsf{F}} \cdot\left(\frac{\mathcal{C}_{\theta}}{B}\cdot \frac{\partial y}{\partial X_k}\right),\\
\textsf{M}_{\zeta \varepsilon} =& \partial_{\zeta_i \varepsilon}\text{\textsf{F}} - \sum_k \partial_{\zeta_i \zeta_k}\text{\textsf{F}} \cdot \left( \frac{\mathcal{C}_{\varepsilon}}{B} \cdot\frac{\partial y}{\partial X_k}\right),\\
\textsf{M}_{\zeta \theta} =&\partial_{\zeta_i \theta}\text{\textsf{F}} - \sum_k \partial_{\zeta_i \zeta_k}\text{\textsf{F}} \cdot\left( \frac{\mathcal{C}_{\theta}}{B}\cdot \frac{\partial y}{\partial X_k}\right),\\
\textsf{M}_{\theta \varepsilon} =&\partial_{\theta \varepsilon}\text{\textsf{F}} - \sum_k \partial_{\theta \zeta_k}\text{\textsf{F}} \cdot \left(\frac{\mathcal{C}_{\varepsilon}}{B}\cdot\frac{\partial y}{\partial X_k} \right),\\
\textsf{M}_{\theta \theta} =&\partial_{\theta \theta}\text{\textsf{F}} - \sum_k \partial_{\theta \zeta_k}\text{\textsf{F}} \cdot \left( \frac{\mathcal{C}_{\theta}}{B}\cdot \frac{\partial y}{\partial X_k}\right).
\end{eqnarray*}
In case of $y <0$, relation (\ref{eq:incr_pl_}) becomes
\begin{equation}
\label{eq:incr_pl}
\dot{\Xi} = \mathcal{M}|_{y<0} \;\dot{\xi}, \quad  \text{ with } \quad \mathcal{M}|_{y<0}
= 
\begin{bmatrix} 
\partial_{\varepsilon \varepsilon}\text{\textsf{F}} & \partial_{\varepsilon \theta}\text{\textsf{F}} \\
\partial_{\zeta_i \varepsilon}\text{\textsf{F}} & \partial_{\zeta_i \theta}\text{\textsf{F}}\\
\partial_{\theta \varepsilon}\text{\textsf{F}} & \partial_{\theta \theta}\text{\textsf{F}} \\
0& 0\\
0& 0
\end{bmatrix}.
\end{equation}
\newpage
\section*{References}
\bibliography{Bibliography}  

\end{document}